\pdfoutput=1
\documentclass[pdflatex,sn-nature,iicol]{sn-jnl}

\usepackage{graphicx}
\usepackage{multirow}
\usepackage{amsmath,amssymb,amsfonts}
\usepackage{booktabs}
\usepackage{array}
\usepackage{xurl}

\usepackage{float}
\usepackage{tabularx}
\usepackage{enumitem}
\usepackage{xcolor}
\usepackage{colortbl}
\usepackage[T1]{fontenc}
\usepackage{newtxtext}
\usepackage[cmintegrals]{newtxmath}
\usepackage[scaled=0.92]{helvet}
\usepackage{textcomp}
\makeatletter
\definecolor{natgrey}{RGB}{242,242,242}
\def\sectionfont{\reset@font\sffamily\fontsize{11bp}{13bp}\bfseries\selectfont\raggedright}
\def\subsectionfont{\reset@font\sffamily\fontsize{9bp}{11bp}\bfseries\selectfont\raggedright}
\def\subsubsectionfont{\reset@font\sffamily\fontsize{8.5bp}{10.5bp}\bfseries\itshape\selectfont\raggedright}
\def\paragraphfont{\reset@font\bfseries\selectfont}
\renewcommand\section{\@startsection{section}{1}{\z@}{-10pt \@plus -3pt \@minus -2pt}{4pt}{\sectionfont}}
\renewcommand\subsection{\@startsection{subsection}{2}{\z@}{-8pt \@plus -3pt \@minus -2pt}{2pt}{\subsectionfont}}
\renewcommand\subsubsection{\@startsection{subsubsection}{3}{\z@}{-6pt \@plus -2pt \@minus -1pt}{1pt}{\subsubsectionfont}}
\renewcommand\paragraph{\@startsection{paragraph}{4}{\z@}{4pt \@plus 1pt \@minus 1pt}{-0.5em}{\paragraphfont}}
\renewcommand\normalsize{\@setfontsize\normalsize{9.5bp}{11.5bp}%
  \abovedisplayskip 6\p@ \@plus2\p@ \@minus2\p@ \abovedisplayshortskip \z@ \@plus2\p@
  \belowdisplayshortskip 4\p@ \@plus2\p@ \@minus2\p@ \belowdisplayskip \abovedisplayskip
  \let\@listi\@listI}
\normalsize
\def\fnum@figure{{\sffamily\bfseries\selectfont\figurename\space\thefigure\space|}}
\def\fnum@table{{\sffamily\bfseries\selectfont\tablename\space\thetable\space|}}
\def\figurecaptionfont{\reset@font\fontsize{7.5bp}{9bp}\selectfont}
\def\tablecaptionfont{\reset@font\fontsize{7.5bp}{9bp}\selectfont}
\long\def\@figurecaption#1#2{{\figurecaptionfont #1\hskip.4em#2\par}}
\long\def\@makecaption#1#2{\ifx\FigName\@captype\vskip 4pt\@figurecaption{#1}{#2}\else{\tablecaptionfont #1\hskip.4em#2\par}\vskip 4pt\fi}
\long\def\abstract#1{\def\@abstract{{\fontsize{10bp}{12.5bp}\bfseries\selectfont #1\par}\vskip 6pt}}
\newcounter{natbox}
\newsavebox{\natboxbin}
\newenvironment{natbox}[1]{\begin{figure*}[t]\refstepcounter{natbox}\begin{lrbox}{\natboxbin}\begin{minipage}{\dimexpr\textwidth-2\fboxsep\relax}\vspace{3pt}%
  {\sffamily\bfseries\small Box \thenatbox\space|\space #1\par}\vspace{4pt}\small}{\vspace{3pt}\end{minipage}\end{lrbox}\noindent\colorbox{natgrey}{\usebox{\natboxbin}}\end{figure*}}
\newcommand{\tabsize}{\fontsize{7bp}{8.6bp}\selectfont\renewcommand{\arraystretch}{1.1}}

\definecolor{stpos}{RGB}{31,122,118}\definecolor{stneg}{RGB}{196,72,60}\definecolor{stoff}{RGB}{226,229,229}

\makeatother
\makeatletter
\def\Titlefont{\reset@font\sffamily\fontsize{18bp}{22bp}\bfseries\selectfont\raggedright}
\makeatother
\makeatletter
\providecommand{\natauthors}{}
\renewcommand{\@maketitle}{\null\vskip-14pt\hsize\textwidth\parindent0pt
  {\sffamily\fontsize{8.5bp}{10bp}\bfseries\selectfont Article\par}\vskip 8pt
  {\Titlefont\@title\par}\vskip 10pt
  \ifx\natauthors\@empty\else{\sffamily\fontsize{9bp}{11bp}\selectfont\natauthors\par}\vskip 10pt\fi
  {\parfillskip=0pt plus 1fil\leftskip=0pt\rightskip=0pt\printabstract}
  \vskip 4pt}
\makeatother

\newcommand{\fitdisplay}[1]{\begingroup\sbox0{$\displaystyle #1$}\ifdim\wd0>\linewidth\resizebox{\linewidth}{!}{\usebox0}\else\usebox0\fi\endgroup}

\hypersetup{pdftitle={Hippocampal consolidation without replay in fully fine-tuned networks},pdfauthor={Keuntae Kim, Yong Suk Choi}}
\begin{document}
\title[DeltaHippo]{Hippocampal consolidation without replay in fully fine-tuned networks}
\renewcommand{\natauthors}{Keuntae Kim\textsuperscript{1} and Yong Suk Choi\textsuperscript{1,*}\par\vskip 3pt
{\fontsize{8bp}{10bp}\selectfont\textsuperscript{1}Department of Computer Science, Hanyang University, Seoul, Republic of Korea\par
\textsuperscript{*}Correspondence and requests for materials should be addressed to Yong Suk Choi (\href{mailto:cys@hanyang.ac.kr}{cys@hanyang.ac.kr}).}}

\abstract{Networks that keep learning overwrite what they know. Prevailing remedies replay stored data, distil from old models or freeze the backbone. The brain suggests a different division of labour: the hippocampus indexes where experience lies in cortex while the cortex keeps learning. We show that this division lets a fully trainable network keep learning without cross-task replay, a teacher or inference-time memory. After each task, a training-only hippocampal memory indexes the input geometry of every group of synapses sharing an input, residual energy included, and limits each synapse's movement per direction before and after the optimizer. This class-free geometry carries most of the protection; with conflict detection on top, the method cuts absorption of established classes by near-duplicates several-fold against every compared method. To our knowledge, this is the first demonstration of one principle keeping whole networks learning across four architecture families and models up to 8 billion parameters. Methods that freeze the backbone hold up only as far as its representation already separates the classes; here the whole network keeps learning to tell near-duplicates apart, leading the frozen-backbone baseline by 17.5 points on TACRED. Without the memory, much old knowledge stays decodable but beyond the network's reach.}
\maketitle

Artificial neural networks overwrite what they know when they learn something new, a failure known as catastrophic forgetting\textsuperscript{1–3}. Gradient descent reduces the error on the data in front of it, so every step pulls the shared weights towards the newest task. Large pre-trained models are no exception. In class-incremental learning (Fig. 1a) a model learns a sequence of tasks, each adding new categories, sees only the present task's data, and must afterwards answer for every category seen so far without being told the task. When Llama-3.1-8B learns the 40 relations of TACRED\textsuperscript{4} as eight tasks by ordinary fine-tuning, following the protocol of Zheng et al.\textsuperscript{5}, it answers 97.5\% of the first task's questions right after learning them and none after one more task (one run; Fig. 1b); averaged over all tasks it ends at 37.6\% (three runs).

The methods that work best avoid forgetting by keeping the past at hand or by keeping the network from changing (Fig. 1c). Replay stores or generates earlier examples\textsuperscript{6–9}, distillation keeps a copy of the earlier model\textsuperscript{10}, and freezing stops the backbone and trains only prompts, adapters, low-rank modules or a classifier on top\textsuperscript{11–15}. With large pre-trained models freezing has become the default\textsuperscript{5}. A frozen backbone, however, leans on what pre-training already gave it, and holds up only as far as that representation already separates the classes to come. Projection methods let every weight change but steer updates away from the input directions that earlier tasks used\textsuperscript{16–18}; carried to a whole pre-trained network, the constraint must survive adaptive optimizers that rescale every step, the part of the past that a compressed memory does not store must still be protected, and every weight, output head included, must be covered.

One failure is shared by every remedy we test. Later tasks bring near-duplicate classes, categories that ask for the same kind of attribute and differ in one respect, such as a person's \emph{city of death} learnt after their \emph{city of residence}. When the later class is learnt, its internal states at the answer token fall on the earlier class's decision region, and the earlier class's examples begin to be answered as the later one; we call this absorption. Protecting stored data, an old model, a fixed representation or the input directions of earlier tasks does not stop it, because the two classes rely on the same directions. The failure is local and nearly invisible in average accuracy, and preventing it requires recognising the collision as it happens and changing how the network learns there.

Hippocampal accounts suggest a different design. In complementary learning systems, the neocortex keeps learning while the hippocampus governs how experience is laid down in it\textsuperscript{19,20}. The hippocampus keeps an index of where experience lies rather than the experience itself\textsuperscript{21,22}, plasticity depends on what a synapse has carried\textsuperscript{23,24}, and incoming experience is compared with what is stored, so that what is new can be learnt without overwriting what is old\textsuperscript{25–27}. We call the resulting hypothesis \emph{consolidation by plasticity indexing}: a memory can protect what the cortex knows without reinstating it, by indexing where that knowledge lies, detecting when new learning collides with it, and steering the cortex to learn the new while keeping the old apart.

DeltaHippo tests this hypothesis by pairing a fully fine-tuned network, the cortex, with a memory used only during training, the hippocampus (Fig. 1d–f). No earlier task is replayed, no model copy is kept and no parameter tensor is frozen; at test time the memory is removed. The same principle lets the whole network keep learning in ten models from four families, from a 25-million-parameter convolutional network to 8-billion-parameter language models. We then show, in turn, that what the memory holds decides what is kept, that a class-free geometric core with a two-sided optimizer constraint and a residual floor carries most of the effect, that conflict detection keeps near-duplicate classes that every compared method loses, and that what all of this preserves is the compatibility between a changing representation and the output that answers with it.

\begin{figure*}[!p]
\centering
\includegraphics[width=0.865\textwidth]{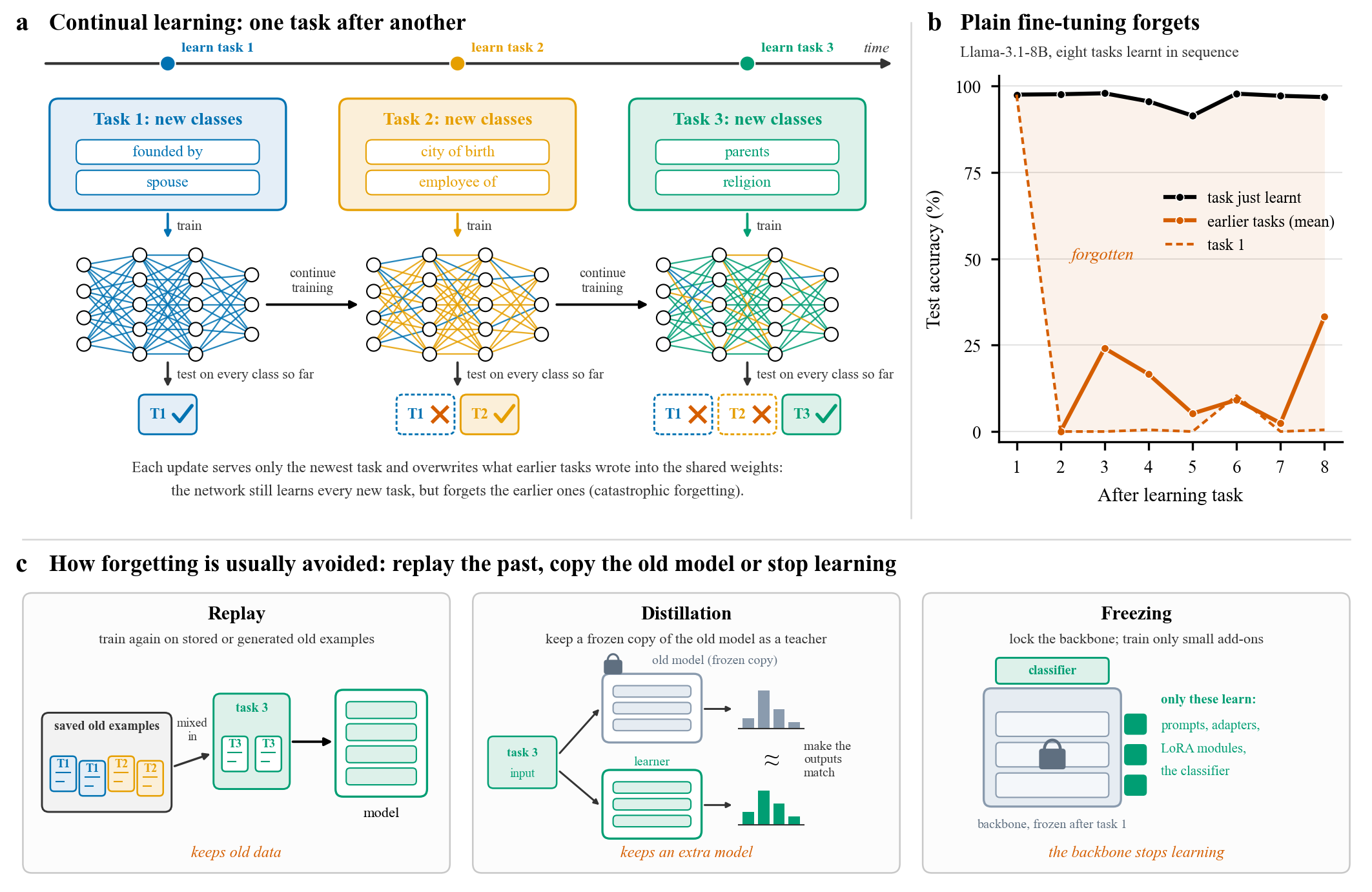}\par\vspace{2pt}
\includegraphics[width=0.865\textwidth]{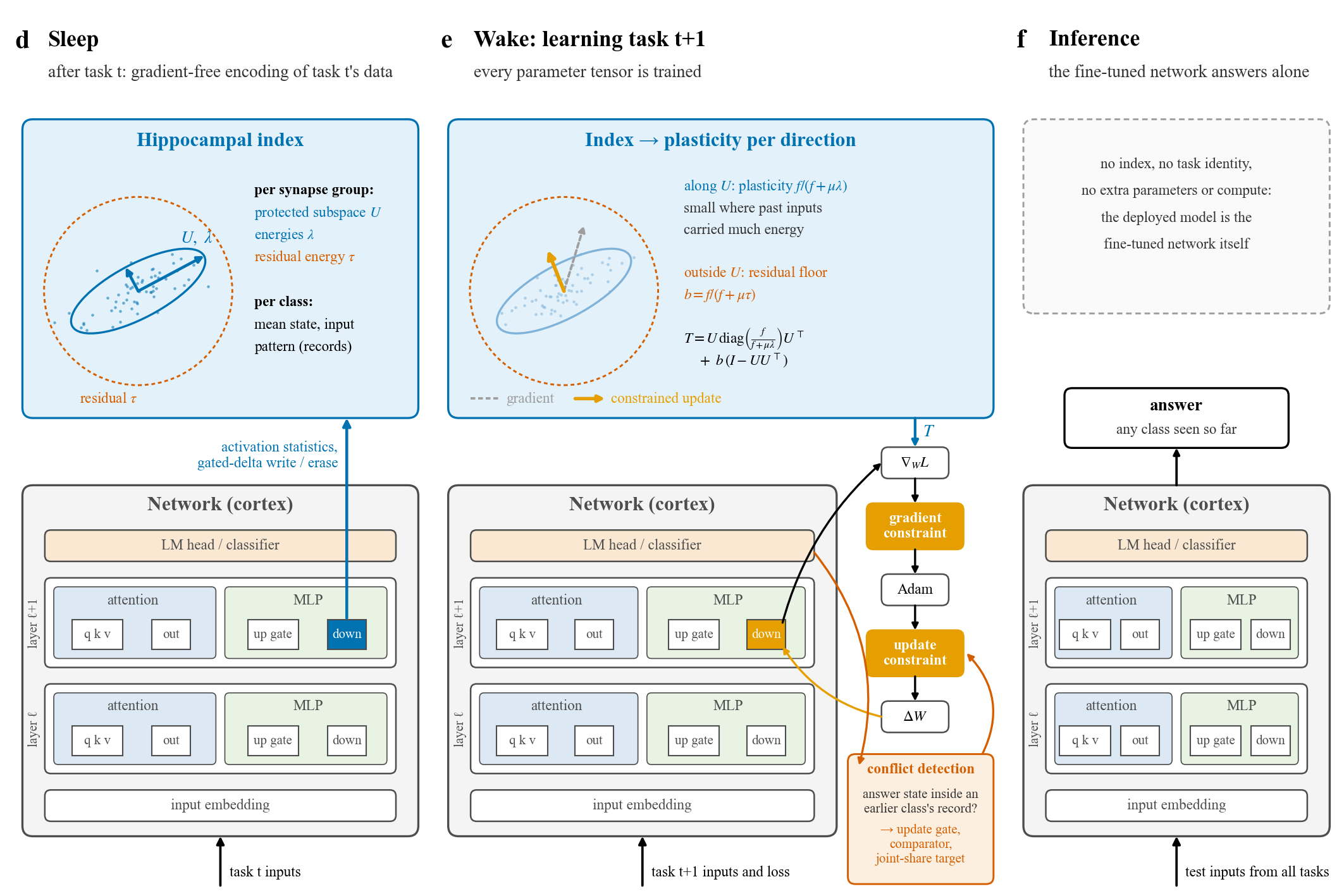}\par\vspace{2pt}
\caption{\textbf{Continual learning, catastrophic forgetting and DeltaHippo.} \textbf{a}, Class-incremental learning: one network trains on one task after another, each with new classes, and after every task is tested on all classes seen so far. \textbf{b}, Plain sequential fine-tuning of Llama-3.1-8B on eight TACRED tasks (seed 42; full accuracy matrices in Extended Data Fig. 1). \textbf{c}, The usual remedies: replay, distillation and freezing. \textbf{d–f}, DeltaHippo; the maps that read one input form a synapse group, and one group (MLP down) is followed. \textbf{d}, Sleep: after task $t$, the group's input geometry (protected subspace $U$, energies $\lambda$, residual energy $\tau$) and the class records are written into the hippocampal index. \textbf{e}, Wake: while task $t + 1$ is learnt, the index sets the operator $T$, which constrains the gradient before Adam and the update after it; conflict detection reads the answer state at the output head and, where it falls inside an earlier class's record, sets the update gate, comparator and joint-share target. \textbf{f}, Inference: the fine-tuned network alone.}
\end{figure*}

\section{Results}

\subsection{A training-only memory guides learning}

The model is the plain network, trained with cross-entropy and AdamW; every parameter tensor, including the output head, the input embedding matrix and the normalisation gains, stays in the optimization on every task. The memory changes only how each step is applied, per synapse group: the set of linear maps that read the same input tensor, through which a change alters earlier outputs (Methods: Synapse groups).

\textbf{Sleep.} At the end of each task, a gradient-free encoding phase re-reads that task's own training data (Fig. 1d). For each synapse group it eigendecomposes the second-moment matrix of the group's inputs, keeps the directions above a data-dependent cutoff\textsuperscript{28} and merges them into the memory by the gated delta rule of Gated DeltaNet\textsuperscript{29}. The memory holds a low-rank subspace $U$ with energies $\lambda$, the mean residual energy $\tau$ outside $U$ and the task's noise floor $f$, and, per class, the mean answer state and input pattern. No sentence, logit or answer token is kept, and earlier tasks are never re-read.

\textbf{Wake.} While a later task is learnt, every step passes through a closed-form operator derived from a local objective (Fig. 1e). Each remembered direction keeps a plasticity factor $f/(f + \mu\lambda_i)$, which falls with its stored energy and with the amount of earlier experience $\mu$; every direction outside $U$ keeps the residual floor $f/(f + \mu\tau)$; and the base update leaves earlier class means outside $U$ unchanged (class-pattern constraint). Because Adam's per-coordinate normalisation rotates a constrained gradient back into protected directions (Proposition 1, Methods), the operator is applied both to the gradient and to the update Adam takes.

\textbf{Conflict.} At every answer-token position the network's answer state is compared with the records of earlier classes; it is in conflict when it lies closer to an earlier class's mean than that class's nearest other class does. Only the state is read. A conflict sets three responses: an update gate lowers the probability that the step is applied, a comparator learns only the part of the input that no earlier class pattern explains, and a joint-share target keeps the new class from taking over a decision region it shares with an earlier near-duplicate (Methods).

At test time the plain model answers alone (Fig. 1f). A training step costs 1.25–1.33 times a plain fine-tuning step, inference costs nothing extra, and the added cost is one offline sleep per task (Methods: Cost).

\subsection{One principle across models and scales}

We evaluate the same principle on ten models and six class-incremental benchmarks: the decoder language models and the text encoders on TACRED, FewRel, Banking77 and CLINC150, and the vision transformer and the convolutional network on CIFAR-100 and ImageNet-R (Table 1; Extended Data Tables 1–2). Only the operators are fitted to each architecture's structure (Methods: Encoders and vision models).

\begin{table*}[!tp]
\caption{\textbf{Decoder LLMs.} Final average accuracy (\%) after the last task; mean ± sample standard deviation over three runs. SEQ, sequential full fine-tuning; SEQ* adapts the backbone on the first task and then freezes it.}
{\tabsize\setlength{\tabcolsep}{3pt}\begin{tabular*}{\textwidth}{@{\extracolsep{\fill}}lllllll@{}}
\toprule
\textbf{Model} & \textbf{Method} & \textbf{Relies on} & \textbf{TACRED} & \textbf{FewRel} & \textbf{Banking77} & \textbf{CLINC150}\\
\midrule
\textbf{Qwen3-0.6B} & SEQ & none (full FT) & 37.20 ± 3.31 & 24.46 ± 0.97 & 19.90 ± 0.25 & 39.00 ± 1.22\\
 & L2KD & teacher (distillation) & 10.66 ± 1.06 & 12.04 ± 0.95 & 17.52 ± 0.15 & 24.83 ± 0.45\\
 & LAMOL & generative replay + teacher & 33.42 ± 0.42 & 27.58 ± 1.44 & 29.32 ± 1.97 & 55.09 ± 0.92\\
 & PCLL & generative replay + teacher & 32.83 ± 1.25 & 24.85 ± 1.07 & 37.15 ± 2.68 & 56.01 ± 2.25\\
 & Sculpting (fixed budget) & weights' singular subspaces; full FT & 41.33 ± 0.48 & 40.38 ± 2.23 & 37.33 ± 1.18 & 63.91 ± 3.51\\
 & CODE-CL (task-agnostic, $K = 0$) & stored input subspaces (conceptors); full FT & 40.00 ± 4.39 & 34.78 ± 3.15 & 47.27 ± 3.67 & 70.59 ± 2.45\\
 & SEQ* & frozen backbone & 53.35 ± 1.07 & 69.21 ± 0.39 & 74.47 ± 1.71 & \textbf{87.79 ± 0.17}\\
 & \textbf{DeltaHippo} & \textbf{training-only geometric memory; full FT} & \textbf{72.46 ± 1.80} & \textbf{71.01 ± 1.40} & \textbf{82.91 ± 0.41} & 87.13 ± 2.96\\
\textbf{Llama-3.2-1B} & SEQ & none (full FT) & 35.09 ± 2.83 & 38.87 ± 1.82 & 24.90 ± 1.04 & 58.24 ± 0.89\\
 & L2KD & teacher (distillation) & 21.04 ± 0.50 & 32.51 ± 2.78 & 20.91 ± 0.52 & 58.40 ± 0.53\\
 & LAMOL & generative replay + teacher & 52.63 ± 2.73 & 42.29 ± 3.03 & 64.29 ± 0.85 & 72.86 ± 1.42\\
 & PCLL & generative replay + teacher & 35.64 ± 4.09 & 43.45 ± 1.65 & 43.64 ± 2.39 & 64.33 ± 1.56\\
 & Sculpting (fixed budget) & weights' singular subspaces; full FT & 26.54 ± 7.56 & 39.72 ± 4.01 & 35.63 ± 3.61 & 57.67 ± 4.55\\
 & CODE-CL (task-agnostic, $K = 0$) & stored input subspaces (conceptors); full FT & 62.10 ± 12.36 & 58.04 ± 2.29 & 56.24 ± 4.96 & 82.87 ± 2.35\\
 & SEQ* & frozen backbone & 54.95 ± 1.41 & 72.56 ± 0.35 & 80.28 ± 0.57 & \textbf{90.41 ± 0.05}\\
 & \textbf{DeltaHippo} & \textbf{training-only geometric memory; full FT} & \textbf{71.58 ± 0.49} & \textbf{79.09 ± 2.36} & \textbf{82.28 ± 1.13} & 87.04 ± 1.46\\
\textbf{Qwen3-4B} & SEQ & none (full FT) & 43.62 ± 0.99 & 39.68 ± 0.97 & 32.13 ± 1.36 & 61.97 ± 2.02\\
 & SEQ* & frozen backbone & 57.35 ± 0.56 & \textbf{77.58 ± 0.27} & 78.48 ± 1.13 & 90.58 ± 0.08\\
 & \textbf{DeltaHippo} & \textbf{training-only geometric memory; full FT} & \textbf{76.52 ± 1.46} & 76.52 ± 4.48 & \textbf{87.46 ± 0.67} & \textbf{92.33 ± 0.19}\\
\textbf{Llama-3.2-3B} & SEQ & none (full FT) & 41.43 ± 2.28 & 42.03 ± 0.97 & 29.20 ± 0.81 & 59.94 ± 1.30\\
 & SEQ* & frozen backbone & 53.99 ± 1.35 & 75.85 ± 0.66 & 81.11 ± 0.49 & \textbf{91.93 ± 0.22}\\
 & \textbf{DeltaHippo} & \textbf{training-only geometric memory; full FT} & \textbf{76.40 ± 1.25} & \textbf{81.48 ± 1.60} & \textbf{85.84 ± 0.47} & 91.68 ± 0.83\\
\textbf{Qwen3-8B} & SEQ & none (full FT) & 43.32 ± 2.34 & 39.63 ± 1.83 & 33.29 ± 1.71 & 68.55 ± 2.66\\
 & SEQ* & frozen backbone & 59.32 ± 0.59 & 77.94 ± 0.33 & 77.37 ± 1.65 & 92.29 ± 0.28\\
 & \textbf{DeltaHippo} & \textbf{training-only geometric memory; full FT} & \textbf{76.79 ± 2.14} & \textbf{85.95 ± 0.79} & \textbf{88.36 ± 0.29} & \textbf{93.70 ± 0.77}\\
\textbf{Llama-3.1-8B} & SEQ & none (full FT) & 37.61 ± 3.82 & 47.51 ± 0.94 & 29.77 ± 0.73 & 65.16 ± 2.93\\
 & SEQ* & frozen backbone & 66.94 ± 2.52 & 80.48 ± 0.82 & 84.59 ± 0.19 & \textbf{93.08 ± 0.64}\\
 & \textbf{DeltaHippo} & \textbf{training-only geometric memory; full FT} & \textbf{77.35 ± 0.69} & \textbf{82.92 ± 1.71} & \textbf{87.80 ± 0.54} & 92.53 ± 0.69\\
\bottomrule
\end{tabular*}}
\par\vspace{2pt}{\fontsize{7bp}{8.5bp}\selectfont Bold, best in the model's column. Every model is evaluated after the last task, with no checkpoint selection; the three runs use seeds 42, 11 and 3, batch 32 and 3 epochs per task; every run is listed in Supplementary Table 13.\par}
\end{table*}

DeltaHippo exceeds sequential fine-tuning on every cell, by 25–63 points on the language models, 39–77 on the encoders and 41–67 on the vision models. It exceeds the subspace-constrained full fine-tuning methods Sculpting Subspaces\textsuperscript{30} and CODE-CL\textsuperscript{31} on every cell, by 4.2–46.7 points, generative replay with a teacher (LAMOL, PCLL) and teacher distillation (L2KD) on every cell where they are evaluated, and the replay buffers DER++ and CLS-ER on every encoder and vision cell. Against the frozen-backbone baseline SEQ*, which no longer has representational interference to manage, DeltaHippo ends higher on 19 of the 24 language-model cells, on every encoder cell (by 18–33 points) and on three of the four vision cells; the remaining cells are those where the frozen representation already separates the classes. The margin depends on the task: averaged over the six decoders it is +17.5 points on TACRED, where many relations have near-duplicates learnt in other tasks, against +6.4 on Banking77 and +3.9 on FewRel, with the two level on CLINC150, and on TACRED it holds under two further class orders (Supplementary Note 2; Supplementary Table 17).

The backbone does not simply stop changing. Sequential fine-tuning acquires every task but keeps 0–38\% of the earlier ones on the language models, whereas DeltaHippo retains 63–95\% throughout while acquiring every task (Fig. 2a). Within one system, continued backbone learning pays: on ViT-B/16 ImageNet-R, the same method with the backbone frozen after task 0 ends 10.9 points lower (Fig. 2b; Extended Data Table 3). The gain is present in every family and scale (Fig. 2c–e).

\begin{figure*}[!p]
\centering
\includegraphics[width=0.827\textwidth]{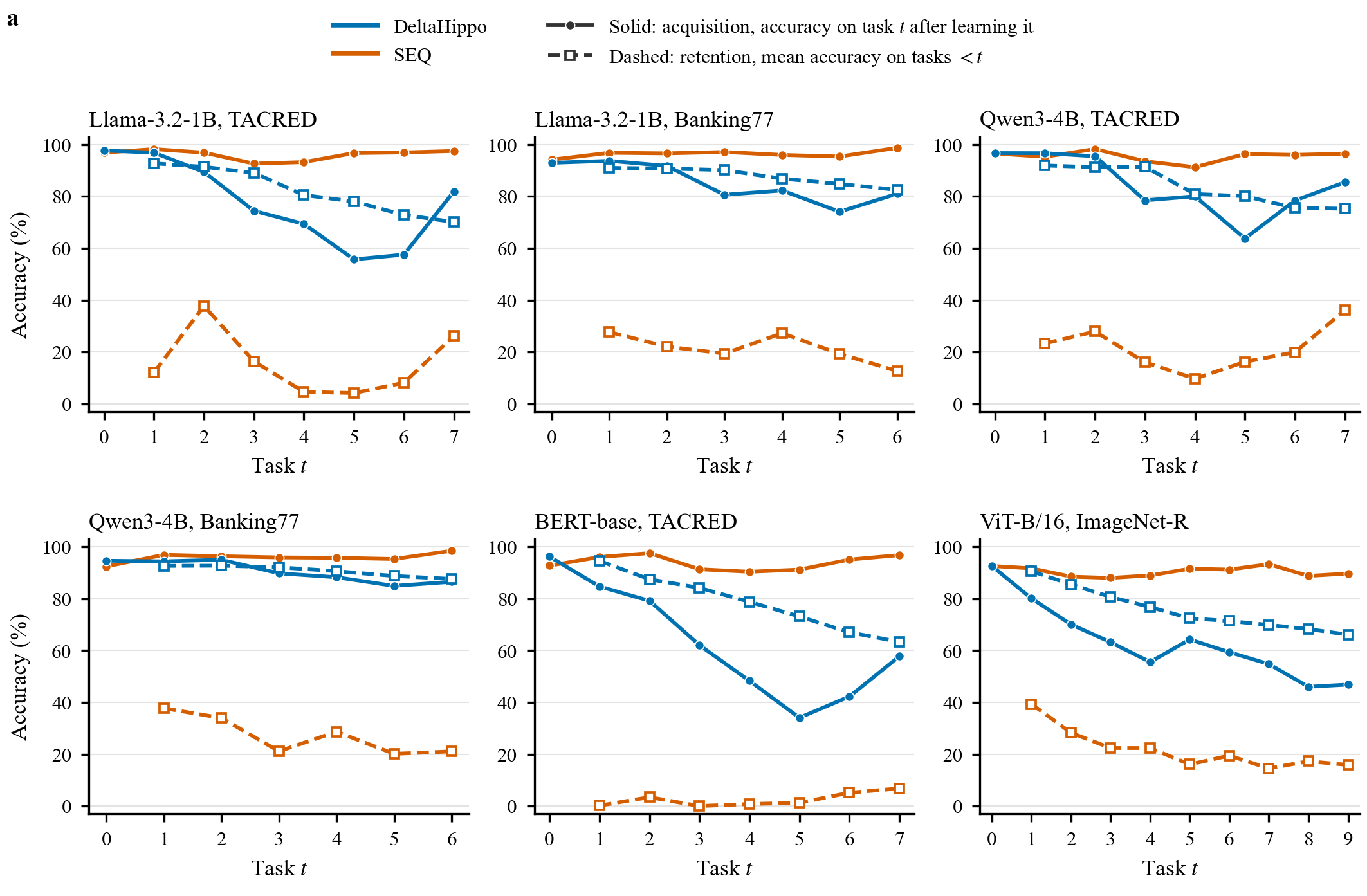}\par\vspace{2pt}
\includegraphics[width=0.827\textwidth]{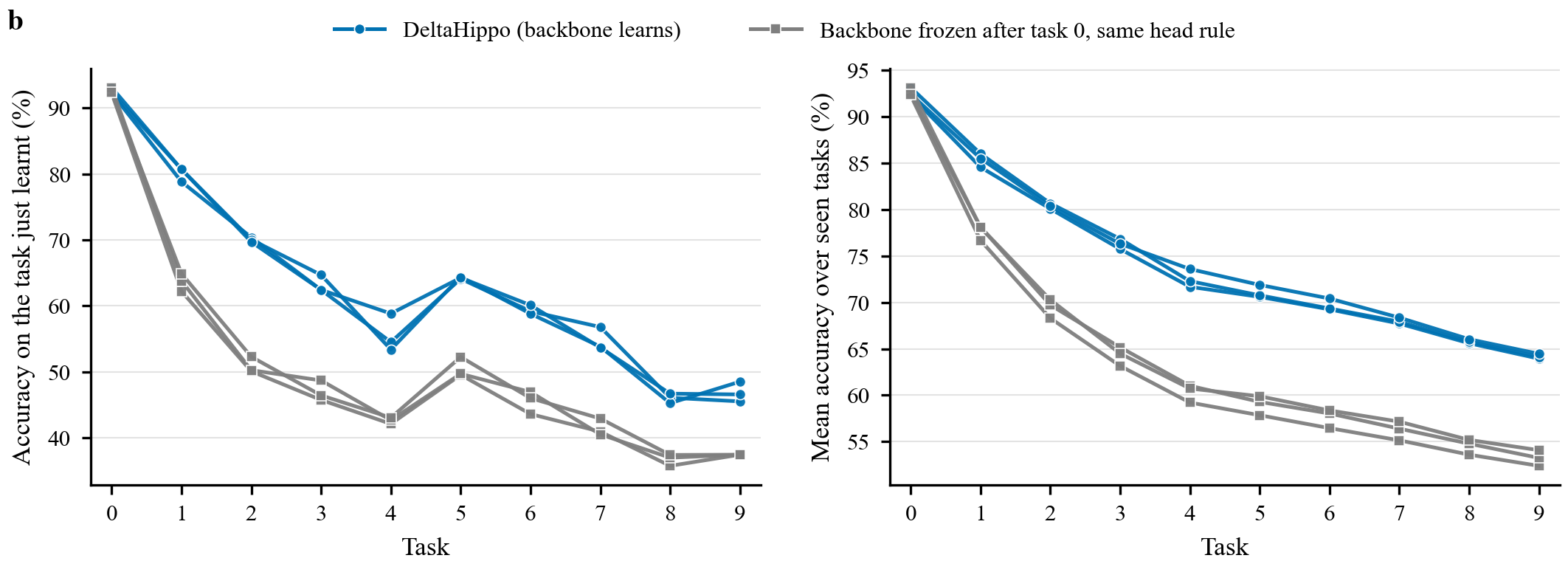}\par\vspace{2pt}
\includegraphics[width=0.827\textwidth]{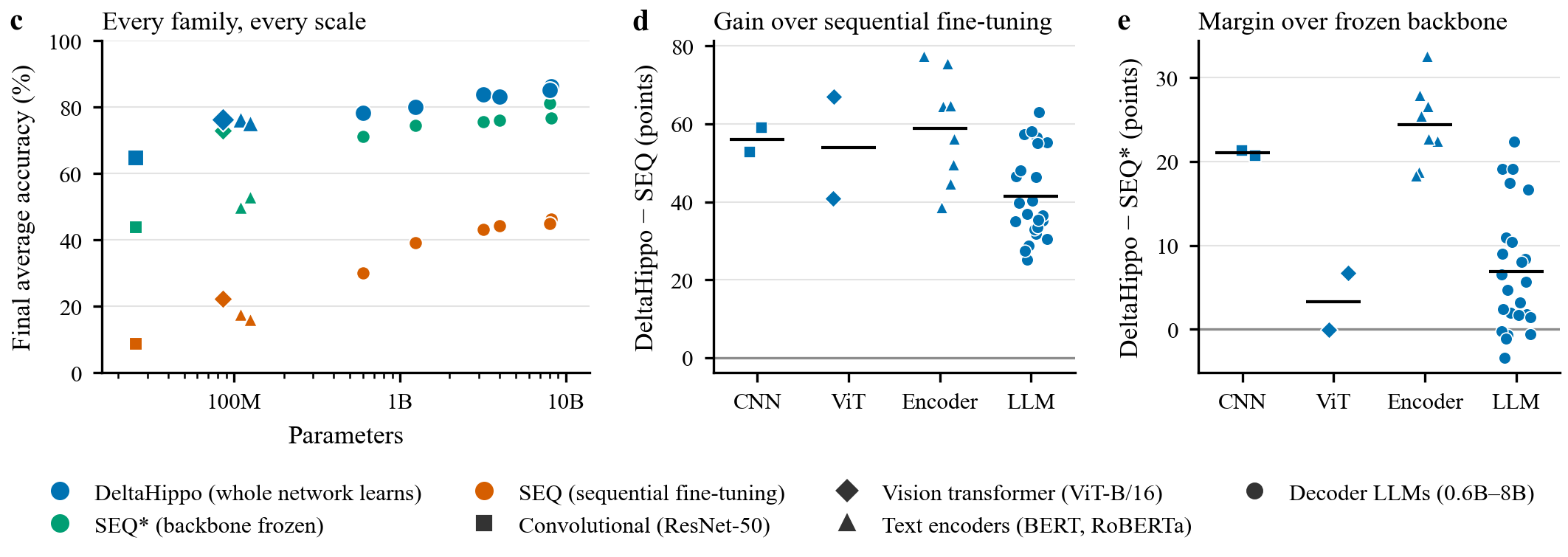}\par\vspace{2pt}
\caption{\textbf{One principle across families and scales.} \textbf{a}, Acquisition (accuracy on each task right after learning it) and retention (mean accuracy on the earlier tasks) after each task, SEQ and DeltaHippo: Llama-3.2-1B on TACRED and Banking77, Qwen3-4B on TACRED and Banking77, BERT-base on TACRED and ViT-B/16 on ImageNet-R; mean over three runs. \textbf{b}, Same-head control (ViT-B/16, ImageNet-R): DeltaHippo and the same system with the backbone frozen after task 0. \textbf{c}, Final average accuracy (mean over each model's benchmarks) against parameter count; marker shape, family. \textbf{d}, Gain over SEQ, one point per setting (36). \textbf{e}, Margin over SEQ\textbackslash{}*; bars, family means. Values from Table 1 and Extended Data Tables 1–2.}
\end{figure*}

\subsection{Memory content, not step size, matters}

To ask why the method works, we intervene on the memory at a task transition (Llama-3.2-1B, TACRED). We train the normal method through task 3, capture the full training state, and learn task 4 under six conditions that share the same minibatches, learning rate and evaluation; every condition is identical to the normal run up to the point where it is changed (Fig. 3a–d; accuracies at the end of task 4 over three runs, Fig. 3e).

Without the read, every earlier task collapses within the first quarter of task 4 (B). Rotating the stored directions while keeping every eigenvalue and plasticity factor collapses them as well (C): the directions the hippocampus protects must match what the network experienced. Moving each synapse group by exactly the method's amount, at the same steps, in the plain direction also collapses them (F), so the benefit is not a smaller effective step; the memory organises the direction of change. Reconnecting the read does not restore access already lost (D): the memory holds no copy of the knowledge, which lives in the network, and regulates how much the next experience damages it. Each sleep selectively protects the task it encoded: blocking only task 3's sleep costs task 3 32.0 points and the older tasks 2.5 (E).

\begin{figure*}[!tp]
\centering
\includegraphics[width=1.000\textwidth]{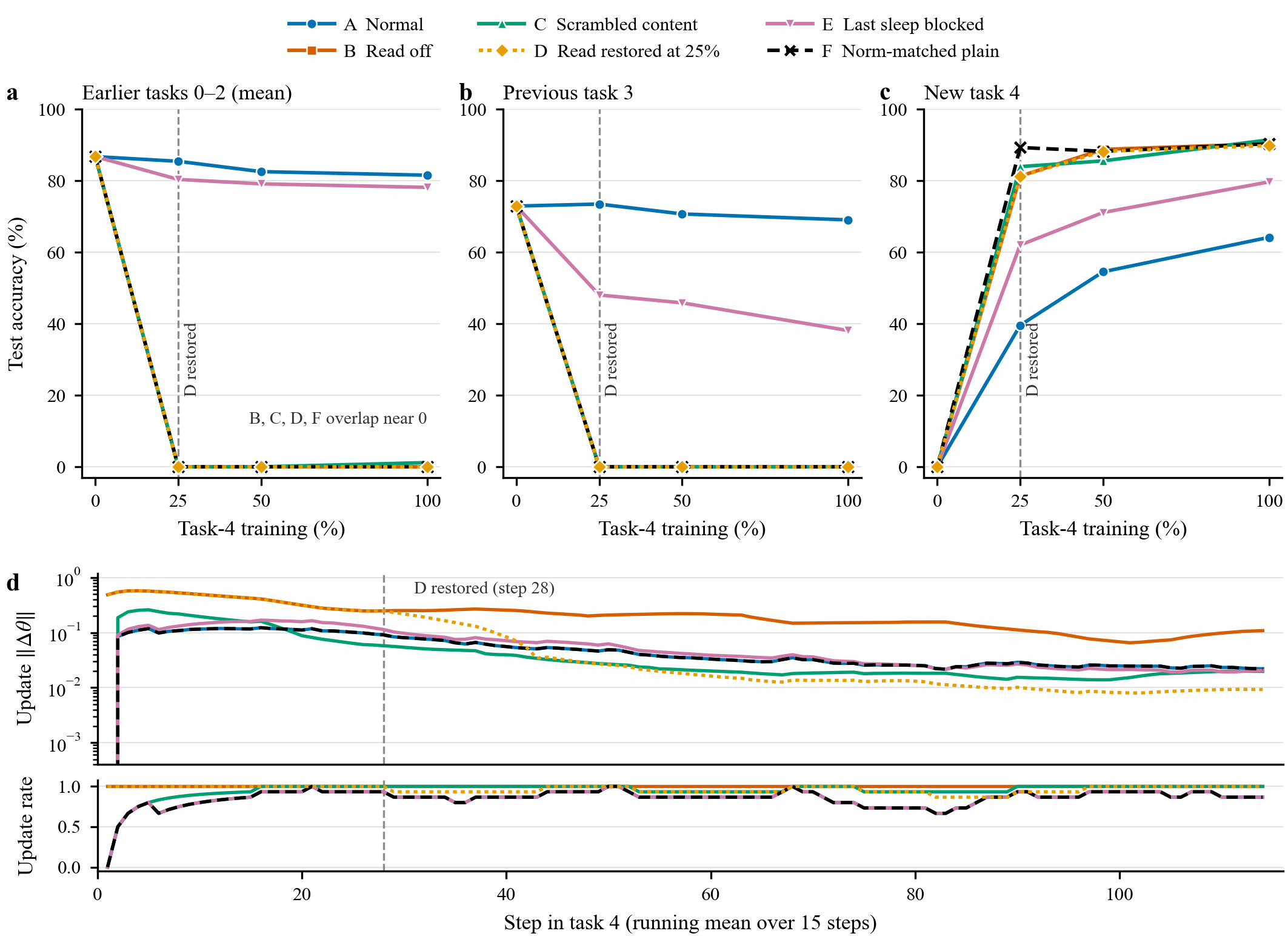}\par\vspace{2pt}
{\raggedright\sffamily\bfseries\small e\par}\vspace{2pt}
{\tabsize\setlength{\tabcolsep}{3pt}\begin{tabularx}{1.000\textwidth}{@{}>{\raggedright\arraybackslash}X>{\raggedright\arraybackslash}Xlll@{}}
\toprule
\textbf{Condition} & \textbf{What changes} & \textbf{Old tasks 0–2} & \textbf{Task 3} & \textbf{New task 4}\\
\midrule
A. Normal & nothing & \textbf{83.8 ± 3.2} & \textbf{70.7 ± 3.4} & 69.3 ± 4.9\\
B. Read off & the hippocampus is not read; plain update & 0.0 ± 0.0 & 0.0 ± 0.0 & 91.4 ± 1.4\\
C. Scrambled content & same rank and eigenvalues, directions rotated inside each synapse group & 4.6 ± 5.5 & 0.0 ± 0.0 & 91.1 ± 0.6\\
D. Read restored & as B, then the original read is reconnected at 25\% of task 4 & 0.0 ± 0.0 & 0.0 ± 0.0 & 91.6 ± 1.6\\
E. Last sleep blocked & task 3's sleep is not written; the earlier memory is read & 81.3 ± 3.1 & 38.7 ± 5.8 & 79.7 ± 1.1\\
F. Norm-matched plain & plain direction, with A's per-synapse-group applied norm and A's applied/skipped steps & 2.0 ± 3.4 & 0.0 ± 0.0 & 91.3 ± 1.5\\
\bottomrule
\end{tabularx}}
\caption{\textbf{Memory intervention at the transition to task 4 (Llama-3.2-1B, TACRED).} \textbf{a–c}, The three task groups at 0, 25, 50 and 100\% of task 4 (one run); dashed line, D's restore. \textbf{d}, Applied update norm and update rate per step (same run). \textbf{e}, Accuracy (\%) at the end of task 4 under each condition, mean ± sample standard deviation over three independent runs. Runs 2 and 3 of a–d, Extended Data Fig. 2.}
\end{figure*}

\subsection{A class-free core gives most of the gain}

Removing every class-conditioned part (the class records, class-pattern constraint, conflict detection with the update gate, comparator, joint-share target and erase) leaves the residual-aware index and its two constraints (Methods, Box 1). Over three runs this class-free core reaches 67.1 ± 2.2 against 71.6 ± 0.5 for the full method: of the full method's 36.5-point gain over sequential fine-tuning, the core delivers 32.0 (Extended Data Table 4). Nothing in the core reads a class label; what protects earlier knowledge is the geometry of past inputs itself. The same holds where the output is a classifier rather than generated text: on ViT-B/16 ImageNet-R the core ends at 63.4 against 64.0 for the full method in the matched run (Supplementary Note 4).

Two design conditions carry the core, each confirmed by removing it from the full method in three runs. First, the constraint must act on both sides of the adaptive optimizer. Constraining the gradient but applying Adam's update as it comes lowers retention from 82.1 to 60.3 and final accuracy by 9.4 points, while acquisition rises from 77.9 to 87.3; at the first transition the applied update then carries 3.0–80 times the interference energy under the stored metric of earlier inputs. Constraining only the realised update lowers retention to 53.1 and final accuracy by 15.2 points. The gradient constraint shapes the moments Adam accumulates; the update constraint re-imposes it on the update those moments produce. Second, the complement of what the memory stores must not be treated as free. Opening the residual subspace lowers retention from 82.1 to 79.2 as acquisition rises, whereas making the protected subspace hard while keeping the residual floor leaves retention intact (82.9). The less the operator protects explicitly, the more the floor carries: when it protects only the top quarter of the stored directions, the residual floor is worth 18.9 points of final accuracy and 26.6 of retention. The same conditions hold in BERT-base and ResNet-50, and removing the update constraint costs ViT-B/16 17.0 points (Extended Data Table 5). Erase lowers the stored energy (54 against 96 without it) while the protected subspace stays well below full on these sequences, and is designed for sequences that saturate the memory (Supplementary Note 4).

\subsection{Conflict detection keeps near-duplicates}

Average accuracy dilutes the loss of one class among forty, so we measure absorption directly. We count two relations as near-duplicates when their labels, after the person/organisation prefix, differ in a single word (a person's \emph{city of birth} and \emph{city of death}) and are learnt in different tasks; the set is fixed by this rule on the label strings alone (Supplementary Note 2). On TACRED, 15 of the 40 relations have a near-duplicate, and eight of them, the earlier classes, are learnt before one of their near-duplicates. Every method first learns these classes (72–96\% right after their own task). At the end of the sequence (Llama-3.2-1B, three runs of each Table 1 configuration), every compared method answers fewer of their test sentences correctly than DeltaHippo (0.9–63.0\% against 76.6\%), absorbs 3 to 14 times as many into their later near-duplicates (9.0–40.5\% against 2.8\%, 6.2–37.7 percentage points more), and, when it errs on these relations, names a near-duplicate up to 2.8 times as often (Fig. 4a–c; Extended Data Table 6). The same holds at every scale: on all six decoders DeltaHippo answers more of the earlier classes correctly than SEQ* (73.5–81.2\% against 51.1–64.5\%; Extended Data Table 7). Whatever a method keeps, earlier outputs, a teacher, subspaces of the weights or inputs, or the whole backbone, it does not register when a later class takes an earlier class's region.

Conflict detection supplies what the index lacks: it flags the answer-token positions of the present task whose state falls inside an earlier class's record, exactly where absorption begins. Without the class-conditioned parts, the task-2 class \emph{state or provinces of residence} falls to 0 of 41 test sentences in every run once its near-duplicate \emph{state or province of birth} is learnt, whereas the full method keeps 29 on average (up to 39). The two labels form a near-duplicate pair when singular and plural count as the same word, and under that rule DeltaHippo again leads on every measure of Extended Data Table 6 (Supplementary Note 2). With both classes weighted equally, the full method keeps the pair better in every run (mean recall 59, 70 and 48\% against 44, 39 and 39\%; the lower of the two recalls 29, 44 and 11\% against 0\%; Fig. 4d–f). The comparator is what lets colliding classes be learnt: switching it off lowers final accuracy by 12.9 points over three runs (71.6 ± 0.5 to 58.7 ± 0.6), as \emph{city of death} is answered as \emph{city of residence} 21 times instead of at most 3, and placing it where conflict detection fires raises acquisition by 3.5 points over the same number of other positions. The joint-share target keeps the established class's share of a decision region it shares with a newcomer. Because the whole network keeps adapting while conflict detection holds these boundaries, the earlier distinction stays usable and the later class is still acquired.

\begin{figure*}[!p]
\centering
\includegraphics[width=1.000\textwidth]{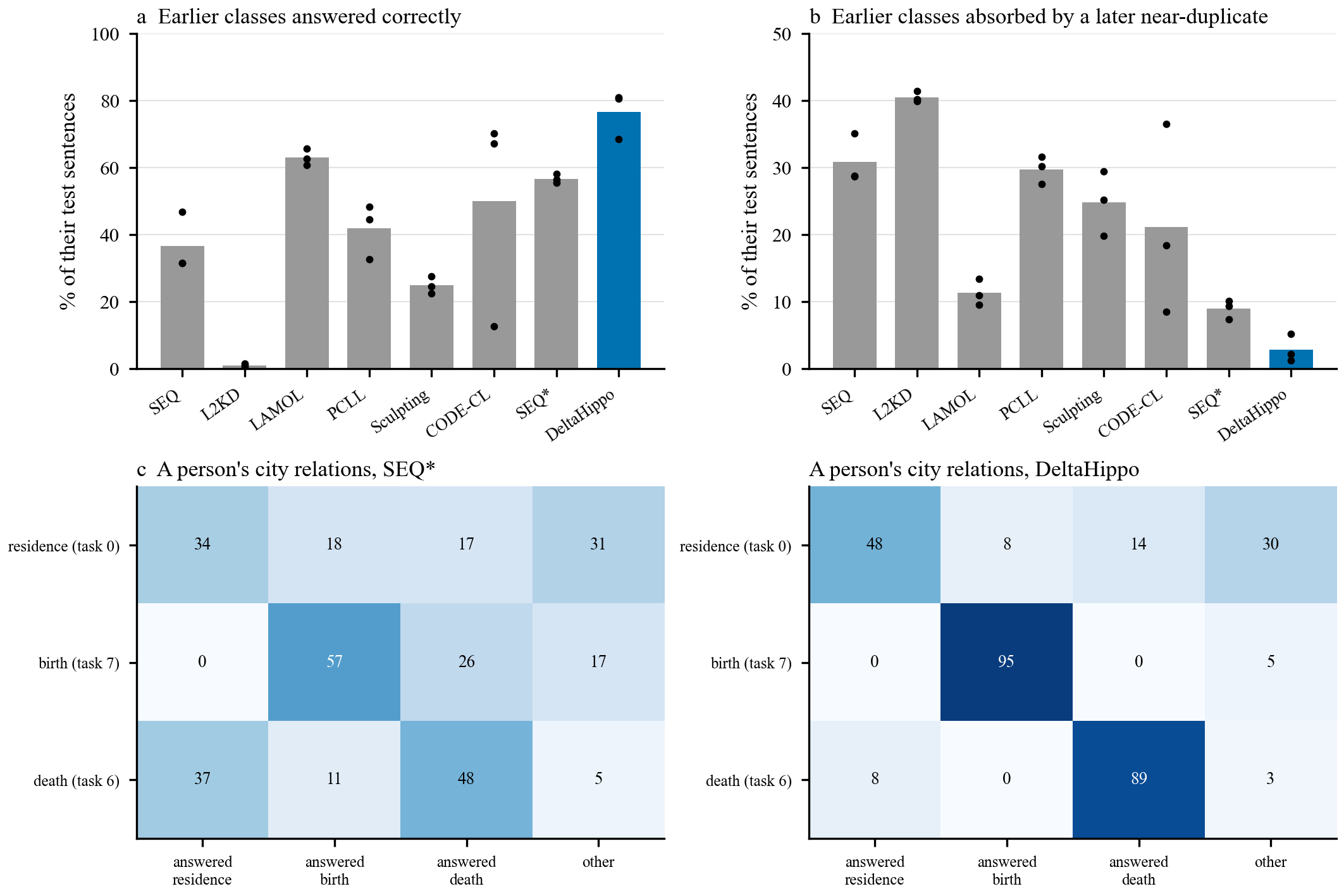}\par\vspace{2pt}
\includegraphics[width=1.000\textwidth]{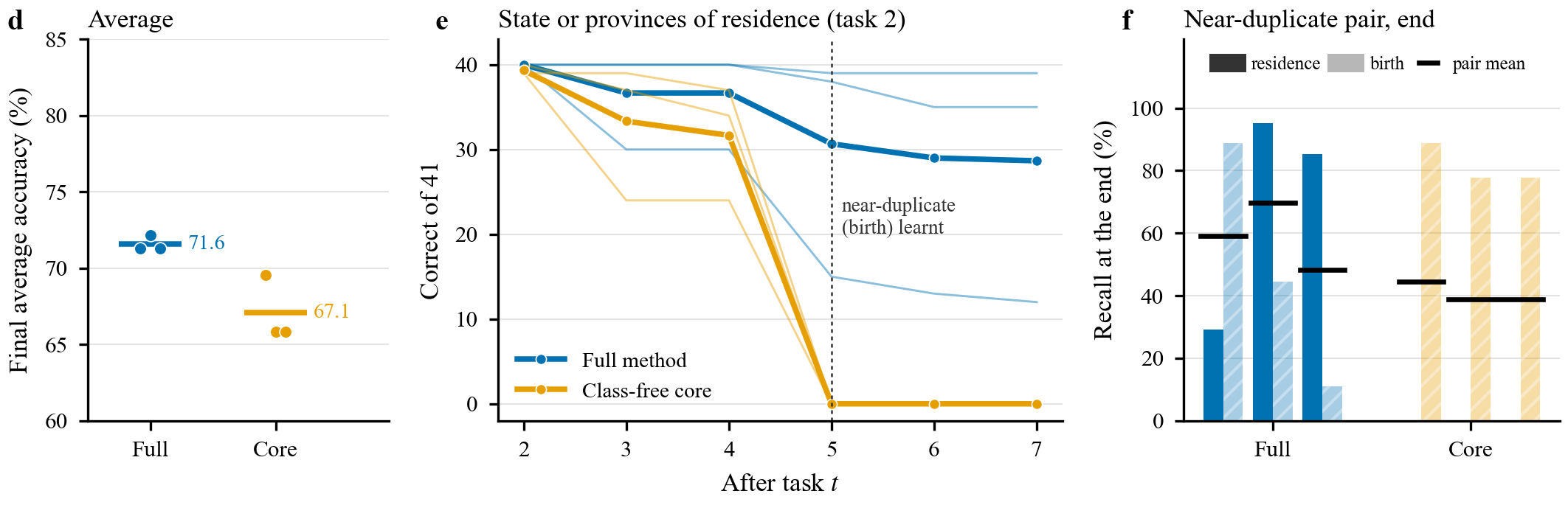}\par\vspace{2pt}
\caption{\textbf{Near-duplicate classes on TACRED (Llama-3.2-1B; three runs).} \textbf{a}, Share of the eight earlier classes' test sentences answered correctly at the end of the task sequence, every method. \textbf{b}, Share absorbed, answered as one of the class's later near-duplicates. Bars, mean over runs; dots, runs. \textbf{c}, A person's three city relations under SEQ\textbackslash{}* and DeltaHippo: share of each relation's test sentences (rows; task in brackets) answered as each of the three or as anything else. \textbf{d}, Final average accuracy of the class-free core and the full method. \textbf{e}, Correct test sentences (of 41) of the task-2 class ‘state or provinces of residence’ after each task; its near-duplicate ‘state or province of birth’ is learnt at task 5. \textbf{f}, The pair at the end, per run: recall of residence (solid) and birth (hatched) and their mean with both classes weighted equally (black).}
\end{figure*}

\subsection{Earlier knowledge stays within reach}

What the geometric core and conflict detection protect, in the end, is the compatibility between the changing representation and the output that answers with it. In the task-4 intervention, a probe fitted on each condition's own final-layer states still separates the earlier classes nearly as well as in the normal run, although generation on them falls by 80–82 points; the normal run's probe applied unchanged does not separate them (Supplementary Table 18). The collapse is mainly a loss of access, not of linear separability.

Access needs both sides. Guiding only the backbone, or only the output head (in the tied model, the input embedding rows and final norm), over the whole sequence ends at 53.6 ± 11.0 and 12.1 ± 0.2 against 71.6 ± 0.5, although at task 4 a probe on each run's own states still reads the earlier tasks at 82.4 and 85.5, against 83.3 for the whole model (Fig. 5a). Swapping output heads between runs locates the break (Fig. 5b): a guided head reads the whole-model states at 80.7 while the unguided head of the backbone-only run reads them at 43.4, and the whole-model head recovers the backbone-only states from 23.8 to 61.8 but not to 80.4. Neither side's guidance keeps its part intact alone: the two pathways learn interdependently, which is why the hippocampus indexes every synapse group, output head included. In the untied Qwen3-8B, joint guidance again keeps the earlier classes most usable (Supplementary Note 4). The lost access is recovered by a low-rank correction of the states fitted on the same classes: a rank-1 map closes 95\% of the backbone-only run's gap to a re-fitted probe, against at most half for a bias and temperature (Supplementary Tables 19–20). Over the full sequence, the earlier tasks remain as linearly separable as in a backbone frozen right after task 0 (77.2 against 75.4; Fig. 5c).

\begin{figure*}[!tp]
\centering
\includegraphics[width=0.490\textwidth]{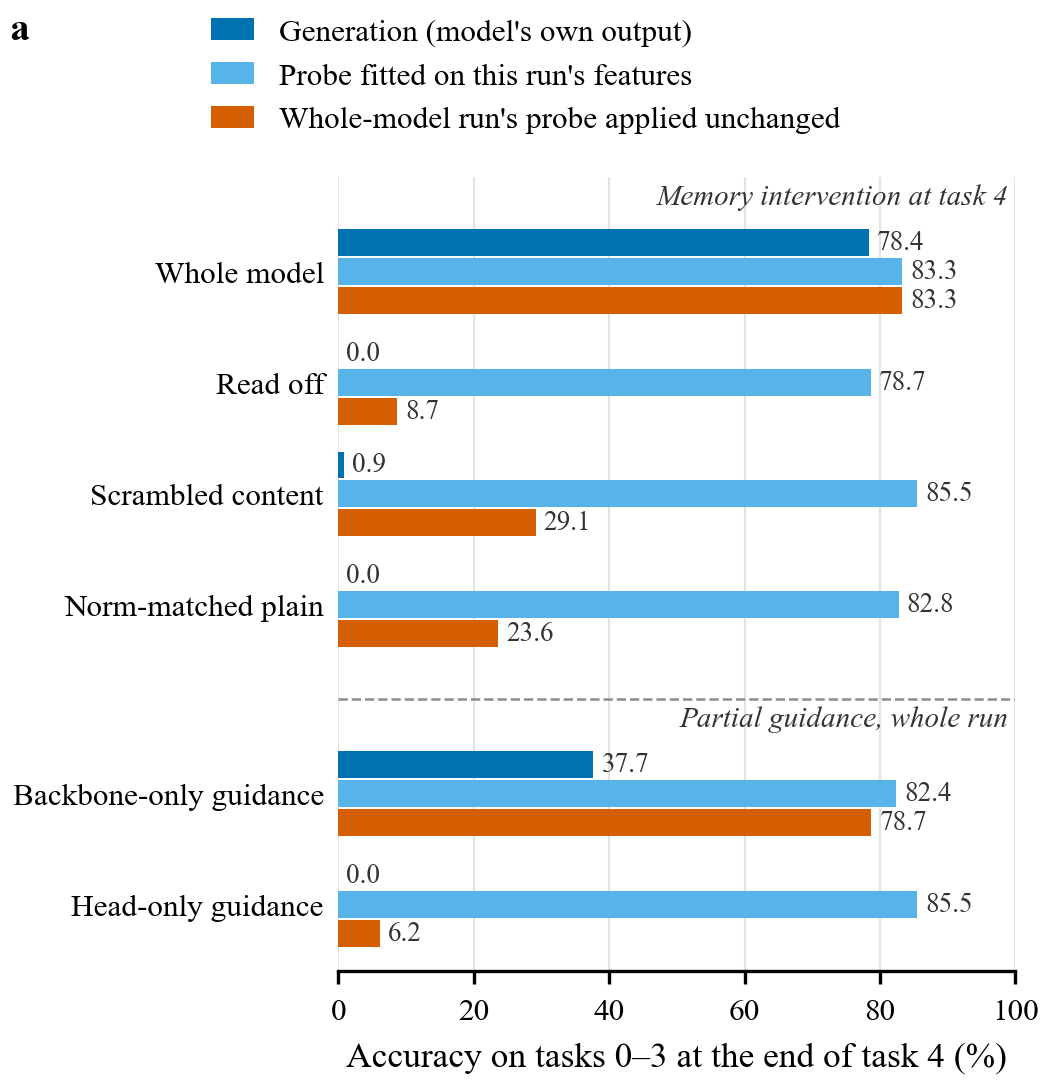}\hfill\includegraphics[width=0.490\textwidth]{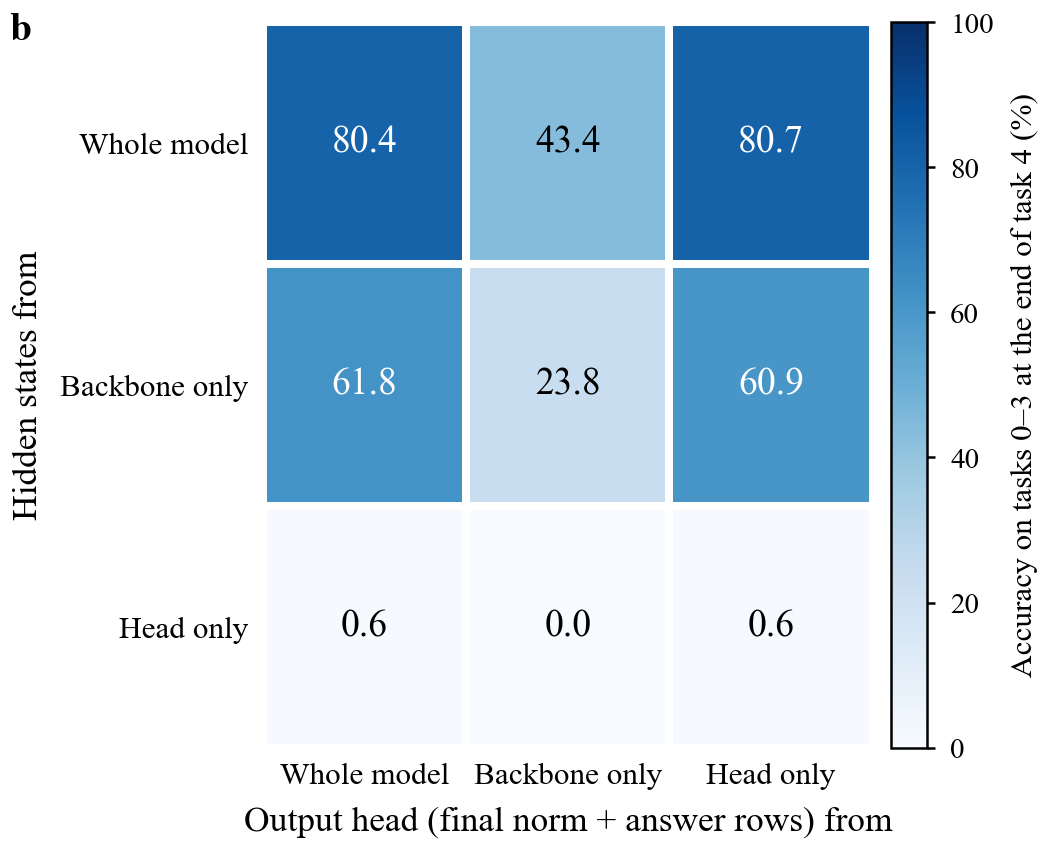}\par\vspace{2pt}
\includegraphics[width=1.000\textwidth]{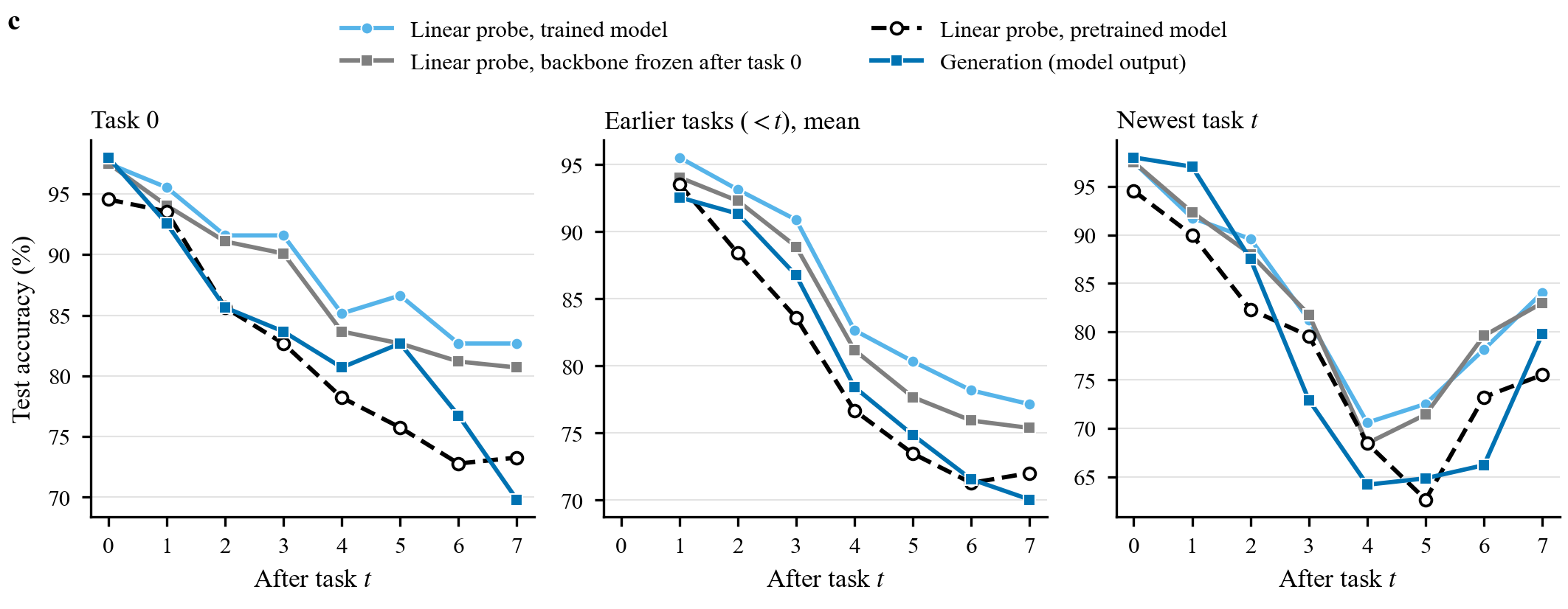}\par\vspace{2pt}
\caption{\textbf{Separable versus usable (Llama-3.2-1B, TACRED).} \textbf{a}, Tasks 0–3 at the end of task 4 (one run): the model's own generation, a probe fitted on each run's features (an analysis readout trained on the earlier tasks' training data, not part of the method) and the whole-model run's probe applied unchanged. \textbf{b}, Teacher-forced accuracy of the fixed hidden states of each guidance run (rows) read by each run's output head (columns): tasks 0–3 at the end of task 4, label-level, mean over three runs. \textbf{c}, A common linear probe on the final-layer state before the first answer token: trained model, backbone frozen after task 0 and pre-trained model; runs 2 and 3, Extended Data Fig. 3.}
\end{figure*}

\section{Discussion}

A network whose every parameter tensor keeps learning can accumulate classification skills without cross-task replay, a teacher or an inference-time memory, when a training-only memory organises how it changes. The evidence forms one chain: what the memory holds, not how much it slows learning, decides what is kept (Fig. 3); its class-free core works when it constrains both sides of the adaptive optimizer and protects the residual energy that compression discards; conflict detection keeps the earlier member of a near-duplicate pair (Fig. 4); and what all of this preserves is the compatibility between a changing representation and the output that uses it (Fig. 5). The same principle, with operators fitted to each architecture, holds from convolutional networks to 8-billion-parameter language models.

\textbf{A memory that organises future learning.} Continual learning has mostly used memory to keep material for relearning the past. The memory here supplies only geometric constraints on future learning, and its effect depends on what it holds, not on how much it shrinks the step. Past content need not be replayed if its geometry keeps organising future change. Projection methods introduced the protection of earlier input directions; the step taken here is to tie an index of experience to the optimizer that actually trains a whole pre-trained network, on both sides of it and in every synapse group, output head included, and to let it change how the network learns where new classes collide with old ones. Any memory that fits beside a large model compresses, and the directions it does not store are not unused: earlier inputs left aggregate energy there. The complement of a compressed memory is not a null space of past experience. Because the memory is defined on what every architecture is made of, linear maps reading shared inputs, the class-free core is a candidate principle beyond class-incremental classification.

\textbf{Where the hippocampus matters most.} The advantage is largest where continual learning is hardest, when later tasks bring near-duplicates of classes already learnt. There every compared method loses the earlier class, whether it keeps earlier outputs, a teacher, subspaces of the weights or inputs, or the whole backbone. What fails is not a stored quantity but the boundary between two classes, which only the later task puts under pressure: absorption is retroactive interference of the kind that grows with the similarity of the competing memories\textsuperscript{32}. It has to be found during learning, from the network's own state, and held by learning differently there, while the index keeps the rest of the network free to adapt. Methods that freeze the backbone hold up wherever the representation already separates the classes to come, and converge with DeltaHippo there; telling near-duplicates apart asks the whole network to keep adapting, which a fixed network does not do.

\textbf{Separable is not usable.} Without guidance, information that a newly fitted probe can recover largely survives while the model's own access to it is lost. This is the distinction between the availability and the accessibility of a memory\textsuperscript{33}, here measured inside a network. Usable knowledge behaves as a relation between representation and output, best kept when both are guided while they learn. Earlier work responded to the gap between decodable and usable knowledge by freezing part of the model\textsuperscript{5,34}, and recent work analyses how access is lost and recovered after fine-tuning\textsuperscript{35}. Here the loss of access is prevented as learning proceeds, while every parameter tensor learns.

\textbf{A computational hypothesis.} We use hippocampal accounts as functional design principles, not as a biological model (Methods: Correspondence with hippocampal accounts). Sleep here builds an index that protects knowledge rather than transferring it. Within this functional reading, the model bears out the predictions of indexing, regulation and conflict detection, and its erase is designed for task sequences that press on the memory. Supplementary Note 1 relates DeltaHippo to input-space projection, subspace fine-tuning, frozen-backbone and hippocampus-inspired methods.

{\fontsize{8.5bp}{10.5bp}\selectfont
\section{Methods}

\textbf{Overview.} The network is the cortex: it holds all long-term knowledge and is fully fine-tuned on every task (Model training). The hippocampus is a gated-delta memory in the style of Gated DeltaNet over the network's synapse groups (Synapse groups; What the memory stores). It alternates between two phases:

\begin{itemize}[leftmargin=1.4em,itemsep=1pt,topsep=2pt,parsep=0pt]
\item[\textbullet] \textbf{sleep}, at the end of each task except the last, when it encodes the finished episode by gated-delta write and erase (Sleep);
\item[\textbullet] \textbf{wake}, during the next tasks, when it is read to build the operators that shape every cortical update (Operators), to detect conflicts with earlier records (Conflict detection), and to gate and correct the steps at those conflicts (Learning at detected conflicts; The constrained update).
\end{itemize}

Consolidation happens without cross-task replay: the index governs cortical plasticity, and the model answers alone (One training step and one task). The sections below first describe DeltaHippo for generative LLMs and then its adaptations to encoders and vision models.

\textbf{Layers of the method.} Supplementary Table 3 summarises the components of DeltaHippo.

\begin{natbox}{The class-free core}
Everything the core needs, per synapse group a (the linear maps reading one input tensor; Synapse groups). It reads no class label. The scaling rules (relative step, width rule, embedding-row and gain plasticity factors) apply as in the full method.
\par\vspace{3pt}\emph{After task $e$} (sleep; a gradient-free encoding phase over task $e$'s own training data):
\begin{itemize}[leftmargin=1.4em,itemsep=1pt,topsep=2pt,parsep=0pt]
\item[\textbullet] $\hat C_e$ = position-averaged $x x^\top$, answer-token positions reweighted as in Sleep (LM head: answer-token states);
\item[\textbullet] $(K, \lambda_K)$ = eigenpairs of $\hat C_e$ with $\lambda > f$, $f = \theta^2$ ($\theta$: Gavish–Donoho cutoff on $\sigma = \sqrt{\lambda}$; $f$: the episode's noise floor);
\item[\textbullet] $(U_a, \lambda_a) \leftarrow$ write $(K, \lambda_K)$ by the gated delta rule, erase off;
\item[\textbullet] $\mathrm{tail}_a \leftarrow$ running mean of the energy outside $U_a$; $n_a \leftarrow n_a + n_e$; $n_a^{\mathrm{last}} \leftarrow n_e$.
\end{itemize}
\par\vspace{3pt}\emph{At the start of task $e + 1$:}
\[\fitdisplay{\mu = \frac{n_a}{n_a^{\mathrm{last}}}, \qquad \tau = \frac{\mathrm{tail}_a}{d_a - r_a}, \qquad b = \frac{f}{f + \mu\tau}, \qquad T_a = b\,(I - U_a U_a^\top)}\]
\par\vspace{3pt}$U_a$ is protected exactly and the residual floor $b$ acts outside it; no $d \times d$ matrix is formed (the reported core runs use this exact form; Extended Data Table 4).
\par\vspace{3pt}\emph{Every training step} (gradient constraint, Adam, update constraint):
\[\fitdisplay{\frac{\partial L}{\partial W} = \sum_t \delta_t^\top\,(x_t T_a), \qquad u = \frac{\hat m}{\sqrt{\hat v} + \epsilon}, \qquad u' = u\,T_a, \qquad W \leftarrow W - \eta\,u'}\]
\par\vspace{3pt}with $\delta_t = \partial L / \partial y_t$ at position $t$; the forward pass is unchanged.
\par\vspace{3pt}\emph{At test time} the hippocampus is detached; the model answers alone.
\end{natbox}

The full method adds to this the class-conditioned index (records and class-pattern constraint), state-only conflict detection with its update gate and comparator, the joint-share target and erase.

The model is the sole predictor but not the sole learner: while later tasks are learnt, the hippocampus must keep being read, and switching its read off collapses earlier tasks (Fig. 3).

\subsection{Constraints}

The first five constraints, listed in Supplementary Table 4, exclude each regime of Supplementary Note 1.

\textbf{Further rules.}

\begin{itemize}[leftmargin=1.4em,itemsep=1pt,topsep=2pt,parsep=0pt]
\item[6.] No task-specific backbone expansion; only the output rows required by newly introduced classes are added.
\item[7.] No task- or dataset-specific tuning. Settings may depend on the architecture and model size and are stated explicitly: the learning rate per model family (decoders and BERT/RoBERTa 4e-5 at batch 32, the batch-8 rate 2e-5 scaled by $\sqrt{32/8}$, the square-root batch rule; ViT-B/16 3e-5 and ResNet-50 1e-4 at batch 64) and the width rule for the 8B models and the fp64 basis for models whose widest synapse-group input exceeds 9,000 (Qwen3-4B, Qwen3-8B and Llama-3.1-8B). Each is fixed once per model and used unchanged on every benchmark.
\item[8.] Rules may depend on architecture (e.g. normalisation layout, tied embeddings, width), never on the dataset or label strings.
\item[9.] No stored answer-token ids. Conflict detection reads states only and never checks whether a competing token is an earlier answer.
\item[10.] Training-step cost close to plain fine-tuning: no extra forward pass during a training step, and no $d \times d$ matrices kept between tasks.
\end{itemize}

\subsection{Setting and notation}

\begin{itemize}[leftmargin=1.4em,itemsep=1pt,topsep=2pt,parsep=0pt]
\item[\textbullet] Tasks $e = 0, 1, \dots$ arrive in sequence. Each task brings new classes whose labels are generated as text after a prompt.
\item[\textbullet] The model is trained with cross-entropy on answer tokens. At test time it generates greedily, and accuracy is exact match of the label. No task id is given.
\item[\textbullet] An \textbf{answer-token position} is a position whose next token belongs to the label. $o_t$ is its offset from the first answer-token position of the sentence (0, 1, 2, …).
\item[\textbullet] A class $c$ is an integer id assigned to a label string when it is first seen. Strings and token ids are not kept by the hippocampus.
\end{itemize}

\begin{itemize}[leftmargin=1.4em,itemsep=1pt,topsep=2pt,parsep=0pt]
\item[\textbullet] $x_t \in \mathbb{R}^{d_a}$: input of synapse group $a$ at position $t$;
\item[\textbullet] $h_t \in \mathbb{R}^{d}$: LM-head input (last hidden state) at answer-token position $t$;
\item[\textbullet] $n_{\mathrm{all}}$, $n_{\mathrm{ans}}$: number of valid positions / answer-token positions in a batch.
\end{itemize}

\subsection{Model training}

\emph{Motivation.} In complementary learning systems, the slow cortical learner keeps learning from its own experience and alone produces behaviour\textsuperscript{19}. The model is therefore the plain network, trained on the current task's supervision with the base optimizer; the hippocampus modulates the gradient, the answer-token target at detected conflicts and the realised update (Methods: Operators used while learning a new task, Learning at detected conflicts, The constrained update).

\begin{itemize}[leftmargin=1.4em,itemsep=1pt,topsep=2pt,parsep=0pt]
\item[\textbullet] Plain full fine-tuning: every parameter trains, with answer-token cross-entropy.
\item[\textbullet] AdamW with $\beta_1 = 0.9$, $\beta_2 = 0.999$, $\epsilon$ = 1e-8 and weight decay 5e-4. Learning rate $\eta$ = 4e-5 (2e-5 × $\sqrt{32/8}$, the square-root batch rule; at 8B the width rule scales the step on top), batch size 32, 3 epochs per task, seeds 42, 11 and 3. At this learning rate the decoupled decay factor $1 - \eta \cdot 5 \times 10^{-4} = 1 - 2 \times 10^{-8}$ rounds to 1 in fp32, so weight decay leaves the LLM weights unchanged in practice.
\item[\textbullet] fp32 weights and optimizer state; bf16 autocast forward.
\item[\textbullet] One AdamW instance serves the whole task sequence: its moments are carried across task boundaries, not reset. Moments carried into a new task can point along directions the new memory protects; the update constraint removes that component from the applied update (Proposition 1).
\item[\textbullet] \textbf{Relative step (architecture rule).} Each transformer matrix $m$ uses
\end{itemize}

\[
\fitdisplay{\eta_m = \eta \cdot \min\!\left(1,\ \frac{\operatorname{rms}(W_m)}{\operatorname{mean}_{m'} \operatorname{rms}(W_{m'})}\right)}
\]

computed once from the pre-trained weights. Adam moves every coordinate by about $\eta$ regardless of the weight's size, so small-norm matrices would otherwise change fastest in relative terms.

\begin{itemize}[leftmargin=1.4em,itemsep=1pt,topsep=2pt,parsep=0pt]
\item[\textbullet] \textbf{Width rule (architecture rule).} Under Adam, the change of a matrix's output per step grows with the matrix's width (the observation behind $\mu$P)\textsuperscript{36}. For the 8B models, the backbone matrices and an untied LM head step at
\end{itemize}

\[
\fitdisplay{\begin{gathered}
\eta_m \leftarrow \eta_m \cdot \frac{d_{\mathrm{ref}}}{d_{\mathrm{hidden}}} \\
d_{\mathrm{ref}} = 2560
\end{gathered}}
\]

Vector-like parameters (the input embedding matrix and the normalisation gains) keep $\eta$. $d_{\mathrm{ref}}$ is the hidden width of Qwen3-4B. The rule is set once per model size and applied to the 8B models only ($d_{\mathrm{hidden}} = 4096$, factor 0.625); the models up to 4B, including Llama-3.2-3B ($d_{\mathrm{hidden}} = 3072$), use no width scaling.

\begin{itemize}[leftmargin=1.4em,itemsep=1pt,topsep=2pt,parsep=0pt]
\item[\textbullet] Task 0 is trained with plain AdamW, because the memory is still empty.
\item[\textbullet] Evaluation is the plain model's greedy generation. The hippocampus is detached at test time and never touches outputs.
\end{itemize}

\subsection{Synapse groups: what the memory indexes}

\emph{Motivation.} Indexing theory places the index over the cortical regions whose patterns it points to. Here the natural unit is a group of synapses that read the same input vector: a step on any of them changes earlier outputs through that shared input, so one input statistic measures their interference with the past.

A \textbf{synapse group} is the set of linear maps that read the same input tensor. Per transformer layer there are four synapse groups (Supplementary Table 5).

The \textbf{LM head} is one more synapse group under the same rule. In tied models the LM head is tied to the input embedding matrix.

\textbf{Per-coordinate and per-row synapse groups} are added with a simpler rule (Methods: Operators used while learning a new task): one per normalisation gain (per coordinate) and one for the input embedding matrix (per row). Supplementary Table 6 gives the counts per decoder.

\begin{itemize}[leftmargin=1.4em,itemsep=1pt,topsep=2pt,parsep=0pt]
\item[\textbullet] Linear synapse groups = $4 \times$ layers + 1, and linear maps = $7 \times$ layers + 1.
\item[\textbullet] Gain synapse groups are the RMSNorm gains:
\begin{itemize}[leftmargin=1.2em,itemsep=1pt,topsep=2pt,parsep=0pt]
\item[$\circ$] Qwen3: input, post-attention, q-norm and k-norm per layer, plus the final norm;
\item[$\circ$] Llama: two per layer, plus the final norm.
\end{itemize}
\end{itemize}

The memory of each linear synapse group is a moment over the positions where that group receives a learning signal:

\begin{itemize}[leftmargin=1.4em,itemsep=1pt,topsep=2pt,parsep=0pt]
\item[\textbullet] all positions for backbone synapse groups, with answer-token positions reweighted;
\item[\textbullet] answer-token positions only for the LM head.
\end{itemize}

This is the only difference between synapse groups, and it follows from where the loss enters.

\subsection{What the memory stores}

\emph{Motivation.} Following indexing theory\textsuperscript{21,22}, the store keeps compact summaries of where earlier experience lies (subspaces, energies, class means and counts), never the experience itself.

\begin{table*}[!t]
{\tabsize\setlength{\tabcolsep}{3pt}\begin{tabular*}{\textwidth}{@{\extracolsep{\fill}}llll@{}}
\toprule
\textbf{Part of the hippocampus} & \textbf{Entry} & \textbf{Size} & \textbf{Content}\\
\midrule
Linear synapse group $a = 1, \dots, A$ & $U_a$ & $d_a \times r_a$ & protected subspace (orthonormal)\\
 & $\lambda_a$ & $r_a$ & energy per protected direction\\
 & $\mathrm{tail}_a$ & scalar & per-position energy outside $U_a$ (running mean over episodes)\\
 & $f_a$ & scalar & noise floor of the latest episode (spectral cutoff, in energy units)\\
 & $n_a$, $n_a^{\mathrm{last}}$ & scalars & positions seen in all episodes / in the latest episode\\
 & $P_a$ & $R \times d_a$ & class-level mean input pattern per record (running sums)\\
State records & $N_j$, $\sum h$ & per record & per record $j = (\text{class } c, \text{offset } o)$: count and sum, giving the mean $H_j$\\
LM-head spread sketch & $\sum x\,(x^\top \Omega)$ & per record & $\Omega$ a fixed $d \times 16$ Gaussian\\
Gain group $g$ & energies & $d_g$ & per-coordinate energies, protected coordinates\\
Input embedding matrix & $\nu_v$ & per token & prompt use (running mean), protected tokens, floor $f_{\mathrm{tok}}$, weight $\mu_{\mathrm{tok}}$\\
\bottomrule
\end{tabular*}}
\end{table*}

$R$ is the number of (class, offset) records, the sum of the label lengths of all classes seen (a few hundred at the end of TACRED).

No $d \times d$ matrix is kept between tasks. At 8B the subspace memory takes 11–12 GB after the eight TACRED tasks and 24–26 GB after the 15 CLINC150 tasks (Llama-3.1-8B), two-thirds of it on the host.

\subsection{Sleep: writing and erasing}

\emph{Motivation.} In sleep-dependent consolidation, recent experience is reactivated offline, after it was acquired\textsuperscript{37,38}. Here the reactivated experience is the finished task's own training set, read again without gradients in an encoding phase, and what the phase writes is the hippocampal memory, not the network's weights. The encoding uses the read, write and erase mechanism of Gated DeltaNet\textsuperscript{29}, the gated form of the delta rule\textsuperscript{39}: new keys are written with gate $\beta_e$, old content along them is erased by $\beta_e(1 - \rho_i)$, where $\rho_i$ (below) is the part of key $i$ that conflicting sentences used, and directions that no new key overwrote keep their content.

After each task $e$ except the last (nothing would read the memory after it), the model, in inference mode, re-reads task $e$'s own training data without gradients. Earlier tasks' data is never read.

\textbf{Episode moment (per linear synapse group).}

\[
\fitdisplay{\begin{aligned}
\text{backbone:}\quad & C_e = \sum_{t\,\in\,\text{all positions}} x_t x_t^\top + \frac{n_{\mathrm{all}}}{n_{\mathrm{ans}}} \sum_{t\,\in\,\text{answer positions}} x_t x_t^\top, \\
 & \qquad n_e = \textstyle\sum n_{\mathrm{all}} \\
\text{LM head:}\quad & C_e = \sum_{t\,\in\,\text{answer positions}} h_t h_t^\top, \\
 & \qquad n_e = \textstyle\sum n_{\mathrm{ans}} \\
& \hat C_e = C_e / n_e
\end{aligned}}
\]

The $d \times d$ moments exist only during this pass. They are accumulated for a few synapse groups at a time, chosen so that accelerator memory stays below the training peak; each further set re-reads the same batches in the same order, so the sleep never needs more accelerator memory than training did.

\textbf{New keys (spectral cutoff adapted from \textsuperscript{28}).} The Gavish–Donoho rule is optimal for a low-rank signal in white noise; transformer activations, with correlated positions and reweighted answer-token positions, need not meet that model, so it serves here as a data-dependent cutoff.

\[
\fitdisplay{\begin{aligned}
\hat C_e &= V\,\mathrm{diag}(\lambda)\,V^\top, \qquad \sigma_i = \sqrt{\lambda_i}\ \text{in decreasing order} \\
\zeta &= \frac{\min(n_e, d)}{\max(n_e, d)}, \qquad \omega = 0.56\,\zeta^3 - 0.95\,\zeta^2 + 1.82\,\zeta + 1.43 \\
\theta &= \omega \cdot \operatorname{median}(\sigma), \qquad k_e = \#\{\, i : \sigma_i > \theta \,\} \\
K &= [k_1, \dots, k_{k_e}]\ \text{(top eigenvectors)}, \qquad \lambda_K\ \text{their energies} \\
f_e &= \theta^2\ \text{(the episode's noise floor)}, \qquad \mathrm{tr}_e = \textstyle\sum_i \lambda_i\ \text{(per-position total energy)}
\end{aligned}}
\]

\textbf{Conflict fraction per key (second read of the same episode).} From task 1 on, a second read tests each answer-token position for a state conflict with the earlier records (Methods: Conflict detection). Then, per key:

\[
\fitdisplay{\rho_i = \frac{\sum_{t\,\in\,\text{conflicting answer positions}} (k_i^\top x_t)^2}{\sum_{t\,\in\,\text{all answer positions}} (k_i^\top x_t)^2}}
\]

Only $k_e$-long sums are kept per synapse group. $\rho_i$ is the part of key $i$ that sentences colliding with earlier records relied on.

\textbf{Gated delta write.} Let the stored memory be $U$ ($d \times r$), $\lambda_U$ and $n$.

\[
\fitdisplay{\begin{aligned}
\beta_e &= \frac{n_e}{n + n_e} \\
 & \qquad \text{write strength: the episode's fraction of all positions} \\
B &= \operatorname{orth}([U, K]) \\
 & \qquad \text{merge basis, } d \times (r + k_e) \\
S &= (B^\top U)\,\mathrm{diag}(\lambda_U)\,(B^\top U)^\top \\
 & \qquad \text{earlier memory in } B \\
\kappa_i &= B^\top k_i \\
 & \qquad \text{new keys in } B
\end{aligned}}
\]

\[
\fitdisplay{\begin{aligned}
\text{erase:}\quad & \varepsilon_i = \beta_e (1 - \rho_i), \quad w_i = 1 - \sqrt{1 - \varepsilon_i}, \quad G = I - \textstyle\sum_i w_i\,\kappa_i \kappa_i^\top, \quad S \leftarrow G S G^\top \\
\text{write:}\quad & S \leftarrow S + \beta_e \textstyle\sum_i \lambda_{K,i}\,\kappa_i \kappa_i^\top \\
\text{prune:}\quad & S = V'\,\mathrm{diag}(m)\,V'^\top, \quad U' = B\,V'_{[:,\,m > f_e]}, \quad \lambda_{U'} = m_{[m > f_e]} \\
& n' = n + n_e, \quad n^{\mathrm{last}} = n_e, \quad f = f_e \\
\text{tail:}\quad & \mathrm{tail}_e = \mathrm{tr}_e - \textstyle\sum_j u_j'^\top \hat C_e\, u_j', \qquad \mathrm{tail}' = (1 - \beta_e)\,\mathrm{tail} + \beta_e\,\mathrm{tail}_e
\end{aligned}}
\]

The erase removes less where conflicting sentences used the key. As a congruence it keeps $S$ positive semi-definite, and the energy along $\kappa_i$ scales by exactly $(1 - \varepsilon_i)$.

\emph{What the residual floor tracks.} The stored residual energy $\mathrm{tail}$ is a running mean, weighted by $\beta_e$, of each task's own per-position energy outside the span $U'$ that is current when the episode is written. Energy that the merge prunes from the stored memory (merged directions at or below $f_e$) is not transferred to $\mathrm{tail}$; the earlier residual energy persists only through the $(1 - \beta_e)$ decay. The floor therefore measures how much recent experience lay outside the current span, in the current basis. The term $\tau(I - UU^\top)$ is a positive semi-definite protection term, not a conserved account of every mass ever discarded, and one scalar per synapse group summarises the residual structure on average: residual directions of equal mean energy can matter differently for a class, which the class-pattern constraint covers for the directions that earlier class means occupy. If only the residual energy is known, the floor charges the average case, $\tau \lVert \Delta W (I - UU^\top) \rVert_F^2$ with $\tau = \mathrm{tail}/(d - r)$, whereas the worst case over all residual covariances with that energy is $\mathrm{tail} \cdot \lVert \Delta W (I - UU^\top) \rVert_2^2$, reached when the residual energy sits on the single direction the step moves most. Requiring zero interference for every residual covariance with that energy would freeze the complement; instead, the floor protects the residual on average and the class-pattern constraint removes the components of earlier class means that lie outside $U$. The spectral cutoff takes the median over the $\min(n_e, d)$ singular values that are not structurally zero (when $n_e < d$ the $d - n_e$ structural zeros are excluded); $f$ is used as computed, guarded only against zero; when $r = d$ the residual term vanishes and only the stored span is protected.

At the first task, $U = K$ and $\lambda_U = \lambda_K$.

\textbf{Class-level records.} The first read also accumulates:

\begin{itemize}[leftmargin=1.4em,itemsep=1pt,topsep=2pt,parsep=0pt]
\item[\textbullet] State record (LM-head input): for $j = (c, o_t)$, $N_j \mathrel{+}= 1$ and $\Sigma h_j \mathrel{+}= h_t$, giving $H_j = \Sigma h_j / N_j$.
\item[\textbullet] Pattern record (every group $a$): $\Sigma x^a_j \mathrel{+}= x_t$ at answer-token positions, giving $P^a_j = \Sigma x^a_j / N_j$.
\item[\textbullet] Gain group $g$, coordinate $i$: $e_i$ = running mean of $(x / \operatorname{rms}(x))_i^2$; coordinates above the Gavish–Donoho cutoff are protected.
\item[\textbullet] Embedding rows: $\nu_v$ = running mean prompt use of token $v$; tokens above the Gavish–Donoho cutoff are protected; $f_{\mathrm{tok}}$ = the smallest protected $\nu_v$; $\mu_{\mathrm{tok}}$ = all positions so far / the latest episode's.
\end{itemize}

These are class means and counts. No sentence-level state is stored. The memory thus holds geometric statistics formed per earlier class, but it never re-supplies earlier sentences, answer strings or answer-token targets.

\subsection{Operators used while learning a new task}

\emph{Motivation.} At wake the hippocampus is read, and its content becomes the operators that shape every cortical update. Cascade models of synaptic memory hold that the more a memory has been consolidated, the less plastic the synapses that carry it become, and that this happens gradually rather than by freezing\textsuperscript{24,40}. Metaplasticity makes the same point for a synapse's history of use\textsuperscript{23}. The operators below implement this per input direction: a direction's plasticity factor falls with its stored energy and with the amount of earlier experience, and an embedding row's plasticity factor falls with its earlier use.

\textbf{State neighbourhoods (task start).}

\[
\fitdisplay{\begin{gathered}
\bar c = \operatorname{mean}_j H_j \\
M_j = \frac{H_j - \bar c}{\lVert H_j - \bar c \rVert} \\
r_j = \max\{\, M_j^\top M_{j'} : o_{j'} = o_j,\ c_{j'} \neq c_j \,\}
\end{gathered}}
\]

where $r_j$ is the neighbourhood radius of record $j$.

\textbf{Pattern basis and protected subspace (task start), with the class-pattern constraint.}

\[
\fitdisplay{\begin{aligned}
Q_a &= \operatorname{orth}(\{\, P^a_j : \text{all earlier records} \,\}) \\
 & \qquad \text{(QR; near-zero diagonal entries dropped)} \\
R_a &= \operatorname{orth}\big((I - U U^\top)\, Q_a\big) \\
 & \qquad \text{the patterns' part outside } U \\
W_a &= [U, R_a] \\
 & \qquad \text{the protected basis}
\end{aligned}}
\]

$U$ holds the high-energy directions of earlier inputs. An earlier class's mean can still have a component in the low-energy residual subspace. The class-pattern constraint adds those components to the protected subspace, so a new near-duplicate's output row cannot grow along them and win on the earlier class's own sentences. It costs at most one extra column per record and stores nothing new, because the patterns are the comparator's records.

\emph{Numerically orthonormal protected basis (wide models).} When the patterns lie almost entirely inside $U$, the two-step construction above can leave $R_a$ far from orthonormal in finite precision, and the operator would then amplify the step along those columns instead of constraining it; this concentrates in synapse groups with very wide inputs, the MLP down-projections. Models whose widest synapse-group input exceeds 9,000 (Qwen3-4B, Qwen3-8B and Llama-3.1-8B, with down-projection inputs of 9,728, 12,288 and 14,336) therefore build $W_a$ as one fp64 QR of $[U, P^a]$ with $U$ first, keeping a pattern column only if its diagonal entry exceeds $\max(d_a, c_a) \cdot \epsilon_{\mathrm{fp32}}$ times its norm, with $c_a$ the number of columns of $[U, P^a]$. The narrower models use the two-step construction.

\textbf{Minimum-interference operator (per linear synapse group).} A step $\Delta W$ on synapse group $a$ changes every earlier output by $\Delta W x$. The step is chosen to minimise a local, per-layer surrogate, with the layer's input distribution held fixed: the new loss, the step's size, and the earlier tasks' expected output change, weighted by the amount of earlier experience $\mu$ and measured against the group's noise floor $f$. The surrogate controls the part $\Delta W x$ of a layer's output change, not the change $W \Delta x$ carried in from earlier layers:

\[
\fitdisplay{\min_{\Delta W}\ \langle G, \Delta W \rangle + \frac{1}{2\eta}\Big( \lVert \Delta W \rVert^2 + \frac{\mu}{f}\, \operatorname{tr}\big(\Delta W\, C_{\mathrm{mem}}\, \Delta W^\top\big) \Big) \quad \text{subject to} \quad \Delta W R_a = 0}
\]

\[
\fitdisplay{\begin{aligned}
G &= \partial L / \partial W \\
 & \qquad \text{(the plain gradient)} \\
\mu &= n / n^{\mathrm{last}} \\
 & \qquad \text{(earlier positions relative to one episode)} \\
C_{\mathrm{mem}} &= U\,\mathrm{diag}(\lambda)\,U^\top + \tau\,(I - U U^\top), \quad \tau = \mathrm{tail} / (d - r) \\
\Pi_a &= I - R_a R_a^\top \\
 & \qquad \text{(removes the class-pattern columns)}
\end{aligned}}
\]

\[
\fitdisplay{\Rightarrow\quad \Delta W = -\eta\, G\, \Pi_a \big(I + \mu\, C_{\mathrm{mem}} / f\big)^{-1} = -\eta\, G\, T_a}
\]

The constraint $\Delta W R_a = 0$ is the class-pattern constraint (above). Because $R_a$ lies outside $U$, where $C_{\mathrm{mem}}$ is isotropic, the projector $\Pi_a$ commutes with $(I + \mu C_{\mathrm{mem}} / f)$, and the constrained minimiser is exactly the unconstrained one followed by $\Pi_a$. The same step is the limit $c \to \infty$ of a finite penalty $c R_a R_a^\top$ added to $C_{\mathrm{mem}}$.

$C_{\mathrm{mem}}$ is the hippocampal metric written by sleep, and $\operatorname{tr}(\Delta W C_{\mathrm{mem}} \Delta W^\top)$ is a memory-weighted local functional-change penalty. It equals the expected output change $\mathbb{E}_{\mathrm{old}} \lVert \Delta W x \rVert^2$ on earlier inputs only in the special case where $C_{\mathrm{mem}}$ equals their empirical second moment. With erase disabled, the memory accumulates low-rank records of each episode's input statistics, each entering with weight $\beta_e$ (Methods: Sleep). This is a history-weighted protection metric, not the sample-averaged second moment: for two equal-sized tasks with the same energy $\lambda$ along one direction, it holds $\lambda + \lambda/2 = 1.5\lambda$, whereas the pooled second moment is $\lambda$. The derivation of the operator needs only $C_{\mathrm{mem}} \succeq 0$. Without erase the method gives the same result on TACRED (71.32 against 71.27; Supplementary Table 1), because the protected subspace stays well below full there; the gated erase is designed for task sequences that press on the memory.

In closed form, each protected direction keeps the plasticity factor $a_i$ and the residual subspace keeps the residual plasticity factor $b$:

\[
\fitdisplay{\begin{aligned}
a_i &= \frac{f}{f + \mu \lambda_i} \\
 & \qquad \text{(protected direction } u_i) \\
a_i &= 0 \\
 & \qquad \text{(class-pattern columns } R_a\text{, from the constraint } \Delta W R_a = 0) \\
b &= \frac{f}{f + \mu \tau} \\
 & \qquad \text{(residual)} \\
T_a &= b\, I + W_a\, \mathrm{diag}(a - b)\, W_a^\top
\end{aligned}}
\]

$T_a$ is applied as $x T_a = b\, x + \big((x W_a) \odot (a - b)\big) W_a^\top$ and never formed as $d \times d$.

$T_a$ depends only on the memory. It is built once per task and does not change while the task is learnt.

\emph{Relation to conceptor and regularised-inverse projections.} Leaving the class-pattern constraint aside, $T_a = (I + \alpha^2 C_{\mathrm{mem}})^{-1}$ with $\alpha^2 = \mu/f$, derived here from the memory-weighted output-change objective. Rational spectral filters of this kind also appear as the complement of a conceptor ($I - K = (I + \alpha^2 S)^{-1}$ for $K = S (S + \alpha^{-2} I)^{-1}$, used by CODE-CL) and in OWM's regularised inverse. DeltaHippo's operator differs in the metric it acts on and in how it is set and applied:

\begin{itemize}[leftmargin=1.4em,itemsep=1pt,topsep=2pt,parsep=0pt]
\item[\textbullet] $C_{\mathrm{mem}}$ is a history-weighted low-rank record plus a residual floor $\tau$ for the energy the compression leaves out, rather than a full or truncated second moment;
\item[\textbullet] the operator is applied to the realised Adam step as well as the gradient, in every synapse group, output head, tied input embedding matrix and gains included;
\item[\textbullet] the base update $u T_a$ is exactly orthogonal to class patterns outside $U$;
\item[\textbullet] the aperture $\alpha^2 = \mu/f$ is set per synapse group and per task by the episode's spectral cutoff and the ratio of earlier to new experience, so no global aperture is chosen by hand.
\end{itemize}

\textbf{Input embedding matrix (per row).} A change $\Delta v$ of embedding row $v$ moves every earlier output that used token $v$ as input. The same objective gives each row the plasticity factor

\[
\fitdisplay{s_v = \frac{f_{\mathrm{tok}}}{f_{\mathrm{tok}} + \mu_{\mathrm{tok}}\, \nu_v}}
\]

on its step. In a tied model the row is also an LM-head row, so the LM-head operator acts on it as well, and the factor protects the rows of words that earlier prompts are full of, such as label words. The gradient rows of the embedding matrix are also multiplied by the per-token novelty factor $D_{\mathrm{tok}}$ (below).

\textbf{Normalisation gains (per coordinate).} Rebuilt at steps 1, 2, 4, 8, … of each task from the present energies:

\[
\fitdisplay{\begin{gathered}
D_i = \operatorname{clip}\!\left( \frac{e_i^{\mathrm{now}} - e_i^{\mathrm{past}}}{e_i^{\mathrm{now}}},\, 0,\, 1 \right) \\
D_i = 0 \ \text{on protected coordinates}
\end{gathered}}
\]

The coordinates protected for earlier tasks receive no update; the others pass only their novel energy. $D_{\mathrm{tok}}$ is formed by the same rule per token.

\subsection{Conflict detection}

\emph{Where the difficulty lies.} In class-incremental learning by generation, the loss sits at the answer token where near-duplicate labels first differ. It is dominated by \emph{near-duplicate} classes: a new class whose answer state is nearly the same as an earlier one absorbs it (e.g. \emph{city of residence} $\rightarrow$ \emph{city of death}, \emph{state or provinces of residence} $\rightarrow$ \emph{state or province of birth}). The update must therefore do two things:

\begin{itemize}[leftmargin=1.4em,itemsep=1pt,topsep=2pt,parsep=0pt]
\item[1.] keep the directions that earlier tasks use from being overwritten, in proportion to how much they were used;
\item[2.] specifically keep each earlier class's own answer state from being taken over.
\end{itemize}

All of this has to happen without storing sentences, and without a cost that grows to $d \times d$ per layer.

\emph{Motivation.} Conflict detection is the match–mismatch comparison of the Introduction\textsuperscript{25,26}, made between the present answer state and the earlier class records. In words: an answer-token position is in conflict with an earlier class when its centred answer state is closer to that class's record than the record's nearest other class at the same answer-token offset. The record directions $M_j$, their radii $r_j$ and the centre $\bar c$ are those of Methods: Operators used while learning a new task.

At every answer-token position $t$ of a training batch (computed in the forward pass at the LM-head input):

\[
\fitdisplay{\begin{aligned}
\hat h_t &= \frac{h_t - \bar c}{\lVert h_t - \bar c \rVert}, \qquad s_j = \hat h_t^\top M_j \\
\operatorname{conflict}(t) &\iff \exists\, j : o_j = o_t \ \text{and} \ s_j > r_j \\
j^*(t) &= \mathop{\mathrm{arg\,max}}_{j\,:\,o_j = o_t}\, (s_j - r_j) \\
 & \qquad \text{(the record entered most deeply)}
\end{aligned}}
\]

Here $z_g$ is the logit of the present sentence's own gold token and $z_r$ the largest other logit (its value only).

\begin{itemize}[leftmargin=1.4em,itemsep=1pt,topsep=2pt,parsep=0pt]
\item[\textbullet] "Gold" is the label of the present training sentence, which is part of the current task's supervision.
\item[\textbullet] The identity of the competing token is never read, and nothing checks whether it was an earlier answer.
\item[\textbullet] Class ids enter only through the records: they group states into class means and define the radius against \emph{other} classes.
\end{itemize}

\subsection{Learning at detected conflicts}

\emph{Motivation.} The update gate implements the shift from encoding towards retrieval that familiar input causes\textsuperscript{27}, and the comparator CA1's comparison of recall with input\textsuperscript{26}. The joint-share target keeps the two classes apart as a computational device without a specific biological counterpart.

\textbf{Update gate (per step).}

\[
\fitdisplay{\psi_t = \begin{cases} \dfrac{N_{j^*}}{N_{j^*} + n_c / E} & \text{if } \operatorname{conflict}(t) \\[6pt] 0 & \text{otherwise} \end{cases} \qquad\qquad p = \operatorname{clip}\!\big( \operatorname{mean}_t (1 - \psi_t),\, 0,\, 1 \big)}
\]

with $n_c$ the sentences of the present class seen in this task and $E$ the epochs done ($E \ge 1$).

The update is applied whole with probability $p$, drawn independently of the data order; otherwise the learning rate is zero for that step and only Adam's moments are updated. A probability is used instead of a scaled gradient because Adam's normalisation would cancel the scale.

\textbf{Joint-share target (answer-token positions after the first, $o_t > 0$).}

Let $\Delta$ be the change of the present class's mean answer state since the task start, and $\varphi_j$ the fraction of $\Delta$ lying in record $j$'s space (its centred mean and the span of its spread sketch). Then

\[
\fitdisplay{\begin{aligned}
E_{\mathrm{rec}} &= \frac{\max_j \varphi_j - \operatorname{median}_j \varphi_j}{\max_j \varphi_j} \\
 & \qquad \text{near-duplicate evidence for the present (class, offset)} \\
a_t &= \operatorname{clip}(E_{\mathrm{rec}}\, \psi_t,\, 0,\, 1) \\
y_t &= (1 - a_t)\, \operatorname{onehot}(\mathrm{gold}) + a_t\, \frac{q_{\neq \mathrm{gold}}}{1 - q_{\mathrm{gold}}}
\end{aligned}}
\]

$a_t$ is set only at conflicting positions of a present (class, offset) that has been seen in conflict while its gold lost.

Here $q$ is the model's softmax and $q_{\neq \mathrm{gold}}/(1 - q_{\mathrm{gold}})$ is its distribution over the other tokens. The gradient at the logits is replaced by $(q - y_t)$. Algebraically,

\[
\fitdisplay{q - y_t = \Big( 1 - \frac{a_t}{1 - q_{\mathrm{gold}}} \Big)\, (q - e_{\mathrm{gold}})}
\]

so the target rescales the position's cross-entropy gradient by a scalar. It drives the gold probability towards $1 - a_t$ rather than towards 1: while $q_{\mathrm{gold}} < 1 - a_t$ the gradient keeps the cross-entropy direction with a smaller magnitude, and once $q_{\mathrm{gold}}$ exceeds $1 - a_t$ its sign reverses and the position is pushed back to $1 - a_t$. The other tokens share the remainder in proportion to the model's own distribution; the target never reads which earlier label a competing token is. The effect is that a new class does not take over the decision region it shares with an earlier near-duplicate beyond the share $a_t$ set by their joint experience. On Llama-3.2-1B TACRED the target is active on 2–19\% of the answer-token positions of each task (three runs).

\textbf{Comparator (CA1).}

\[
\fitdisplay{\begin{gathered}
A = \{\, t : \operatorname{conflict}(t) \ \text{and} \ z_g \le z_r \,\} \\
X_p = X - (X Q_a)\, Q_a^\top
\end{gathered}}
\]

$A$ is the comparator set, the conflicting positions whose gold does not win yet; $X$ holds the rows of synapse group $a$'s input at the positions in $A$, and $X_p$ is their mismatch with every earlier pattern.

Supplementary Table 7 summarises the three responses and their effects.

\subsection{The constrained update}

\emph{Motivation.} The network's synapses change only through this update. The operators of Methods: Operators used while learning a new task and the comparator of Methods: Learning at detected conflicts act on it twice, on the gradient and on Adam's update, because Adam's per-coordinate normalisation would otherwise rotate a constrained gradient back into protected directions.

\textbf{Constrained-input gradient (every linear map of every synapse group).} The forward pass is unchanged, $y_t = x_t W^\top$ in bf16. Only the weight gradient changes:

\[
\fitdisplay{\tilde x_t = \begin{cases} x_t\, T_a & t \notin A \\ X_p[t] & t \in A \end{cases}, \qquad \frac{\partial L}{\partial W} = \sum_t \delta_t^\top\, \tilde x_t \quad (\delta_t = \partial L / \partial y_t), \qquad \frac{\partial L}{\partial x} = \delta\, W \ \ \text{(unchanged)}}
\]

\begin{itemize}[leftmargin=1.4em,itemsep=1pt,topsep=2pt,parsep=0pt]
\item[\textbullet] This equals $G T_a$, with $G$ the plain gradient, except at the positions in $A$, whose rows use the mismatch $X_p$.
\item[\textbullet] It is computed on the batch's rows instead of the weight's rows.
\item[\textbullet] $x T_a$ is computed once per synapse group and input, because the maps of one synapse group share their input.
\item[\textbullet] The embedding matrix's gradient rows are multiplied by the token's novelty factor $D_{\mathrm{tok}}$; in tied models they also pass the LM-head operator.
\end{itemize}

\textbf{Proposition 1 (a constrained gradient is not a constrained update).} Let $Z$ span the hard-constrained input directions of a linear map $W$ within one task, so that a step $\Delta W$ leaves the outputs on those inputs unchanged iff $\Delta W Z = 0$, and let every gradient of the task be constrained, $G' = G(I - ZZ^\top)$, so $G'Z = 0$.

\begin{itemize}[leftmargin=1.4em,itemsep=1pt,topsep=2pt,parsep=0pt]
\item[\textbullet] \emph{SGD.} The step is a constrained gradient, so $\Delta W Z = 0$.
\item[\textbullet] \emph{Momentum SGD.} The step is a linear combination of constrained gradients and of the momentum carried into the task. It satisfies $\Delta W Z = 0$ only if that carried momentum also has no component along $Z$.
\item[\textbullet] \emph{Adam.} The step is $u = \hat m \oslash (\sqrt{\hat v} + \epsilon)$. Write row $r$ as $u_r = \hat m_r \Lambda_r$ with $\Lambda_r = \mathrm{diag}\big(1/(\sqrt{\hat v_r} + \epsilon)\big)$. Even if the bias-corrected first moment $\hat m_r$ lies in the constrained complement ($\hat m_r Z = 0$), $\hat m_r \Lambda_r Z$ can be non-zero: the constraint is guaranteed for every such $\hat m_r$ only if $(I - ZZ^\top) \Lambda_r ZZ^\top = 0$, which for a diagonal $\Lambda_r$ is equivalent to $\Lambda_r$ commuting with $ZZ^\top$, that is, to a scale that is constant over the coordinates $Z$ mixes. A moment carried into the task with a component along $Z$ violates the condition even before scaling. In general, therefore, Adam returns part of a constrained gradient into protected directions.
\item[\textbullet] \emph{Update constraint.} Passing the realised base update through the operator, $u' = u T_a$, restores $u'Z = 0$ on the hard-constrained span and scales every other direction by its plasticity factor, whatever Adam's moments contain.
\end{itemize}

The operator is derived from the local surrogate; the applied Adam step is not that surrogate's exact minimiser. Under plain SGD, constraining both the gradient and the update would apply $T_a$ twice ($G T_a^2$). Under Adam the two constraints are separated by the coordinate-wise normalisation: the gradient constraint shapes the moments Adam accumulates, and the update constraint re-imposes it on the update they produce. The statement concerns this constrained base update. Two further terms are added deliberately and are not covered by it: the comparator's restoration, which reopens the update along the mismatch of the inputs at the positions in $A$ (Methods: Learning at detected conflicts) and, because that mismatch is orthogonal to the pattern basis $Q_a$ but not necessarily to its part $R_a$ outside $U$, can re-enter the class-pattern directions; and, in the LLMs, decoupled weight decay, a uniform shrink of $\eta \cdot 5 \times 10^{-4}$ per step outside the constraint (it rounds away in fp32 at $\eta = 4 \times 10^{-5}$; the encoder and vision versions fold decay into the update before the constraint, Methods: Encoders and vision models). Within the full system, removing the update constraint costs 9.4 points of final accuracy over three runs (Extended Data Table 4).

\textbf{Update (AdamW, then the operator, then the comparator's restoration).}

\[
\fitdisplay{\begin{aligned}
m &\leftarrow \beta_1 m + (1 - \beta_1)\, g, \qquad v \leftarrow \beta_2 v + (1 - \beta_2)\, g^2, \qquad u = \frac{m / (1 - \beta_1^{\,t})}{\sqrt{v / (1 - \beta_2^{\,t})} + \epsilon} \\
\tilde Q_a &= \operatorname{orth}(\text{row space of } X_p), \qquad \Gamma_a = (I - T_a)\, \tilde Q_a \\
u' &= u\, T_a + (u\, \Gamma_a)\, \tilde Q_a^\top \\
W &\leftarrow W - \eta_m\, u'
\end{aligned}}
\]

The update is constrained again, and the comparator restores what was removed along the mismatch. Embedding rows take $u_v \leftarrow s_v\, u_v$; gains take $g \leftarrow g \odot D$ and $u \leftarrow u \odot D$.

\subsection{One training step and one task}

\emph{Design.} The memory is used only during training: it is read at every training step and detached when the model answers.

\begin{table*}[!t]
{\tabsize\setlength{\tabcolsep}{3pt}\begin{tabular*}{\textwidth}{@{\extracolsep{\fill}}ll@{}}
\toprule
\textbf{Stage} & \textbf{What is computed}\\
\midrule
Forward (bf16) & gains and embedding rows: present energies on rebuild steps; LM-head input: conflicts, $p$, $a_t$, comparator positions\\
Before backward & step count; on steps 1, 2, 4, …: rebuild the gain and per-token novelty factors $D$, $D_{\mathrm{tok}}$\\
Backward & constrained-input weight gradients (The constrained update); joint-share target at the logits; $D_{\mathrm{tok}}$ on embedding rows\\
After backward & gains $g \odot D$; comparator bases $\tilde Q_a$, $\Gamma_a$\\
Optimizer step & AdamW moments and $u$; $u\, T_a$; comparator restoration; embedding-row factor $s_v$ and gain novelty factor $D$; $W \leftarrow W - \eta\, \mathbf{1}[\text{applied}]\, u'$\\
\bottomrule
\end{tabular*}}
\end{table*}

A task runs as follows:

\begin{itemize}[leftmargin=1.4em,itemsep=1pt,topsep=2pt,parsep=0pt]
\item[1.] task start: build the neighbourhoods, $Q_a$, $W_a$ and the minimum-interference operators;
\item[2.] train for 3 epochs;
\item[3.] sleep: write the memory and records.
\end{itemize}

At evaluation the hippocampus is detached and the model generates on its own.

\subsection{Encoders and vision models}

\emph{Motivation.} The principle is the same for every architecture: index the synapse groups, write the memory at sleep, detect matches with earlier records and shape the step. What changes is how the model is read (which maps form a synapse group, where the loss enters, how the classifier is protected) and which operator shapes the backbone update.

\textbf{Synapse groups.} Synapse groups are found in one probing forward pass: a synapse group is the set of linear maps that read the same input tensor (BERT/RoBERTa: query/key/value, attention output, intermediate, output; ViT: the same four, plus the patch embedding). A convolution is a linear map on its unfolded input, and convolutions that read the same tensor with the same unfold form one synapse group (ResNet-50: every convolution; ViT: the patch embedding). A bias is the weight on a constant input. LayerNorm and BatchNorm gains (ResNet: with fixed pre-trained running statistics) are per-coordinate synapse groups, and every embedding matrix (token, position, token type; ViT: the [CLS] token and position embeddings) is a per-row synapse group.

\textbf{Classifier.} Each task adds a linear head on the decision feature ([CLS] for BERT/RoBERTa/ViT, the global pool for ResNet-50); the logits are the concatenation of all heads so far. There is one decision feature per input, so each record corresponds to one class.

\begin{itemize}[leftmargin=1.4em,itemsep=1pt,topsep=2pt,parsep=0pt]
\item[\textbullet] \emph{Earlier heads} are constrained row by row: the update of row $k$ and its bias is constrained to be orthogonal to the region of class $k$, the Gavish–Donoho top directions of its augmented decision-state moment $\mathbb{E}\big[[h; 1][h; 1]^\top\big]$, a class-level statistic taken at sleep.
\item[\textbullet] \emph{The present head}, which has no memory yet, is constrained only to be orthogonal to the earlier classes' mean decision states (vision: $[w; b]$ jointly orthogonal to $[\mathrm{mean}; 1]$).
\end{itemize}

\textbf{Detection with one decision feature.} A decision state is in conflict with an earlier record if its centred cosine to that record exceeds its centred cosine to the running mean of its own present class.

\textbf{Components per architecture.} Supplementary Table 8 lists the components used in each architecture.

\begin{itemize}[leftmargin=1.4em,itemsep=1pt,topsep=2pt,parsep=0pt]
\item[\textbullet] The \emph{novelty factor} is $c = \operatorname{clip}(1 - \tau_{\mathrm{past}} / \tau_{\mathrm{now}}, 0, 1)$, the novel part of the present energy outside the protected subspace $W$ ($\tau_{\mathrm{past}}$ and $\tau_{\mathrm{now}}$, the earlier and the present energy outside $W$); the class-pattern constraint applies as in Methods: Operators used while learning a new task. This projection blocks the update on the span $W$ completely and passes only the novel part of the rest; it does not use the graded, memory-strength-dependent plasticity factors of Methods: Operators used while learning a new task.
\item[\textbullet] \emph{Residual-stream units}: an embedding row's change enters the first synapse group through the embedding normalisation as $(\gamma/\sigma)\, \Delta v$, so its plasticity factor uses that group's floor $f_A$: $s_v = f_A / \big(f_A + \mu \nu_v \cdot \operatorname{mean}(\gamma^2)/\sigma^2\big)$.
\item[\textbullet] \emph{Attention-sink weight}: in an attention-input synapse group, the decision feature's position is read by every position $i$ with weight $a_{i0}$, so its interference counts with weight $1 + \sum_i a_{i0}^2$, in the memory and in the step.
\item[\textbullet] BERT/RoBERTa: backbone moments are taken over all attended positions alike, the sleep's forward passes run with dropout active, as in training, and each post-LN gain coordinate receives the protection that the synapse group it feeds gives that coordinate.
\item[\textbullet] ViT: the last block's attention output, MLP and final LayerNorm count only the [CLS] row, which alone reaches the decision.
\end{itemize}

\subsection{Cost}

Mean over three runs. Llama-3.2-1B was measured on one NVIDIA H100 80GB GPU (one run per GPU), Llama-3.1-8B on one NVIDIA B200 GPU (183 GB) running alone, with gradient checkpointing and the width rule. Wall time is the sum of training, sleep and evaluation; training time at 8B includes building each task's operators (Supplementary Table 9).

\begin{itemize}[leftmargin=1.4em,itemsep=1pt,topsep=2pt,parsep=0pt]
\item[\textbullet] \textbf{Training step.} On Llama-3.2-1B, 0.16–0.29 s per step (mean 0.24 s on TACRED, 0.20 s on FewRel), against 0.16–0.20 s (mean 0.18 and 0.16 s) for plain fine-tuning of the same model on the same GPU (the same training loop with the hippocampus switched off, one run per benchmark). At 8B, 0.32–0.71 s per step.
\item[\textbullet] \textbf{Sleep.} 15–38 s per task at 1B (TACRED and FewRel); at 8B (Llama-3.1-8B) 70–95 s per task on TACRED and 63–139 s per CLINC150 task. It never needs more GPU memory than training.
\item[\textbullet] \textbf{Subspace memory.} 1.96–2.09 GB at 1B (TACRED, FewRel); at 8B (Llama-3.1-8B) 11.2–11.5 GB after TACRED and 23.8–26.2 GB after CLINC150 (fp32 on the host, bf16 copy on the device).
\item[\textbullet] \textbf{Inference.} The plain model; the memory is not consulted.
\end{itemize}

\textbf{Why the step is cheap.}

\begin{itemize}[leftmargin=1.4em,itemsep=1pt,topsep=2pt,parsep=0pt]
\item[\textbullet] The gradient constraint is applied to activations (batch rows $\times\, d \times r$) rather than to weights.
\item[\textbullet] The update constraint is two low-rank matrix products per synapse group.
\item[\textbullet] No $d \times d$ matrix survives a sleep.
\item[\textbullet] Evaluation runs without the memory.
\end{itemize}

Supplementary Table 10 profiles one run (Llama-3.2-1B, TACRED, 8 tasks; seed 42; NVIDIA H100).

\textbf{What the memory holds} (Llama-3.2-1B, FewRel, at the start of the last task): the subspace bases 1.42 GiB fp32 on the host and 0.72 GiB bf16 on the device; class patterns 0.64 GiB; class-level state records 7 MiB; gain and embedding statistics about 2 MiB. The steady footprint is 1.06 GiB on the device and 1.74 GiB on the host.

A training step costs 25–33\% more than plain fine-tuning and inference is unchanged; the main added cost is the sleep after each task.

\subsection{Benchmarks, models and baselines}

Text benchmarks: TACRED\textsuperscript{4} and FewRel\textsuperscript{41} for relation extraction, and Banking77\textsuperscript{42} and CLINC150\textsuperscript{43} for intent detection. Vision benchmarks: CIFAR-100\textsuperscript{44} and ImageNet-R\textsuperscript{45}. Models: Qwen3\textsuperscript{46}, Llama-3\textsuperscript{47}, BERT\textsuperscript{48}, RoBERTa\textsuperscript{49}, ViT-B/16\textsuperscript{50} and ResNet-50\textsuperscript{51}; Supplementary Table 11 summarises the benchmarks.

\begin{itemize}[leftmargin=1.4em,itemsep=1pt,topsep=2pt,parsep=0pt]
\item[\textbullet] \textbf{LLMs:} Qwen3-0.6B, 4B and 8B; Llama-3.2-1B and 3B; Llama-3.1-8B. Learning rate 4e-5 (2e-5 × $\sqrt{32/8}$, the square-root batch rule), batch 32, 3 epochs per task, seeds 42, 11 and 3, the same code and configuration on every model (8B: the width rule; Qwen3-4B and both 8B models: gradient checkpointing, which changes memory only). Mean ± sample standard deviation over three runs.
\item[\textbullet] \textbf{Encoders:} BERT-base and RoBERTa-base on the four text benchmarks. Learning rate 4e-5 (heads 1e-3), batch 32, 3 epochs per task. Mean ± sample standard deviation over three runs (seeds 42, 11, 3).
\item[\textbullet] \textbf{Vision:} ViT-B/16 (ImageNet-21k) and ResNet-50 on CIFAR-100 and ImageNet-R, class-incremental with 10 tasks and no task id. Batch 64, 3 epochs per task, backbone learning rate 3e-5 (ViT) and 1e-4 (ResNet-50), heads 1e-3. Mean ± sample standard deviation over three runs (seeds 1993, 42, 3).
\item[\textbullet] \textbf{Metric:} final average accuracy (AA) over all tasks after the last task.
\item[\textbullet] \textbf{Baselines:}
\begin{itemize}[leftmargin=1.2em,itemsep=1pt,topsep=2pt,parsep=0pt]
\item[$\circ$] LLMs: SEQ (sequential fine-tuning, same constraints as ours); SEQ*\textsuperscript{5} (backbone adapted for one epoch on the first task and then frozen, cosine classifier with prototype prediction, the official configuration; a frozen-backbone method with looser constraints than ours); L2KD\textsuperscript{10} (teacher distillation); LAMOL\textsuperscript{8} and PCLL\textsuperscript{9} (generative replay with a teacher); Sculpting Subspaces\textsuperscript{30}, with a fixed budget protecting a fraction $t/n$ of each matrix's singular directions at task $t$ ($t = 0, \dots, n - 1$); and CODE-CL\textsuperscript{31} in its form without task-specific mixing matrices ($K = 0$), which needs no task identity at test time. The two projection methods use the same optimizer, batch and epochs as DeltaHippo with their own learning rate (2e-5), and CODE-CL's projection acts on the AdamW step. SEQ, SEQ*, L2KD, LAMOL and PCLL keep their own learning rate of 1e-5 (batch 32, 3 epochs per task). L2KD, LAMOL, PCLL, Sculpting Subspaces and CODE-CL are evaluated on Qwen3-0.6B and Llama-3.2-1B; from 3B on, the comparison is with SEQ and SEQ*.
\item[$\circ$] Encoders and vision: SEQ, and DER++\textsuperscript{6} and CLS-ER\textsuperscript{7} with a buffer of one stored sample per class; SEQ* adapts the backbone on the first task and then freezes it (a frozen-backbone baseline with looser constraints than ours). On the encoders it trains a linear head with the same one-sample-per-class buffer, and on the vision models a cosine classifier. The encoder baselines use learning rate 1e-5, batch 32 and 3 epochs per task. The replay baselines keep samples (with logits or averaged model copies) and DeltaHippo keeps geometric statistics, so these comparisons are by what each method may keep, not by memory size.
\item[$\circ$] ViT-B/16 additionally: EASE\textsuperscript{13} and SD-LoRA\textsuperscript{52}, which train adapters or LoRA modules on a frozen pre-trained backbone (frozen-backbone regime).
\end{itemize}
\end{itemize}

\subsection{Analysis protocol}

The analyses examine why the method works. Unless stated otherwise the testbed is Llama-3.2-1B on TACRED, the representative setting where both retention and acquisition are visible. All LLM interventions use the method's code with deterministic kernels, and every condition is identical to the normal run, including its random draws, up to the point where it is changed.

\subsection{Correspondence with hippocampal accounts}

The central computational hypothesis is \emph{consolidation by plasticity indexing}. The cortex (the network) holds long-term knowledge and keeps learning; the hippocampus (a training-only memory written by a gated-delta rule) encodes each task at sleep and, when read during later learning (wake), governs how the cortex is updated. It never reinstates content into the cortex. Supplementary Table 12 groups the accounts by the overall division of labour, the three functions of the Introduction (index, regulation, conflict) and memory capacity, and lists, for each, the quantity that plays the corresponding role and the evidence for it in this study. The mappings are functional, not anatomical.

\subsection*{Data availability}

TACRED is available from the Linguistic Data Consortium (LDC2018T24) under its licence; FewRel, Banking77, CLINC150, CIFAR-100 and ImageNet-R are available from their public releases. The class-incremental splits and task orders are given in Supplementary Note 2 and are released with the code.

\subsection*{Code availability}

The code for DeltaHippo, with its run configurations and our implementations of Sculpting Subspaces and CODE-CL, is available at \url{https://github.com/kkt94/deltahippo}; the version used for this paper is tagged v1.0 (commit 7b9b0a9).

\section*{Acknowledgements}

This work was supported by the Institute of Information \& Communications Technology Planning \& Evaluation (IITP) grant (No. RS-2025-25422680, No. RS-2020-II201373), and the National Research Foundation of Korea (NRF) grant (No. RS-2025-00520618) funded by the Korean Government (MSIT).

\section*{Author contributions}

K.K. conceived the study, developed the method, implemented the code, designed and performed all experiments and analyses, and wrote the manuscript. Y.S.C. supervised the project and reviewed the manuscript.

\section*{Competing interests}

The authors declare no competing interests.
}

\section*{References}
{\fontsize{7.5bp}{9bp}\selectfont
\begin{itemize}[leftmargin=2.2em,labelsep=0.5em,itemsep=0.5pt,topsep=2pt,parsep=0pt]
\item[1.] McCloskey, M. \& Cohen, N. J. Catastrophic interference in connectionist networks: the sequential learning problem. \emph{Psychol. Learn. Motiv.} \textbf{24}, 109–165 (1989).
\item[2.] French, R. M. Catastrophic forgetting in connectionist networks. \emph{Trends Cogn. Sci.} \textbf{3}, 128–135 (1999).
\item[3.] Kirkpatrick, J. et al. Overcoming catastrophic forgetting in neural networks. \emph{Proc. Natl Acad. Sci. USA} \textbf{114}, 3521–3526 (2017).
\item[4.] Zhang, Y., Zhong, V., Chen, D., Angeli, G. \& Manning, C. D. Position-aware attention and supervised data improve slot filling. In \emph{Proc. Conference on Empirical Methods in Natural Language Processing} (2017).
\item[5.] Zheng, J., Qiu, S. \& Ma, Q. Learn or recall? Revisiting incremental learning with pre-trained language models. In \emph{Proc. Annual Meeting of the Association for Computational Linguistics} (2024).
\item[6.] Buzzega, P., Boschini, M., Porrello, A., Abati, D. \& Calderara, S. Dark experience for general continual learning: a strong, simple baseline. In \emph{Advances in Neural Information Processing Systems} (2020).
\item[7.] Arani, E., Sarfraz, F. \& Zonooz, B. Learning fast, learning slow: a general continual learning method based on complementary learning system. In \emph{Proc. International Conference on Learning Representations} (2022).
\item[8.] Sun, F.-K., Ho, C.-H. \& Lee, H.-Y. LAMOL: Language modeling for lifelong language learning. In \emph{Proc. International Conference on Learning Representations} (2020).
\item[9.] Zhao, Y. et al. Prompt conditioned VAE: enhancing generative replay for lifelong learning in task-oriented dialogue. In \emph{Proc. Conference on Empirical Methods in Natural Language Processing} (2022).
\item[10.] Chuang, Y.-S., Su, S.-Y. \& Chen, Y.-N. Lifelong language knowledge distillation. In \emph{Proc. Conference on Empirical Methods in Natural Language Processing} (2020).
\item[11.] Wang, Z. et al. Learning to prompt for continual learning. In \emph{Proc. IEEE/CVF Conference on Computer Vision and Pattern Recognition} (2022).
\item[12.] Zhou, D.-W., Cai, Z.-W., Ye, H.-J., Zhan, D.-C. \& Liu, Z. Revisiting class-incremental learning with pre-trained models: generalizability and adaptivity are all you need (SimpleCIL / APER). \emph{Int. J. Comput. Vis.} (2024).
\item[13.] Zhou, D.-W., Sun, H.-L., Ye, H.-J. \& Zhan, D.-C. Expandable subspace ensemble for pre-trained model-based class-incremental learning. In \emph{Proc. IEEE/CVF Conference on Computer Vision and Pattern Recognition} (2024).
\item[14.] Liang, Y.-S. \& Li, W.-J. InfLoRA: Interference-free low-rank adaptation for continual learning. In \emph{Proc. IEEE/CVF Conference on Computer Vision and Pattern Recognition} (2024).
\item[15.] McDonnell, M. D., Gong, D., Parvaneh, A., Abbasnejad, E. \& van den Hengel, A. RanPAC: Random projections and pre-trained models for continual learning. In \emph{Advances in Neural Information Processing Systems} (2023).
\item[16.] Zeng, G., Chen, Y., Cui, B. \& Yu, S. Continual learning of context-dependent processing in neural networks. \emph{Nat. Mach. Intell.} \textbf{1}, 364–372 (2019).
\item[17.] Saha, G., Garg, I. \& Roy, K. Gradient projection memory for continual learning. In \emph{Proc. International Conference on Learning Representations} (2021).
\item[18.] Wang, S., Li, X., Sun, J. \& Xu, Z. Training networks in null space of feature covariance for continual learning. In \emph{Proc. IEEE/CVF Conference on Computer Vision and Pattern Recognition} (2021).
\item[19.] McClelland, J. L., McNaughton, B. L. \& O'Reilly, R. C. Why there are complementary learning systems in the hippocampus and neocortex. \emph{Psychol. Rev.} \textbf{102}, 419–457 (1995).
\item[20.] Kumaran, D., Hassabis, D. \& McClelland, J. L. What learning systems do intelligent agents need? \emph{Trends Cogn. Sci.} \textbf{20}, 512–534 (2016).
\item[21.] Teyler, T. J. \& DiScenna, P. The hippocampal memory indexing theory. \emph{Behav. Neurosci.} \textbf{100}, 147–154 (1986).
\item[22.] Teyler, T. J. \& Rudy, J. W. The hippocampal indexing theory and episodic memory: updating the index. \emph{Hippocampus} \textbf{17}, 1158–1169 (2007).
\item[23.] Abraham, W. C. \& Bear, M. F. Metaplasticity: the plasticity of synaptic plasticity. \emph{Trends Neurosci.} \textbf{19}, 126–130 (1996).
\item[24.] Fusi, S., Drew, P. J. \& Abbott, L. F. Cascade models of synaptically stored memories. \emph{Neuron} \textbf{45}, 599–611 (2005).
\item[25.] Kumaran, D. \& Maguire, E. A. Match–mismatch processes underlie human hippocampal responses to associative novelty. \emph{J. Neurosci.} \textbf{27}, 8517–8524 (2007).
\item[26.] Lisman, J. E. \& Grace, A. A. The hippocampal-VTA loop: controlling the entry of information into long-term memory. \emph{Neuron} \textbf{46}, 703–713 (2005).
\item[27.] Hasselmo, M. E. The role of acetylcholine in learning and memory. \emph{Curr. Opin. Neurobiol.} \textbf{16}, 710–715 (2006).
\item[28.] Gavish, M. \& Donoho, D. L. The optimal hard threshold for singular values is 4/$\surd$3. \emph{IEEE Trans. Inf. Theory} \textbf{60}, 5040–5053 (2014).
\item[29.] Yang, S., Kautz, J. \& Hatamizadeh, A. Gated delta networks: improving Mamba2 with delta rule. In \emph{Proc. International Conference on Learning Representations} (2025).
\item[30.] Nayak, N. S. et al. Sculpting subspaces: constrained full fine-tuning in LLMs for continual learning. Preprint at \url{https://arxiv.org/abs/2504.07097} (2025).
\item[31.] Apolinario, M. P. E., Choudhary, S. \& Roy, K. CODE-CL: Conceptor-based gradient projection for deep continual learning. In \emph{Proc. IEEE/CVF International Conference on Computer Vision} (2025).
\item[32.] McGeoch, J. A. \& McDonald, W. T. Meaningful relation and retroactive inhibition. \emph{Am. J. Psychol.} \textbf{43}, 579–588 (1931).
\item[33.] Tulving, E. \& Pearlstone, Z. Availability versus accessibility of information in memory for words. \emph{J. Verbal Learn. Verbal Behav.} \textbf{5}, 381–391 (1966).
\item[34.] Zheng, J., Cai, X., Qiu, S. \& Ma, Q. Spurious forgetting in continual learning of language models. In \emph{Proc. International Conference on Learning Representations} (2025).
\item[35.] Palit, V., Draye, F., Zucchet, N., Jin, Z. \& Schölkopf, B. When forgetting is not catastrophic: on the mechanics of spurious forgetting. Preprint at \url{https://arxiv.org/abs/2610.08718} (2026).
\item[36.] Yang, G. et al. Tensor Programs V: Tuning large neural networks via zero-shot hyperparameter transfer. Preprint at \url{https://arxiv.org/abs/2203.03466} (2022).
\item[37.] Diekelmann, S. \& Born, J. The memory function of sleep. \emph{Nat. Rev. Neurosci.} \textbf{11}, 114–126 (2010).
\item[38.] Rasch, B. \& Born, J. About sleep's role in memory. \emph{Physiol. Rev.} \textbf{93}, 681–766 (2013).
\item[39.] Widrow, B. \& Hoff, M. E. Adaptive switching circuits. In \emph{IRE WESCON Convention Record} Part 4, 96–104 (1960).
\item[40.] Benna, M. K. \& Fusi, S. Computational principles of synaptic memory consolidation. \emph{Nat. Neurosci.} \textbf{19}, 1697–1706 (2016).
\item[41.] Han, X. et al. FewRel: a large-scale supervised few-shot relation classification dataset with state-of-the-art evaluation. In \emph{Proc. Conference on Empirical Methods in Natural Language Processing} (2018).
\item[42.] Casanueva, I., Temčinas, T., Gerz, D., Henderson, M. \& Vulić, I. Efficient intent detection with dual sentence encoders. In \emph{Proc. 2nd Workshop on Natural Language Processing for Conversational AI} (2020).
\item[43.] Larson, S. et al. An evaluation dataset for intent classification and out-of-scope prediction. In \emph{Proc. Conference on Empirical Methods in Natural Language Processing and the International Joint Conference on Natural Language Processing} (2019).
\item[44.] Krizhevsky, A. Learning multiple layers of features from tiny images. Technical report, Univ. Toronto (2009).
\item[45.] Hendrycks, D. et al. The many faces of robustness: a critical analysis of out-of-distribution generalization. In \emph{Proc. IEEE/CVF International Conference on Computer Vision} (2021).
\item[46.] Yang, A. et al. Qwen3 technical report. Preprint at \url{https://arxiv.org/abs/2505.09388} (2025).
\item[47.] Grattafiori, A. et al. The Llama 3 herd of models. Preprint at \url{https://arxiv.org/abs/2407.21783} (2024).
\item[48.] Devlin, J., Chang, M.-W., Lee, K. \& Toutanova, K. BERT: pre-training of deep bidirectional transformers for language understanding. In \emph{Proc. Conference of the North American Chapter of the Association for Computational Linguistics} (2019).
\item[49.] Liu, Y. et al. RoBERTa: a robustly optimized BERT pretraining approach. Preprint at \url{https://arxiv.org/abs/1907.11692} (2019).
\item[50.] Dosovitskiy, A. et al. An image is worth 16x16 words: transformers for image recognition at scale. In \emph{Proc. International Conference on Learning Representations} (2021).
\item[51.] He, K., Zhang, X., Ren, S. \& Sun, J. Deep residual learning for image recognition. In \emph{Proc. IEEE/CVF Conference on Computer Vision and Pattern Recognition} (2016).
\item[52.] Wu, Y. et al. SD-LoRA: Scalable decoupled low-rank adaptation for class incremental learning. In \emph{Proc. International Conference on Learning Representations} (2025).
\end{itemize}
}

\clearpage
\setcounter{table}{0}\setcounter{figure}{0}
\renewcommand{\tablename}{Extended Data Table}
\renewcommand{\figurename}{Extended Data Fig.}
\begin{figure*}[!tp]
\centering
\includegraphics[width=1.000\textwidth]{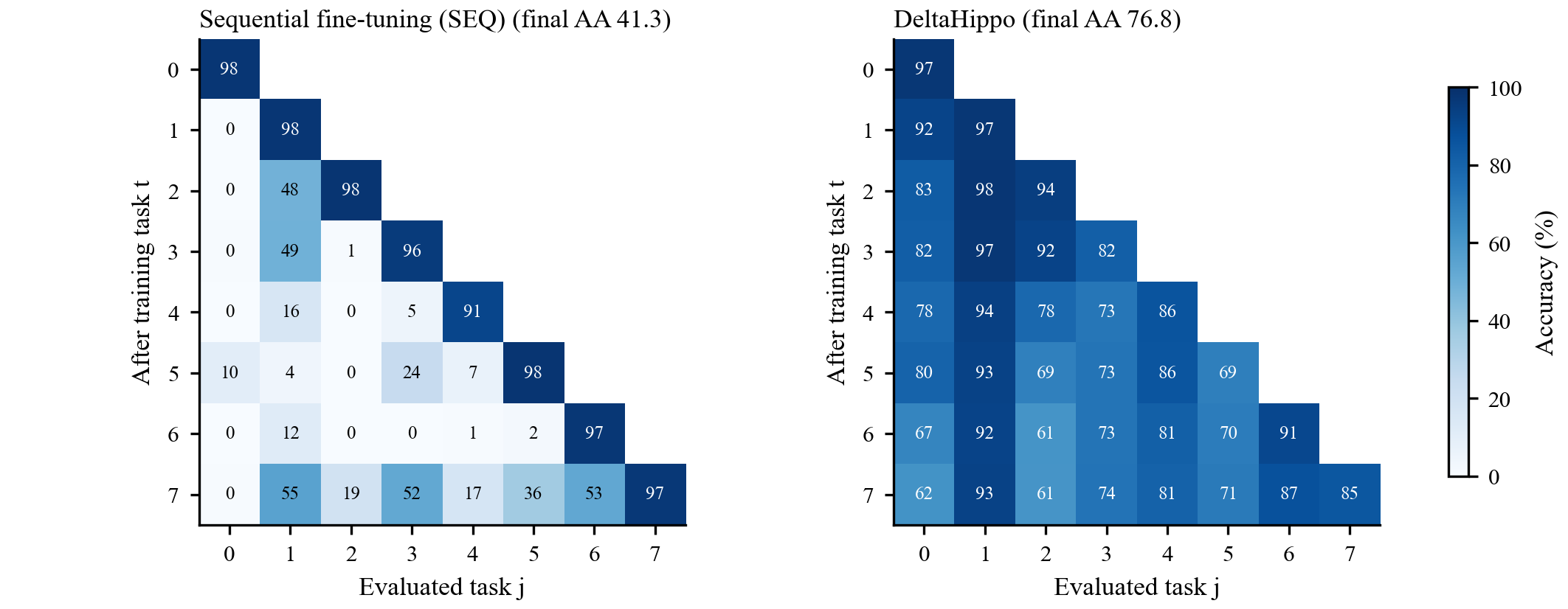}\par\vspace{2pt}
\caption{\textbf{Accuracy matrices of sequential fine-tuning and DeltaHippo (Llama-3.1-8B, TACRED; seed 42).} Entry $(t, j)$ is the test accuracy on task $j$ after training task $t$; the final average accuracy is the mean of the last row. Under sequential fine-tuning the earlier tasks fall sharply as later tasks are learnt (final AA 41.3); under DeltaHippo they stay in use while every new task is learnt by the same fully trained network (final AA 76.8).}
\end{figure*}

\begin{figure*}[!p]
\centering
\includegraphics[width=0.743\textwidth]{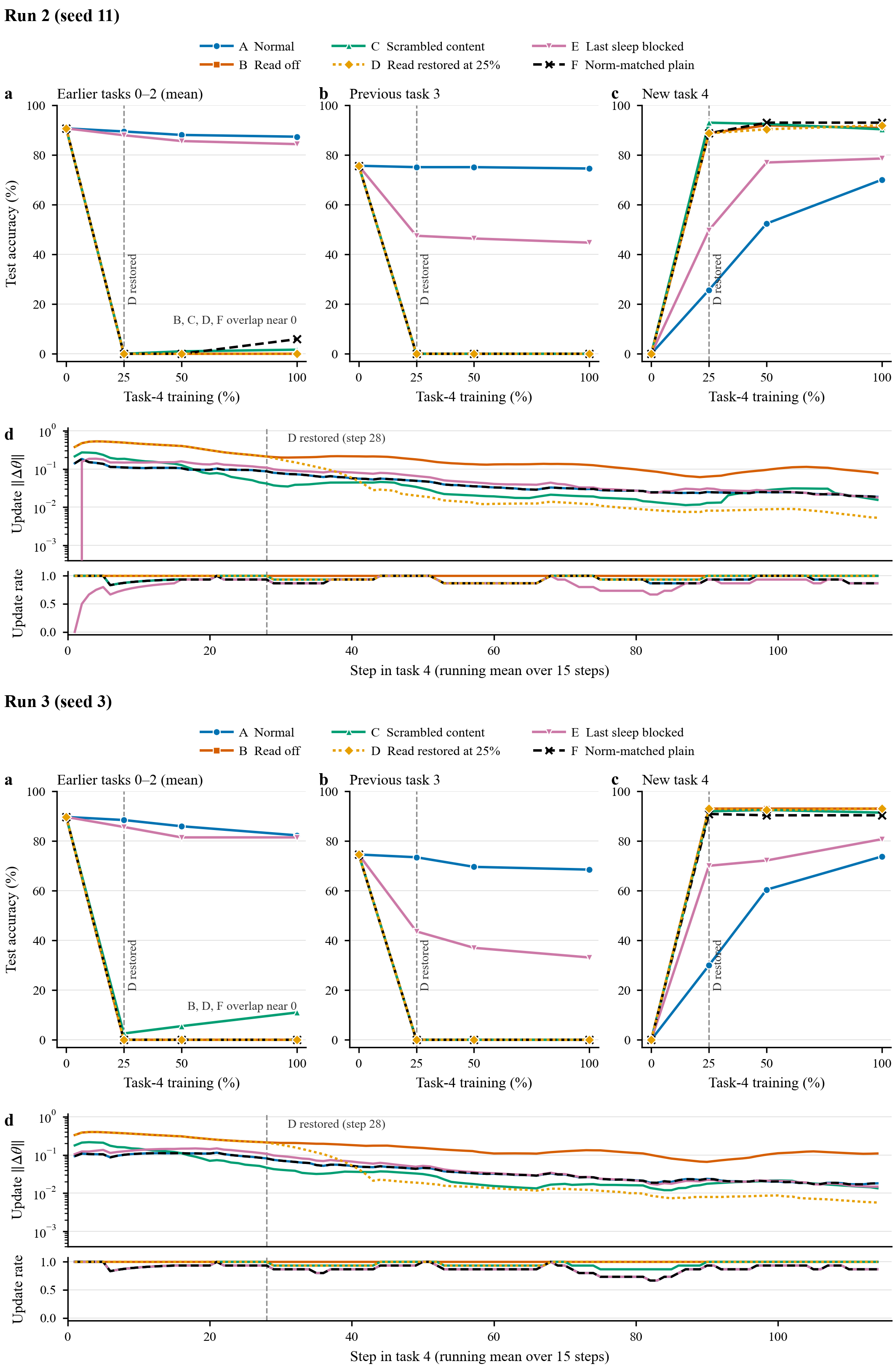}\par\vspace{2pt}
\caption{\textbf{Memory intervention at the transition to task 4, runs 2 and 3 (Llama-3.2-1B, TACRED; seeds 11 and 3).} Panels and conditions as in Fig. 3: \textbf{a–c}, the three task groups at 0, 25, 50 and 100\% of task 4; dashed line, D's restore; \textbf{d}, applied update norm and update rate per step.}
\end{figure*}

\begin{figure*}[!tp]
\centering
\includegraphics[width=1.000\textwidth]{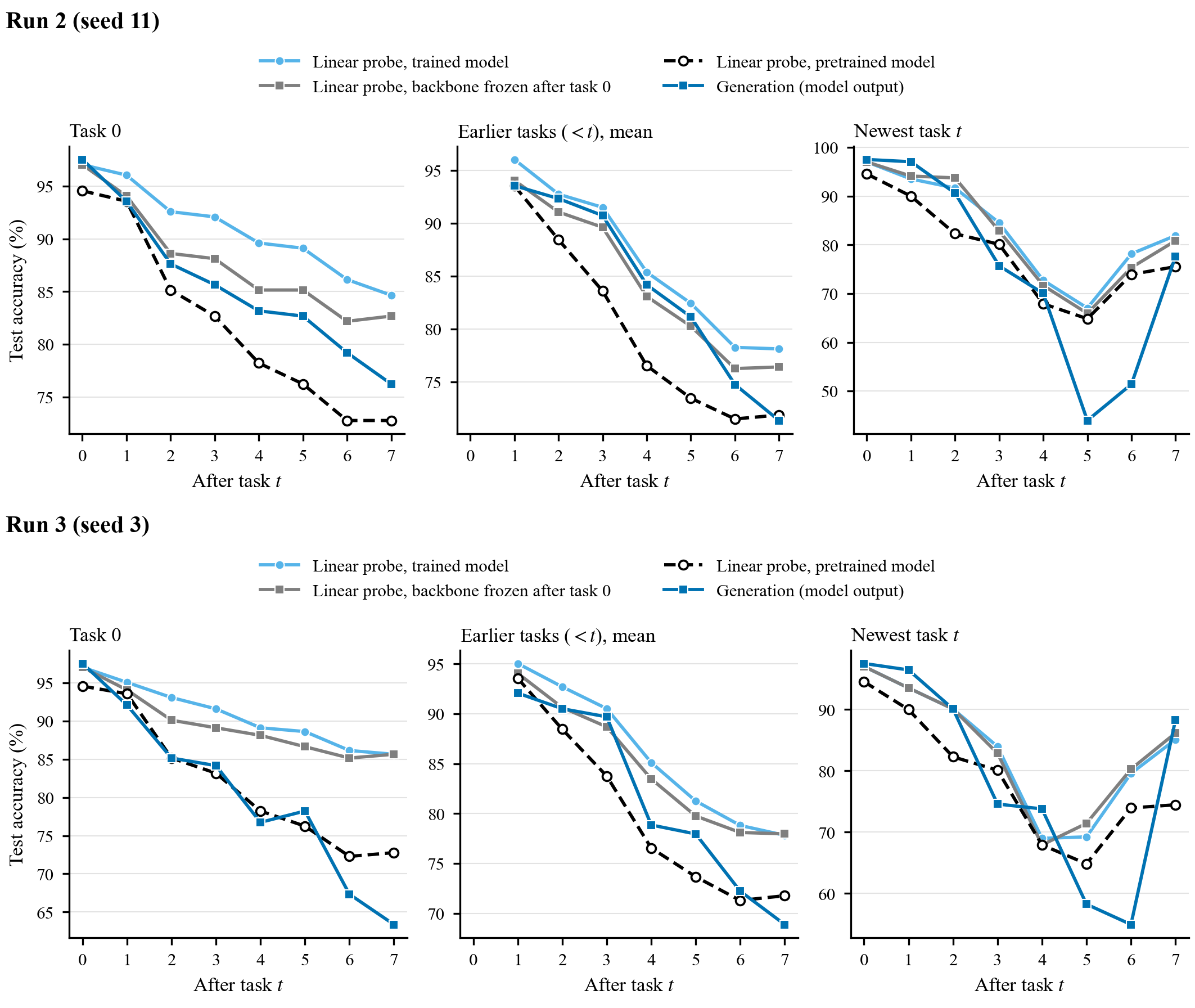}\par\vspace{2pt}
\caption{\textbf{Common linear probe, runs 2 and 3 (Llama-3.2-1B, TACRED; seeds 11 and 3).} As in Fig. 5c: a linear probe on the final-layer state before the first answer token of the trained model, of a backbone frozen after task 0 and of the pre-trained model, with the model's own generation; task 0, the mean of the earlier tasks and the newest task after each task.}
\end{figure*}

\begin{table*}[!tp]
\caption{\textbf{Text encoders: final average accuracy (\%).} Mean ± sample standard deviation over three runs (seeds 42, 11, 3; batch 32); SEQ*, DER++ and CLS-ER keep a buffer of one sample per class. On the encoders SEQ* trains a linear head (without a buffer it falls to the level of SEQ, 14–27). Bold, best in the model's column.}
{\tabsize\setlength{\tabcolsep}{3pt}\begin{tabular*}{\textwidth}{@{\extracolsep{\fill}}lllllll@{}}
\toprule
\textbf{Model} & \textbf{Method} & \textbf{Relies on} & \textbf{TACRED} & \textbf{FewRel} & \textbf{Banking77} & \textbf{CLINC150}\\
\midrule
BERT & SEQ & none (full FT) & 18.04 ± 0.26 & 17.40 ± 1.39 & 20.45 ± 0.17 & 14.97 ± 0.75\\
 & SEQ* & frozen backbone + replay buffer (samples) & 30.01 ± 1.23 & 48.18 ± 2.79 & 58.45 ± 2.28 & 62.50 ± 1.68\\
 & DER++ & replay buffer (samples + logits) & 42.01 ± 0.89 & 50.57 ± 1.79 & 54.29 ± 0.68 & 75.22 ± 1.29\\
 & CLS-ER & replay buffer + averaged model copies & 44.97 ± 3.76 & 52.34 ± 1.77 & 59.44 ± 0.73 & 75.25 ± 1.76\\
 & \textbf{DeltaHippo} & \textbf{training-only geometric memory; full FT} & \textbf{62.62 ± 0.63} & \textbf{66.99 ± 2.94} & \textbf{85.02 ± 0.20} & \textbf{90.42 ± 0.35}\\
RoBERTa & SEQ & none (full FT) & 14.61 ± 0.96 & 15.45 ± 1.62 & 19.72 ± 1.12 & 14.44 ± 0.52\\
 & SEQ* & frozen backbone + replay buffer (samples) & 27.75 ± 2.25 & 49.05 ± 1.83 & 65.96 ± 0.53 & 69.13 ± 1.81\\
 & DER++ & replay buffer (samples + logits) & 43.65 ± 0.40 & 47.88 ± 1.88 & 61.42 ± 0.56 & 80.50 ± 1.44\\
 & CLS-ER & replay buffer + averaged model copies & 45.72 ± 2.39 & 51.79 ± 1.80 & 66.68 ± 1.50 & 81.81 ± 0.39\\
 & \textbf{DeltaHippo} & \textbf{training-only geometric memory; full FT} & \textbf{53.24 ± 7.89} & \textbf{71.55 ± 3.84} & \textbf{84.34 ± 1.21} & \textbf{91.89 ± 0.36}\\
\bottomrule
\end{tabular*}}
\end{table*}

\begin{table*}[!tp]
\caption{\textbf{Vision models: final average accuracy (\%).} Ten tasks; mean ± sample standard deviation over three runs (seeds 1993, 42, 3; batch 64). DER++ and CLS-ER keep one image per class. Bold, best in the model's column.}
{\tabsize\setlength{\tabcolsep}{3pt}\begin{tabular*}{\textwidth}{@{\extracolsep{\fill}}lllll@{}}
\toprule
\textbf{Model} & \textbf{Method} & \textbf{Relies on} & \textbf{CIFAR-100} & \textbf{ImageNet-R}\\
\midrule
ViT-B/16 & SEQ & none (full FT) & 21.25 ± 0.85 & 23.24 ± 0.60\\
 & SEQ* & frozen backbone & 81.52 ± 0.34 & 64.23 ± 0.34\\
 & DER++ & replay buffer (samples + logits) & 45.83 ± 1.90 & 42.93 ± 1.17\\
 & CLS-ER & replay buffer + averaged model copies & 80.42 ± 1.03 & 51.79 ± 0.70\\
 & EASE & frozen backbone + expandable adapters & 87.02 ± 0.03 & \textbf{67.81 ± 0.17}\\
 & SD-LoRA & frozen backbone + LoRA modules & 87.47 ± 0.19 & 66.57 ± 0.09\\
 & \textbf{DeltaHippo} & \textbf{training-only geometric memory; full FT} & \textbf{88.21 ± 0.25} & 64.17 ± 0.26\\
ResNet-50 & SEQ & none (full FT) & 9.08 ± 0.19 & 8.55 ± 0.24\\
 & SEQ* & frozen backbone & 46.84 ± 1.69 & 40.61 ± 0.75\\
 & DER++ & replay buffer (samples + logits) & 18.30 ± 0.57 & 23.82 ± 0.05\\
 & CLS-ER & replay buffer + averaged model copies & 54.60 ± 0.95 & 40.09 ± 0.42\\
 & \textbf{DeltaHippo} & \textbf{training-only geometric memory; full FT} & \textbf{68.20 ± 0.72} & \textbf{61.31 ± 0.62}\\
\bottomrule
\end{tabular*}}
\end{table*}

\begin{table*}[!tp]
\caption{\textbf{Same-head control, ViT-B/16 on ImageNet-R.} Mean ± sample standard deviation over three runs.}
{\tabsize\setlength{\tabcolsep}{3pt}\begin{tabular*}{\textwidth}{@{\extracolsep{\fill}}lll@{}}
\toprule
\textbf{} & \textbf{Final AA} & \textbf{Mean acquisition, tasks 1–9}\\
\midrule
DeltaHippo (backbone learns) & \textbf{64.17 ± 0.26} & \textbf{60.0}\\
Backbone frozen after task 0, same classifier rule & 53.25 ± 0.84 & 46.2\\
\bottomrule
\end{tabular*}}
\end{table*}

\begin{table*}[!tp]
\caption{\textbf{The core of the operator, removed one part at a time, and the class-free core on its own.} Llama-3.2-1B, TACRED; mean ± sample standard deviation of final AA over three runs per row (seeds 42, 11 and 3), acquisition and retention as means over the same runs. Acquisition, mean accuracy on each task right after it is learnt; retention, mean over $t = 1$–7 of the mean accuracy on the tasks before $t$.}
{\tabsize\setlength{\tabcolsep}{3pt}\begin{tabularx}{\textwidth}{@{}>{\raggedright\arraybackslash}Xlll@{}}
\toprule
\textbf{Variant} & \textbf{Final AA} & \textbf{Mean acquisition} & \textbf{Retention}\\
\midrule
DeltaHippo & 71.58 ± 0.49 & 77.9 & 82.1\\
Class-free geometric core only (Box 1: protected subspace constrained exactly, residual floor outside; no class records, class-pattern constraint, conflict detection and update gate, comparator, joint-share target or erase) & 67.07 ± 2.16 & 74.4 & 80.9\\
No update constraint (gradient constrained only) & 62.14 ± 2.16 & 87.3 & 60.3\\
Hard constraint inside the protected subspace, residual plasticity factor kept & 71.07 ± 1.45 & 75.9 & 82.9\\
Graded inside the protected subspace, residual subspace fully open & 71.18 ± 2.45 & 85.3 & 79.2\\
Operator protects only the top 25\% of the stored directions, residual floor kept & 68.67 ± 2.06 & 77.5 & 79.7\\
Operator protects only the top 25\% of the stored directions, residual subspace fully open & 49.77 ± 5.27 & 91.4 & 53.0\\
No gradient constraint (update constraint only) & 56.42 ± 13.06 & 90.7 & 53.1\\
\bottomrule
\end{tabularx}}
\end{table*}

\begin{table*}[!tp]
\caption{\textbf{The two constraints and the residual floor in the encoder, convolutional and vision-transformer families.} Final average accuracy (\%), mean acquisition / retention in brackets (as in Extended Data Table 4). Each column removes one part of the full method. BERT-base and ResNet-50, one run; ViT-B/16, mean ± sample standard deviation over three runs. In ViT-B/16 only the update constraint was removed (–, not evaluated).}
{\tabsize\setlength{\tabcolsep}{3pt}\begin{tabular*}{\textwidth}{@{\extracolsep{\fill}}lllll@{}}
\toprule
\textbf{Model, benchmark} & \textbf{Full method} & \textbf{No update constraint} & \textbf{No gradient constraint} & \textbf{Residual subspace open}\\
\midrule
BERT-base, TACRED & 62.69 (63.4 / 78.3) & 59.59 (85.4 / 62.0) & 52.85 (63.4 / 71.3) & 46.84 (93.1 / 50.1)\\
ResNet-50, CIFAR-100 & 68.06 (79.2 / 73.7) & 37.51 (74.5 / 44.7) & 30.70 (44.7 / 44.8) & 27.11 (93.4 / 25.1)\\
ViT-B/16, CIFAR-100 & 88.21 ± 0.25 (94.0 / 91.6) & 71.19 ± 0.36 (98.1 / 79.3) & – & –\\
\bottomrule
\end{tabular*}}
\end{table*}

\begin{table*}[!tp]
\caption{\textbf{Near-duplicate classes on TACRED.} Llama-3.2-1B; mean ± sample standard deviation over three runs. Each value is the mean over the eight earlier classes (first two columns) or a share of all errors on the 15 relations that have a near-duplicate (last column). Bold, best in the column.}
{\tabsize\setlength{\tabcolsep}{3pt}\begin{tabularx}{\textwidth}{@{}l>{\raggedright\arraybackslash}X>{\raggedright\arraybackslash}X>{\raggedright\arraybackslash}X>{\raggedright\arraybackslash}X@{}}
\toprule
\textbf{Method} & \textbf{Relies on} & \textbf{Earlier classes answered correctly (\%)} & \textbf{Earlier classes absorbed by a later near-duplicate (\%)} & \textbf{Errors that name a near-duplicate (\%)}\\
\midrule
SEQ & none (full FT) & 36.6 ± 8.8 & 30.8 ± 3.7 & 31.7 ± 3.2\\
L2KD & teacher (distillation) & 0.9 ± 0.6 & 40.5 ± 0.8 & 28.6 ± 0.6\\
LAMOL & generative replay + teacher & 63.0 ± 2.5 & 11.3 ± 2.0 & 20.7 ± 2.2\\
PCLL & generative replay + teacher & 41.8 ± 8.2 & 29.7 ± 2.1 & 32.1 ± 1.0\\
Sculpting Subspaces & weights' singular subspaces; full FT & 24.8 ± 2.5 & 24.8 ± 4.8 & 22.4 ± 4.7\\
CODE-CL & stored input subspaces (conceptors); full FT & 50.0 ± 32.4 & 21.1 ± 14.2 & 22.3 ± 4.4\\
SEQ* & frozen backbone & 56.6 ± 1.3 & 9.0 ± 1.4 & 24.7 ± 3.7\\
\textbf{DeltaHippo} & \textbf{training-only geometric memory; full FT} & \textbf{76.6 ± 7.1} & \textbf{2.8 ± 2.1} & \textbf{11.6 ± 4.0}\\
\bottomrule
\end{tabularx}}
\end{table*}

\begin{table*}[!tp]
\caption{\textbf{Earlier near-duplicate classes on every decoder (TACRED).} Mean ± sample standard deviation over three runs. Earlier classes: the eight relations learnt before one of their near-duplicates (Supplementary Note 2); values are the mean over them. DeltaHippo is read from the Table 1 runs' per-class accuracies; SEQ* from per-sentence answers of three runs of the Table 1 configuration.}
{\tabsize\setlength{\tabcolsep}{3pt}\begin{tabular*}{\textwidth}{@{\extracolsep{\fill}}lll@{}}
\toprule
\textbf{Model} & \textbf{Earlier classes answered correctly, DeltaHippo (\%)} & \textbf{Earlier classes answered correctly, SEQ* (\%)}\\
\midrule
Qwen3-0.6B & \textbf{74.6 ± 2.0} & 54.6 ± 1.6\\
Llama-3.2-1B & \textbf{76.6 ± 7.1} & 56.6 ± 1.3\\
Qwen3-4B & \textbf{74.1 ± 7.0} & 62.1 ± 1.8\\
Llama-3.2-3B & \textbf{81.2 ± 4.4} & 51.1 ± 1.0\\
Qwen3-8B & \textbf{73.5 ± 2.4} & 54.9 ± 2.6\\
Llama-3.1-8B & \textbf{73.6 ± 6.9} & 64.5 ± 6.5\\
\bottomrule
\end{tabular*}}
\end{table*}

\clearpage
\setcounter{table}{0}\setcounter{figure}{0}
\renewcommand{\tablename}{Supplementary Table}\renewcommand{\figurename}{Supplementary Fig.}
\twocolumn[{\noindent{\sffamily\fontsize{18bp}{22bp}\bfseries\selectfont Supplementary Information\par}\vskip 14pt}]

\section*{Supplementary Note 1: Relation to prior work}

\textbf{Operational regimes.} The table places the main lines of work by where their stability comes from. Each is a useful choice in its own regime; the regime studied here excludes all of them.

\begin{table*}[!t]
{\tabsize\setlength{\tabcolsep}{3pt}\begin{tabularx}{\textwidth}{@{}>{\raggedright\arraybackslash}X>{\raggedright\arraybackslash}X>{\raggedright\arraybackslash}X>{\raggedright\arraybackslash}X@{}}
\toprule
\textbf{Regime} & \textbf{Approach} & \textbf{Where stability comes from} & \textbf{What it gives up}\\
\midrule
Keep old data & Replay of stored samples (DER++, CLS-ER) or generated pseudo-samples (LAMOL, PCLL) & old data or a generator & the right to forget data; for generative replay, a generator to train\\
Keep old models & Distillation from a teacher (L2KD; LAMOL and PCLL in our comparison) & a frozen copy of an earlier model & a full model copy kept beside the learner\\
Keep old models & Parameter anchors (EWC, SI) and averaged model copies (CLS-ER) & a stored copy of earlier parameters & a parameter copy; free movement of important weights\\
Do not change the model & Freezing the backbone and classifying frozen features with prototypes or analytic heads (SEQ*, which on the encoders also keeps a one-sample-per-class buffer; SimpleCIL / APER\textsuperscript{1}; RanPAC\textsuperscript{2}) & the pre-trained features, untouched & any representation learning after task 0\\
Do not change the model & Prompts (L2P\textsuperscript{3}; DualPrompt\textsuperscript{4}; CODA-Prompt\textsuperscript{5}; HiDe-Prompt\textsuperscript{6}; CPrompt\textsuperscript{7}), adapters (EASE\textsuperscript{8}; SSIAT\textsuperscript{9}; MOS\textsuperscript{10}; TUNA\textsuperscript{11}) or LoRA modules (InfLoRA\textsuperscript{12}; SD-LoRA\textsuperscript{13}; O-LoRA\textsuperscript{14}) on a frozen backbone & frozen backbone + learned prompts, adapters or low-rank modules & backbone plasticity; growth with tasks\\
Do not answer alone & Task identity or a memory at inference & task identity or an external module at test time & the model being the knowledge carrier\\
\bottomrule
\end{tabularx}}
\end{table*}

\begin{itemize}[leftmargin=1.4em,itemsep=1pt,topsep=2pt,parsep=0pt]
\item[\textbullet] \textbf{Regularisation and synaptic consolidation.} Elastic weight consolidation (EWC)\textsuperscript{15} and synaptic intelligence (SI)\textsuperscript{16} penalise changes to important weights, one importance per parameter. DeltaHippo instead sets plasticity per input direction of a synapse group, from the energy that earlier inputs carried along it.
\item[\textbullet] \textbf{Gradient projection.} Orthogonal weights modification (OWM)\textsuperscript{17}, orthogonal gradient descent (OGD)\textsuperscript{18}, gradient projection memory (GPM)\textsuperscript{19} and Adam-NSCL\textsuperscript{20} restrict updates according to past inputs or gradients: OWM with a regularised inverse of the input correlation, GPM with a hard null space of stored bases, SGP\textsuperscript{21} with gradient steps scaled along stored bases by their importance, and Adam-NSCL by projecting Adam's candidate update onto the approximate null space of the earlier feature covariance, with the classifier frozen after its task and EWC on BatchNorm parameters; its extension Adam-SFCL scales the update along all singular directions of that covariance instead of nulling a subspace\textsuperscript{22}. A recent per-unit projector protects the tokens each MLP unit fires on when fine-tuning language models\textsuperscript{23}. DeltaHippo differs from this line in five ways:
\begin{itemize}[leftmargin=1.2em,itemsep=1pt,topsep=2pt,parsep=0pt]
\item[$\circ$] a residual floor outside the stored span: the energy that compression leaves out keeps constraining updates there, and opening it loses retention (Extended Data Table 4);
\item[$\circ$] enforcement on both sides of the adaptive optimizer: the gradient is formed through the constrained inputs, shaping Adam's moments, and the realised Adam update is constrained again, in every synapse group, output head, tied input embedding matrix and normalisation gains included, all of which stay in training (Proposition 1); removing either constraint costs 9–15 points (Extended Data Table 4);
\item[$\circ$] the protected subspace is extended by class-level patterns (class-pattern constraint);
\item[$\circ$] the plasticity factors come from an explicit objective whose aperture is set per synapse group and per task from the episode's noise floor and the ratio of earlier to new experience, where SGP uses fixed importance scaling and CODE-CL a fixed hyperparameter;
\item[$\circ$] the memory is written by a gated delta rule, so it can erase as well as accumulate, erasing less where conflicts were seen (erase lowers the stored energy; Supplementary Table 1).
\end{itemize}
\item[\textbullet] \textbf{Recent subspace methods for full-model and pre-trained continual learning.} CODE-CL\textsuperscript{24} represents earlier feature spaces as conceptor matrices and projects gradients pseudo-orthogonally to them. Sculpting Subspaces\textsuperscript{25} keeps LLM fine-tuning full-rank but constrains each matrix's update to be orthogonal to the high-singular-value directions of the weights themselves. CoSO\textsuperscript{26} optimizes a pre-trained ViT in a sequence of gradient-SVD subspaces orthogonal to earlier ones. DeltaHippo instead keeps a residual floor outside the stored span, constrains both the gradient and the realised step in every synapse group of LLMs up to 8B, makes the base update exactly orthogonal to class patterns, sets its plasticity factors from the data, and gives a causal account of what the memory preserves (Results).
\item[\textbullet] \textbf{Spurious forgetting and alignment.} Zheng et al.\textsuperscript{27} show that performance drops in continual learning of language models often reflect a loss of task alignment rather than of underlying knowledge, and mitigate it by freezing lower layers. DeltaHippo keeps every parameter tensor learning and instead maintains that alignment with a training-only index, and Fig. 5 decomposes it causally: neither guiding only the backbone nor only the output head matches guiding both. Recent work traces spurious forgetting of facts to a common shift of old representations, separate from fact-specific erosion\textsuperscript{28}. DeltaHippo keeps access during learning rather than recovering it afterwards.
\item[\textbullet] \textbf{Geometry-aware parameter-efficient methods (2026).} PLATE\textsuperscript{29} builds a protected subspace from pre-trained weight geometry and trains structured low-rank adapters with reduced functional drift; CoRe\textsuperscript{30} intervenes in a low-rank subspace of hidden representations; SPACE-LoRA\textsuperscript{31} suppresses a LoRA branch along activation directions important for old tasks. DeltaHippo applies a past-activation metric to the gradient and the realised update of every parameter tensor of a fully trainable model instead.
\item[\textbullet] \textbf{Geometry and forgetting in LLM post-training.} Geometry Conflict\textsuperscript{32} explains forgetting as incompatibility between the covariance geometry of a task's parameter update and the evolving model state, and uses it for data-free model merging. DeltaHippo uses past-activation geometry online, to control the gradient and the realised update during sequential full-model learning, and shows causally that the direction of change, not its size, decides what is kept.
\item[\textbullet] \textbf{Probing analyses of forgetting.} Zheng et al.\textsuperscript{33} show that sequentially fine-tuned pre-trained language models keep much of their knowledge linearly decodable, and that observed forgetting follows the drift between class embeddings and features; this analysis also motivates SEQ*. Instead of freezing the backbone, DeltaHippo keeps that alignment while every parameter tensor stays in training, and Fig. 5 shows that the hippocampal memory is what keeps it.
\item[\textbullet] \textbf{Hippocampus-inspired continual learning.} HippoTune\textsuperscript{34} builds a query–retrieve–feedback loop inside each transformer layer, inspired by the entorhinal cortex–dentate gyrus–CA3–CA1 circuit, on top of parameter-efficient tuning; the hippocampal module runs at inference. HiCL\textsuperscript{35} is another hippocampus-inspired design that adds a hippocampal module to the network. In DeltaHippo the hippocampus is training-only and never computes an output.
\item[\textbullet] \textbf{How DeltaHippo differs from these lines of work.}
\end{itemize}

| Line of work | What it does | DeltaHippo | |---|---|---| | Input-space projection (OWM, GPM, SGP, Adam-NSCL, CODE-CL) | restricts updates by past feature geometry: a regularised inverse (OWM), a hard null space (GPM), importance-scaled projection (SGP), a null-space projection of Adam's update with a frozen classifier (Adam-NSCL), conceptors with a fixed aperture (CODE-CL) | a protected residual subspace outside the stored span; constraints on both the gradient and the realised update in every synapse group, output head, embedding matrix and gains included, none frozen; an exact class-pattern constraint on the low-energy directions that earlier class means occupy; per-direction plasticity factors from a memory-weighted output-change objective whose aperture is set from the data | | Full-model subspace fine-tuning of LLMs (Sculpting Subspaces) | keeps updates orthogonal to the weights' own top singular directions | protection defined by the inputs each synapse group served, i.e. by the function on past data, in every group including the output head, tied input embedding matrix and normalisation gains | | Pre-trained-model subspace optimization (CoSO) | optimizes in gradient-SVD subspaces orthogonal to earlier ones | no tensor or module is frozen; most directions keep a graded plasticity factor, and only the directions of earlier class means outside the stored span are hard-constrained | | Spurious forgetting\textsuperscript{27} | identifies loss of task alignment, not of knowledge, and freezes lower layers | maintains alignment while every tensor learns, and shows that joint guidance of backbone and output head keeps it best (Fig. 5) | | Geometry-aware parameter-efficient tuning (PLATE, CoRe, SPACE-LoRA) | constrains low-rank adapters or representation interventions | applies a residual-aware past-activation metric to the gradient and the realised update of the full model | | Probing analyses\textsuperscript{33} | show that knowledge stays decodable while the classifier–feature alignment drifts, and freeze the backbone (SEQ*) | keeps that alignment while every parameter tensor stays in training, and shows causally that the memory keeps it (Fig. 5) | | Hippocampus-inspired continual learning (HippoTune, HiCL) | adds a hippocampal module to the network or its inference path, on parameter-efficient tuning | a training-only hippocampus that never enters inference and guides full fine-tuning of every parameter | | Gated-delta memories (Gated DeltaNet) | a sequence-model layer | a sleep-written, gated-delta memory over synapse groups that governs plasticity |

\begin{itemize}[leftmargin=1.4em,itemsep=1pt,topsep=2pt,parsep=0pt]
\item[\textbullet] \textbf{Fast weights and gated-delta memories.} The delta rule\textsuperscript{36} underlies fast-weight views of linear attention\textsuperscript{37}, and Gated DeltaNet\textsuperscript{38} combines it with gating: a key–value association is written after erasing what was stored along the key. The hippocampal memory applies this read, write and erase mechanism to the second moment of each synapse group's inputs rather than to token keys and values. In place of Gated DeltaNet's global decay, content is kept wherever no new key overwrote it, and directions below the episode's noise floor are dropped.
\item[\textbullet] \textbf{Replay and distillation.} LAMOL\textsuperscript{39} and PCLL\textsuperscript{40} generate pseudo-samples; L2KD\textsuperscript{41} distils from a teacher. DER++\textsuperscript{42} replays buffered samples with their stored logits, and CLS-ER\textsuperscript{43} adds two exponentially averaged copies of the model as fast and slow memories. All of them relax our constraints.
\item[\textbullet] \textbf{Continual learning with frozen pre-trained backbones.} Most recent methods keep the pre-trained backbone frozen and learn prompts (L2P\textsuperscript{3}; DualPrompt\textsuperscript{4}; CODA-Prompt\textsuperscript{5}; HiDe- Prompt\textsuperscript{6}; CPrompt\textsuperscript{7}), adapters (EASE\textsuperscript{8}; SSIAT\textsuperscript{9}; MOS\textsuperscript{10}; TUNA\textsuperscript{11}) or LoRA modules (InfLoRA\textsuperscript{12}; SD-LoRA\textsuperscript{13}; O-LoRA\textsuperscript{14}), or classify frozen features with prototypes or analytic heads (SimpleCIL / APER\textsuperscript{1}; RanPAC\textsuperscript{2}); SLCA\textsuperscript{44} tunes every parameter slowly but aligns the classifier on features sampled from stored class statistics. The frozen-backbone methods avoid catastrophic forgetting by not continuing to train the original network; they are outside our constraints, which require every parameter to be fine-tuned; SLCA is outside them because it replays sampled features.
\item[\textbullet] \textbf{Frozen-backbone probes.} SEQ*\textsuperscript{33} adapts the backbone on the first task, then freezes it, trains a cosine classifier and predicts with class prototypes. It is a strong baseline of the frozen-backbone regime, which works under looser constraints than a fully fine-tuned learner.
\item[\textbullet] \textbf{Width-aware step sizes.} $\mu$P\textsuperscript{45} scales per-layer step sizes with width so that a step changes outputs by a width-independent amount. Our width rule scales the Adam step of each backbone matrix and of an untied LM head by $d_{\mathrm{ref}}/d_{\mathrm{hidden}}$ ($d_{\mathrm{ref}} = 2{,}560$) so that its effect on outputs does not grow with width.
\item[\textbullet] \textbf{Class-incremental theory.} Class-incremental learning is the hardest scenario\textsuperscript{46} and has been linked to out-of-distribution detection\textsuperscript{47}. Near-duplicate classes are where this detection is hardest: the new class's inputs fall inside an earlier class's region.
\item[\textbullet] \textbf{Neuroscience.} The accounts mapped in Methods (Correspondence with hippocampal accounts) are complementary learning systems, hippocampal indexing, sleep-dependent consolidation, cascade models of memory strength, metaplasticity, match–mismatch comparison, cholinergic gating, the CA1 comparator and capacity-driven forgetting.
\end{itemize}

\section*{Supplementary Note 2: Data and experimental settings}

\textbf{Benchmarks.} TACRED is distributed by the Linguistic Data Consortium under licence; the other benchmarks are publicly released. Text benchmarks follow the class-incremental protocol of Zheng, Qiu \& Ma\textsuperscript{33}: every class belongs to exactly one task, the class order is fixed, and the same order is used by every method.

\begin{table*}[!t]
{\tabsize\setlength{\tabcolsep}{3pt}\begin{tabularx}{\textwidth}{@{}lllll>{\raggedright\arraybackslash}X@{}}
\toprule
\textbf{Benchmark} & \textbf{Task} & \textbf{Classes} & \textbf{Tasks × classes} & \textbf{Train / test} & \textbf{Notes}\\
\midrule
TACRED\textsuperscript{48} & relation extraction & 40 & 8 × 5 & 7,391 / 1,259 & the 41 TACRED relation types except the rarest, per: country of death; no\_relation is not used; 20–320 training and 5–41 test sentences per relation\\
FewRel\textsuperscript{49} & relation extraction & 80 & 8 × 10 & 33,600 / 11,200 & 420 / 140 sentences per relation\\
Banking77\textsuperscript{50} & intent detection & 77 & 7 × 11 & 10,003 / 3,080 & 35–187 training and 40 test utterances per intent\\
CLINC150\textsuperscript{51} & intent detection & 150 & 15 × 10 & 15,000 / 4,500 & in-scope intents; 100 / 30 utterances per intent\\
CIFAR-100\textsuperscript{52} & image classification & 100 & 10 × 10 & 50,000 / 10,000 & images resized from 32 to 224 pixels (bilinear)\\
ImageNet-R\textsuperscript{53} & image classification & 200 & 10 × 20 & 80\% / 20\% per class & resize 256, centre crop 224\\
\bottomrule
\end{tabularx}}
\end{table*}

Vision class orders follow the PyCIL convention\textsuperscript{54} (a fixed random permutation with seed 1993); no training augmentation is used. Accuracy is measured on the test split after every task, on all tasks seen so far.

\textbf{Fixed protocol order.} Every benchmark is used with the class order and task split of its established class-incremental protocol\textsuperscript{33} (for the text benchmarks; the PyCIL order for the vision benchmarks), and every method, ours and each baseline, is run on exactly this sequence. The protocols fix the order so that results are comparable across methods and with published numbers. The fixed sequences already span 7 to 15 tasks and place near-duplicate relations both early and late (for example, TACRED tasks 0 and 7, and tasks 2 and 5; Fig. 4d–f). The margin holds under other orders: on Llama-3.2-1B TACRED with two further orders of the 40 relations (fixed random permutations, the same for both methods), DeltaHippo ends at 72.24 ± 2.99 and 72.62 ± 1.41 against 59.25 ± 0.81 and 54.77 ± 0.36 for SEQ* (three runs each), against 71.58 ± 0.49 and 54.95 ± 1.41 in the protocol order.

\textbf{TACRED task composition} (task index as in the main text; org = organization, per = person).

\begin{table*}[!t]
{\tabsize\setlength{\tabcolsep}{3pt}\begin{tabular*}{\textwidth}{@{\extracolsep{\fill}}ll@{}}
\toprule
\textbf{Task} & \textbf{Relations}\\
\midrule
0 & org: founded by; per: employee of; org: alternate names; per: city of residence; per: children\\
1 & per: title; per: siblings; per: religion; per: age; org: website\\
2 & per: state or provinces of residence; org: member of; org: top members/employees; per: countries of residence; org: city of headquarters\\
3 & org: members; org: country of headquarters; per: spouse; org: state or province of headquarters; org: number of employees/members\\
4 & org: parents; org: subsidiaries; per: origin; org: political/religious affiliation; per: other family\\
5 & per: state or province of birth; org: dissolved; per: date of death; org: shareholders; per: alternate names\\
6 & per: parents; per: schools attended; per: cause of death; per: city of death; per: state or province of death\\
7 & org: founded; per: country of birth; per: date of birth; per: city of birth; per: charges\\
\bottomrule
\end{tabular*}}
\end{table*}

\textbf{Near-duplicate relations (Extended Data Table 6).} Two relations are near-duplicates when their labels differ in a single word and are learnt in different tasks; the rule is applied to the label strings alone, without reference to model predictions, and compares the label after the entity-type prefix (person related / organization related), so a person's and an organisation's relation can be near-duplicates. It selects 15 relations (task in brackets): per: city of residence (0), per: employee of (0), org: city of headquarters (2), org: member of (2), per: countries of residence (2), org: country of headquarters (3), org: state or province of headquarters (3), per: date of death (5), per: state or province of birth (5), per: cause of death (6), per: city of death (6), per: state or province of death (6), per: city of birth (7), per: country of birth (7) and per: date of birth (7). The eight earlier classes, each learnt before at least one of its near-duplicates, and the later near-duplicates that can absorb them (task in brackets) are:

\begin{table*}[!t]
{\tabsize\setlength{\tabcolsep}{3pt}\begin{tabularx}{\textwidth}{@{}>{\raggedright\arraybackslash}X>{\raggedright\arraybackslash}X@{}}
\toprule
\textbf{Earlier class (task)} & \textbf{Later near-duplicates}\\
\midrule
per: city of residence (0) & per: countries of residence (2), org: city of headquarters (2), per: city of death (6), per: city of birth (7)\\
per: employee of (0) & org: member of (2)\\
org: city of headquarters (2) & org: country of headquarters (3), per: city of death (6), per: city of birth (7)\\
org: country of headquarters (3) & per: country of birth (7)\\
org: state or province of headquarters (3) & per: state or province of birth (5), per: state or province of death (6)\\
per: date of death (5) & per: cause of death (6), per: city of death (6), per: date of birth (7)\\
per: state or province of birth (5) & per: state or province of death (6)\\
per: city of death (6) & per: city of birth (7)\\
\bottomrule
\end{tabularx}}
\end{table*}

The first two measures of Extended Data Table 6 are computed per earlier class and then averaged over the eight classes (TACRED classes have 5–41 test sentences); the third pools all errors on the 15 relations. Counting singular and plural as the same word adds per: state or provinces of residence (2), the relation followed in Fig. 4d–f, giving 16 relations and ten earlier classes; DeltaHippo is then again best on all three measures of Extended Data Table 6 (earlier classes answered correctly 75.3\% against 0.7–53.7\%; absorbed 4.8\% against 10.7–42.2\%; errors naming a near-duplicate 25.7\% against 30.4–50.9\%; Llama-3.2-1B, three runs of every method).

\textbf{Prompts and labels (decoder LLMs).} Each label is its natural-language name (for TACRED, prefixed by "organization related:" or "person related:"), generated after the prompt and followed by the end-of-sequence token.

\begin{itemize}[leftmargin=1.4em,itemsep=1pt,topsep=2pt,parsep=0pt]
\item[\textbullet] Relation extraction: \texttt{Input sentence: \{sentence\}\textbackslash{}n The relationship between \{entity 1\} and \{entity 2\} is \{label\}} (entity markers removed from the sentence).
\item[\textbullet] Intent detection: \texttt{Input sentence: \{utterance\}\textbackslash{}n The label: \{label\}}.
\end{itemize}

At test time the model generates greedily (at most 10 new tokens); a prediction is correct when the generated label matches the gold label exactly.

\textbf{Models.}

\begin{table*}[!t]
{\tabsize\setlength{\tabcolsep}{3pt}\begin{tabular*}{\textwidth}{@{\extracolsep{\fill}}llllll@{}}
\toprule
\textbf{Model} & \textbf{Parameters} & \textbf{Layers} & \textbf{Hidden / MLP width} & \textbf{LM head} & \textbf{Checkpoint}\\
\midrule
Qwen3-0.6B\textsuperscript{55} & 0.6B & 28 & 1,024 / 3,072 & tied & Qwen/Qwen3-0.6B\\
Qwen3-4B & 4B & 36 & 2,560 / 9,728 & tied & Qwen/Qwen3-4B\\
Qwen3-8B & 8B & 36 & 4,096 / 12,288 & untied & Qwen/Qwen3-8B\\
Llama-3.2-1B\textsuperscript{56} & 1B & 16 & 2,048 / 8,192 & tied & meta-llama/Llama-3.2-1B\\
Llama-3.2-3B & 3B & 28 & 3,072 / 8,192 & tied & meta-llama/Llama-3.2-3B\\
Llama-3.1-8B & 8B & 32 & 4,096 / 14,336 & untied & meta-llama/Llama-3.1-8B\\
BERT-base\textsuperscript{57} & 110M & 12 & 768 / 3,072 & linear heads on [CLS] & google-bert/bert-base-uncased\\
RoBERTa-base\textsuperscript{58} & 125M & 12 & 768 / 3,072 & linear heads on [CLS] & FacebookAI/roberta-base\\
ViT-B/16\textsuperscript{59} & 86M & 12 & 768 / 3,072 & linear heads on [CLS] & google/vit-base-patch16-224-in21k\\
ResNet-50\textsuperscript{60} & 25M & 50 & 2,048 (pooled feature) & linear heads on the global pool & torchvision ResNet50\_Weights.IMAGENET1K\_V2\\
\bottomrule
\end{tabular*}}
\end{table*}

\textbf{Training.}

\begin{table*}[!t]
{\tabsize\setlength{\tabcolsep}{3pt}\begin{tabular*}{\textwidth}{@{\extracolsep{\fill}}lllll@{}}
\toprule
\textbf{} & \textbf{Decoder LLMs} & \textbf{Text encoders} & \textbf{ViT-B/16} & \textbf{ResNet-50}\\
\midrule
Optimizer & AdamW (0.9, 0.999), $\epsilon$ 1e-8, weight decay 5e-4 & same & same & same\\
Learning rate & 4e-5 (8B: width rule on top) & 4e-5 (heads 1e-3) & 3e-5 (heads 1e-3) & 1e-4 (heads 1e-3)\\
Batch / epochs per task & 32 / 3 & 32 / 3 & 64 / 3 & 64 / 3\\
Seeds & 42, 11, 3 & 42, 11, 3 & 1993, 42, 3 & 1993, 42, 3\\
Loss & answer-token cross-entropy & cross-entropy over all seen classes & same & same\\
Precision & fp32 weights and optimizer state, bf16 autocast & same & same & same\\
\bottomrule
\end{tabular*}}
\end{table*}

No learning-rate schedule, warm-up or gradient clipping is used, and no checkpoint is selected: every reported number is the model after the last task. The architecture rules of Methods (relative step; width rule at 8B; fp64 orthonormal basis on models whose widest synapse-group input exceeds 9,000) are fixed per model and used unchanged on every benchmark. Gradient checkpointing, which changes memory only, is used for Qwen3-4B and the 8B models. The 8B models run on one NVIDIA B200 GPU and all other models on one NVIDIA H100 80GB GPU. Data, task order, batch size, epoch budget, seeds, precision and evaluation are standardized for every method except EASE and SD-LoRA, which keep their original recipes. The decoder and encoder learning rate 4e-5 is the batch-8 rate 2e-5 scaled by $\sqrt{32/8}$ (square-root batch rule); each baseline keeps its own learning rate (SEQ, SEQ*, L2KD, LAMOL, PCLL and the encoder baselines 1e-5; Sculpting Subspaces and CODE-CL 2e-5).

\textbf{Metrics.} $A_{t,j}$ is the test accuracy on task $j$ after training task $t$. Final average accuracy (AA) is the mean of the last row; acquisition is the mean of $A_{t,t}$; retention is the mean over $t \ge 1$ of the mean accuracy on the tasks before $t$.

\section*{Supplementary Note 3: Baseline implementations}

\textbf{Sequential fine-tuning, SEQ\textbackslash{}\emph{, L2KD, LAMOL and PCLL} use the public implementation and configurations of Zheng, Qiu \& Ma\textsuperscript{33}. SEQ fine-tunes the whole model on each task in turn. SEQ\textbackslash{}} adapts the whole model for one epoch on the first task, then freezes the backbone and trains a cosine classifier, with earlier classes' rows frozen, and predicts with class prototypes (the official configuration). These five baselines use learning rate 1e-5, batch 32 and 3 epochs per task. L2KD distils from the previous model; LAMOL and PCLL generate pseudo-examples of earlier tasks and distil on them.

\textbf{Encoder and vision baselines.} DER++ and CLS-ER keep a replay buffer of one example per class. On the encoders SEQ\textbackslash{}* trains a linear head with the same one-sample-per-class buffer; on the vision models it uses a cosine classifier. The encoder baselines (SEQ, SEQ\textbackslash{}*, DER++ and CLS-ER) use learning rate 1e-5, batch 32 and 3 epochs per task. EASE and SD-LoRA train expandable adapters or LoRA modules on the frozen pre-trained ViT-B/16.

\textbf{Subspace-projection baselines.} Sculpting Subspaces\textsuperscript{25} and CODE-CL\textsuperscript{24} were implemented in the same training framework as DeltaHippo from the papers and their official code (github.com/Red-Hat-AI-Innovation-Team/orthogonal-subspace-learning at commit a5f91f6, with the maintained implementation of orthogonal subspace fine-tuning by Nayak et al. in their mini\_trainer repository at commit fd5b552; github.com/mapolinario94/CODE-CL at commit e6566f5). They share DeltaHippo's data, prompts, loss, evaluation, precision and AdamW (with their own learning rate, 2e-5 on every parameter, against DeltaHippo's 4e-5; batch 32, 3 epochs per task); DeltaHippo's relative step and width rule are not applied to them.

\begin{table*}[!t]
{\tabsize\setlength{\tabcolsep}{3pt}\begin{tabularx}{\textwidth}{@{}l>{\raggedright\arraybackslash}X>{\raggedright\arraybackslash}X@{}}
\toprule
\textbf{Item} & \textbf{Sculpting Subspaces} & \textbf{CODE-CL}\\
\midrule
Matrices constrained & q, k, v, o, gate, up and down projections of every block & every linear map, including the LM head (the tied input embedding matrix is constrained through it); q/k/v and gate/up share one conceptor\\
What is protected & the top singular directions of each current weight matrix & the conceptor $C = R(R + \alpha^{-2} I)^{-1}$ of each map's input moment $R$, accumulated over tasks by the conceptor OR\\
Amount & fixed budget: the top $t/n$ of each matrix's directions frozen at task $t$ ($t = 0, \dots, n - 1$) & aperture $\alpha = 6$, aperture-gain target 0.95, singular values of $C$ below 0.2 set to 0; 125 training sentences per conceptor\\
When & SVD of the merged weights at the start of every task after the first & conceptor after every task except the last\\
Update & the singular-vector factors of the remaining (low-singular-value) subspace are trained; their gradients are projected off the frozen singular directions; factors merged into the dense weight at the end of each task & the AdamW step (decoupled weight decay included) is right-multiplied by $(I - C)$, so that the written change has the form the method prescribes\\
Free dimensions & none (not part of the method) & $K = 0$: no task-specific mixing matrices, so no task identity is needed at test time\\
Optimizer state & new for every task & new for every task\\
First task & plain fine-tuning & plain fine-tuning\\
\bottomrule
\end{tabularx}}
\end{table*}

\emph{Implementation checks.} The conceptor operations (conceptor, aperture-gain loop, AND, OR, lower bound) run in fp64, where they agree with the official code on random conceptors to 1e-10 (AND), 4e-15 (OR) and 9e-15 (conceptor with gain loop).

\emph{Configuration.} Both choices are set by the protocol (CODE-CL, $K = 0$) or by the weights (Sculpting Subspaces, fixed budget), not by test accuracy, and are fixed for every model and benchmark. The alternatives, and a sensitivity check of the learning rate and of CODE-CL's aperture, were also run on one setting (Qwen3-0.6B, TACRED; three runs each, seeds 42, 11, 3) and are reported here.

\begin{itemize}[leftmargin=1.4em,itemsep=1pt,topsep=2pt,parsep=0pt]
\item[\textbullet] CODE-CL uses $K = 0$, its task-agnostic form. With $K > 0$ the method mixes task-specific matrices, which need the task identity that the class-incremental protocol withholds at test time. Folding $K = 80$ mixing matrices into the single class-incremental model, after the second task, scores 0.0 on task 0 and 2.4–3.5 on task 1, and ends at 0.0 in all three runs; $K = 0$ ends at 40.00 ± 4.39.
\item[\textbullet] Sculpting Subspaces uses the fixed budget. On these models a matrix's input and output are nearly orthogonal on average (mean per-matrix cosine between \textminus{}0.0007 and 0.0013 over tasks and runs), so Sculpting Subspaces' adaptive rank rule (Eq. 2, importance normalised by its mean, minimum and target retained fractions 0.1 and 0.8) keeps either one direction or all of them per matrix. The adaptive rule ends at 15.12 ± 1.33 and the fixed budget at 41.33 ± 0.48.
\item[\textbullet] Learning rate and aperture. With the fixed budget, Sculpting Subspaces ends at 43.53 ± 0.87, 41.33 ± 0.48 and 23.42 ± 5.43 at learning rates 1e-5, 2e-5 and 5e-5; CODE-CL ($K = 0$) ends at 32.51 ± 2.55, 40.00 ± 4.39 and 48.31 ± 2.44 at the same rates, and at 48.82 ± 4.87 and 22.34 ± 0.44 with aperture $\alpha = 4$ and 16 ($\alpha = 6$: 40.00 ± 4.39). Even the best variant of each stays 23.6 points or more below DeltaHippo (72.46 ± 1.80).
\end{itemize}

\section*{Supplementary Note 4: Detailed results}

This note gives the full account of the analyses summarised in the main text, in the same order.

\subsubsection*{A training-only memory that governs how the whole network learns}

The model is the plain network, trained on the current task's supervision (cross-entropy on the answer tokens) with the base optimizer, AdamW; the memory modulates the gradient and the realised update. Every parameter tensor stays in the optimization on every task, including the output head, the input embedding matrix and the normalisation gains. The memory changes only how each step is applied, per synapse group: the set of linear maps that read the same input tensor, so that a change to any of them alters earlier outputs through that shared input (four per transformer layer, plus the output head; Methods: Synapse groups), and each normalisation gain and embedding row is protected according to its own history.

\textbf{Sleep.} At the end of each task, a gradient-free encoding phase re-reads that task's own training data (Fig. 1d). For each synapse group, it eigendecomposes the second-moment matrix of the group's inputs over the task and keeps the directions above a data-dependent spectral cutoff adapted from the Gavish–Donoho rule\textsuperscript{61}. It merges them into the group's memory by a gated delta rule, the write–erase mechanism of Gated DeltaNet\textsuperscript{38}. The memory of each synapse group thus contains:

\begin{itemize}[leftmargin=1.4em,itemsep=1pt,topsep=2pt,parsep=0pt]
\item[\textbullet] a low-rank subspace $U$ with energies $\lambda$;
\item[\textbullet] the mean residual energy $\tau$ outside $U$, and the task's noise floor $f$;
\item[\textbullet] per class, the mean answer state and the mean input pattern.
\end{itemize}

No sentence, logit, answer string or answer-token id is kept, and earlier tasks are never re-read.

\textbf{Wake.} While a later task is learnt, every step passes through a closed-form operator derived from a local objective (Fig. 1e): the new loss plus a memory-weighted change of each synapse group's output.

\begin{itemize}[leftmargin=1.4em,itemsep=1pt,topsep=2pt,parsep=0pt]
\item[\textbullet] Each remembered direction keeps a plasticity factor $f/(f + \mu\lambda_i)$ of the step, which falls with its stored energy and with the amount of earlier experience $\mu$.
\item[\textbullet] Every direction outside $U$ keeps the residual plasticity factor $f/(f + \mu\tau)$ (the residual floor), set by the residual energy.
\item[\textbullet] The base update leaves the class-mean inputs of earlier classes that fall outside $U$ exactly unchanged (class-pattern constraint).
\end{itemize}

Adam's per-coordinate normalisation rotates a constrained gradient back into protected directions (Proposition 1, Methods). The operator is therefore applied again to the update Adam actually takes (the update constraint).

\textbf{Conflict.} The class records add conflict detection (Fig. 1e, bottom right). At every answer-token position of a training sentence, the network's answer state is compared with the records of earlier classes. It is in conflict with an earlier class when it lies closer to that class's mean state than the nearest other class's mean does (Methods: Conflict detection). Only the state is read; which answer the network would give, and whether a competing token was ever an earlier answer, are never checked. A conflict sets three responses, each where it is needed (Methods: Learning at detected conflicts):

\begin{itemize}[leftmargin=1.4em,itemsep=1pt,topsep=2pt,parsep=0pt]
\item[\textbullet] \textbf{Update gate.} The more of a step's answer states fall inside strongly recorded earlier classes, the lower the probability that the step is applied.
\item[\textbullet] \textbf{Comparator.} At conflicting positions whose own answer does not yet win, the weight update uses only the part of the input that no earlier class pattern explains, so the new class is learnt along what distinguishes it.
\item[\textbullet] \textbf{Joint-share target.} Where a new class shares its decision region with an earlier near-duplicate, its probability is driven to a share set by the two classes' experience instead of to 1, so it cannot take the whole region.
\end{itemize}

\textbf{Evaluation.} The memory never produces an output. At test time the memory is removed and the plain model generates its answer (Fig. 1f). Its training step costs 1.25–1.33 times a plain fine-tuning step, and inference is that of the plain model. The added cost is one offline sleep per task, which never needs more GPU memory than training (Methods: Cost). The design pays an offline memory-formation cost instead of an inference cost.

\subsubsection*{One principle keeps whole networks learning across architectures and scales}

We evaluate the same principle on the ten models of the Introduction and six class-incremental benchmarks. The decoder large language models (LLMs), which learn by generation, and the bidirectional encoders learn four text benchmarks: TACRED\textsuperscript{48}, FewRel\textsuperscript{49}, Banking77\textsuperscript{50} and CLINC150\textsuperscript{51}. The vision transformer and the convolutional network learn CIFAR-100\textsuperscript{52} and ImageNet-R\textsuperscript{53}. The four families present the memory with different units, output heads and normalisations, and each maps onto the same index:

\begin{table*}[!t]
{\tabsize\setlength{\tabcolsep}{3pt}\begin{tabularx}{\textwidth}{@{}>{\raggedright\arraybackslash}X>{\raggedright\arraybackslash}X>{\raggedright\arraybackslash}X>{\raggedright\arraybackslash}X>{\raggedright\arraybackslash}X@{}}
\toprule
\textbf{Family} & \textbf{Models (parameters)} & \textbf{Units the memory indexes} & \textbf{How the model answers} & \textbf{Normalisation}\\
\midrule
Convolutional network & ResNet-50 (25M) & every convolution, as a linear map on its unfolded input & linear head on the global pool & BatchNorm gains (pre-trained running statistics)\\
Vision transformer & ViT-B/16 (86M) & q/k/v, attention output, MLP input and output; patch embedding & linear head on [CLS] & LayerNorm gains\\
Bidirectional text encoders & BERT, RoBERTa (110–125M) & q/k/v, attention output, intermediate, output; token, position and type embeddings & linear head on [CLS] & LayerNorm gains (post-LN)\\
Decoder language models & Qwen3, Llama-3 (0.6–8B) & q/k/v, o, gate/up, down; input embedding matrix; LM head & generation through the LM head, tied or untied & RMSNorm gains\\
\bottomrule
\end{tabularx}}
\end{table*}

The class-free core is the same in every architecture: synapse groups indexed by their shared input, a post-task memory of those inputs with a residual floor, and a constraint on both the gradient and the realised update. Which operator sets the plasticity factors, how the classifier is protected and how embeddings and normalisation gains are treated are fitted to each architecture's structure (Methods: Encoders and vision models). Table 1 (decoders) and Extended Data Tables 1–2 (encoders, vision) report the final average accuracy (AA) after the last task; the baselines are marked by what they rely on. SEQ is sequential full fine-tuning (full FT) of the whole model without protection; SEQ*\textsuperscript{33} adapts the backbone on the first task, then freezes it (on the encoders, with a one-sample-per-class buffer; Extended Data Table 1).

\textbf{Comparison across regimes.} DeltaHippo fine-tunes every parameter and relies on none of the resources marked in the tables above; apart from SEQ and the subspace-projection methods, the baselines work in other regimes.

\begin{itemize}[leftmargin=1.4em,itemsep=1pt,topsep=2pt,parsep=0pt]
\item[1.] \textbf{Against sequential fine-tuning.} Under the same operational constraints, DeltaHippo exceeds SEQ on every cell: by 25–63 points on the LLMs, 39–77 points on the encoders and 41–67 points on the vision models.
\item[2.] \textbf{Subspace-constrained full fine-tuning.} Sculpting Subspaces\textsuperscript{25} and CODE-CL\textsuperscript{24} also update every weight without replay or a teacher, constraining each update by a stored subspace: the weights' own top singular directions (Sculpting) or a conceptor of earlier inputs (CODE-CL). DeltaHippo exceeds both on every cell, by 4.2–46.7 points. The design axes of the classic input-space projections are isolated by the ablations analysed below (Extended Data Table 4; retention is the mean accuracy on the tasks learnt before each task): an open complement around the stored basis, as in GPM\textsuperscript{19} (opening it lowers retention from 82.1 to 79.2); enforcement on the optimizer's update alone, the stage Adam-NSCL acts on\textsuperscript{20} (53.1); and enforcement on the gradient alone, the stage of OWM-type projectors\textsuperscript{17} (60.3; three runs each). Each axis on its own retains less than the two-sided, residual-aware constraint; removing either constraint or opening the residual subspace also costs accuracy in BERT-base and ResNet-50, and removing the update constraint does in ViT-B/16 (Extended Data Table 5); the external comparison of whole methods is the one with Sculpting Subspaces and CODE-CL above.
\item[3.] \textbf{Replay with a teacher.} It exceeds LAMOL\textsuperscript{39} and PCLL\textsuperscript{40} on every cell where they are evaluated (Qwen3-0.6B and Llama-3.2-1B, all four benchmarks).
\item[4.] \textbf{Teacher distillation.} It exceeds L2KD\textsuperscript{41} on every cell where L2KD is evaluated (Qwen3-0.6B and Llama-3.2-1B, all four benchmarks).
\item[5.] \textbf{Replay buffers.} It exceeds DER++\textsuperscript{42} and CLS-ER\textsuperscript{43} on every encoder and vision cell.
\item[6.] \textbf{Frozen backbone with adapters or LoRA.} On ViT-B/16, EASE\textsuperscript{8} and SD-LoRA\textsuperscript{13} train adapters or low-rank adaptation (LoRA) modules on a frozen pre-trained backbone. DeltaHippo is higher on CIFAR-100 (88.21 against 87.02 and 87.47). They work in the frozen-backbone regime, adding task-specific modules to an unchanged backbone, which our constraints exclude; DeltaHippo reaches this level while every parameter tensor of the backbone keeps learning.
\item[7.] \textbf{Frozen backbone.} SEQ* stops updating the backbone after the first task. It therefore belongs to the frozen-backbone regime, which works under looser constraints than ours: a backbone that no longer changes has no representational interference to manage. It leans on the pre-trained representation, so it holds up as far as that representation already separates the classes, and no further. DeltaHippo learns on every task and still ends higher on 19 of the 24 LLM cells, by up to 22.4 points (Llama-3.2-3B TACRED: 76.40 vs 53.99), on every encoder cell (by 18–33 points), and on three of the four vision cells (by 6.7–21.4); on ViT-B/16 ImageNet-R the two end level (64.17 against 64.23, within the run-to-run spread). The remaining five are cells where the frozen backbone already separates the classes, four of them on CLINC150 (87–93).
\item[8.] \textbf{Scale.} With the step-size rule for the widest models (Methods: Model training), the 8B models exceed the smaller models of the same series on every benchmark: Qwen3-8B on TACRED (76.79 vs 76.52 and 72.46), FewRel (85.95 vs 76.52 and 71.01), Banking77 (88.36 vs 87.46 and 82.91) and CLINC150 (93.70 vs 92.33 and 87.13), and Llama-3.1-8B on TACRED (77.35 vs 76.40 and 71.58), FewRel (82.92 vs 81.48 and 79.09), Banking77 (87.80 vs 85.84 and 82.28) and CLINC150 (92.53 vs 91.68 and 87.04).
\item[9.] \textbf{Architectures.} The gain is present in every family (Fig. 2c–e). Averaged over benchmarks, every transformer model, from 86 million to 8 billion parameters, ends at 75–86\%, and the convolutional network ends at 65\% (sequential fine-tuning: 9\%).
\end{itemize}

\textbf{Where near-duplicate classes meet, DeltaHippo leads by the widest margin.} On TACRED, where many relations have near-duplicates learnt in other tasks (the operational rule is given with Extended Data Table 6), DeltaHippo is ahead by 17.5 points over SEQ* averaged over the six decoders (10.4–22.4) and by 33–40 points over sequential fine-tuning, against +6.4 and +3.9 over SEQ* on Banking77 and FewRel, with the two level on CLINC150. The margin holds under other class orders (Llama-3.2-1B: 13.0 and 17.9 points over SEQ* on two further orders, against 16.6 in the protocol order; Supplementary Note 2).

Retaining the earlier tasks would be trivial if the backbone simply stopped changing. The acquisition–retention trajectories show that it does not, and a same-head control shows that its continued learning pays off.

\textbf{Acquisition–retention trajectories.} For every task $t$, acquisition is the accuracy on task $t$ right after learning it, and retention is the mean accuracy on the earlier tasks at that point.

\begin{itemize}[leftmargin=1.4em,itemsep=1pt,topsep=2pt,parsep=0pt]
\item[\textbullet] SEQ acquires every task (88–99) but keeps little of the earlier ones: 0–38 on the language panels and 15–39 on ViT-B/16.
\item[\textbullet] DeltaHippo retains 63–95 throughout while acquiring every task.
\end{itemize}

\textbf{Same head, backbone frozen.} On ViT-B/16 ImageNet-R we freeze the backbone after task 0 and keep everything else of DeltaHippo, including its classifier rule (Methods: Encoders and vision models; Fig. 2b; Extended Data Table 3).

This control keeps DeltaHippo's memory and classifier rule, so the only difference from the method is whether the backbone learns. Learning the backbone is worth 10.9 points: acquisition of new tasks rises by 13.8 points (46.2 $\rightarrow$ 60.0). Within the same learning system, a backbone that keeps changing accumulates more useful knowledge than a frozen one.

\subsubsection*{Memory content, not step size, decides what is kept}

To ask why the method works, we intervene on the memory at a task transition. Unless stated otherwise the testbed is Llama-3.2-1B on TACRED, where both retention and acquisition are visible; every condition is identical to the normal run, including its random draws, up to the point where it is changed (Methods: Analysis protocol).

\textbf{Design.} We train the normal method through task 3 and capture the full training state (weights, Adam moments, hippocampus, counters, random-number generator states). From that state, task 4 is learnt under six conditions that share the same minibatches, learning rate and evaluation (Fig. 3), and we measure three groups: the older tasks 0–2, the just-learnt task 3, and the new task 4.

\begin{table*}[!t]
{\tabsize\setlength{\tabcolsep}{3pt}\begin{tabular*}{\textwidth}{@{\extracolsep{\fill}}lllll@{}}
\toprule
\textbf{Condition} & \textbf{What changes} & \textbf{Old tasks 0–2} & \textbf{Task 3} & \textbf{New task 4}\\
\midrule
A. Normal & nothing & \textbf{83.8 ± 3.2} & \textbf{70.7 ± 3.4} & 69.3 ± 4.9\\
B. Read off & the hippocampus is not read; plain update & 0.0 ± 0.0 & 0.0 ± 0.0 & 91.4 ± 1.4\\
C. Scrambled content & same rank and eigenvalues, directions rotated inside each synapse group & 4.6 ± 5.5 & 0.0 ± 0.0 & 91.1 ± 0.6\\
D. Read restored & as B, then the original read is reconnected at 25\% of task 4 & 0.0 ± 0.0 & 0.0 ± 0.0 & 91.6 ± 1.6\\
E. Last sleep blocked & task 3's sleep is not written; the earlier memory is read & 81.3 ± 3.1 & 38.7 ± 5.8 & 79.7 ± 1.1\\
F. Norm-matched plain & plain direction, with A's per-synapse-group applied norm and A's applied/skipped steps & 2.0 ± 3.4 & 0.0 ± 0.0 & 91.3 ± 1.5\\
\bottomrule
\end{tabular*}}
\end{table*}

Accuracy (\%) at the end of task 4, mean ± sample standard deviation over three independent runs, each with its own prefix through task 3.

\begin{itemize}[leftmargin=1.4em,itemsep=1pt,topsep=2pt,parsep=0pt]
\item[\textbullet] \textbf{The memory is necessary (B).} Without its read, performance on every earlier task collapses within the first quarter of task 4 (Fig. 3a–c).
\item[\textbullet] \textbf{The spectrum alone is insufficient (C).} Rotating the stored directions keeps every eigenvalue (to within $3.6 \times 10^{-5}$) and every plasticity factor, yet performance on the earlier tasks collapses. The directions the hippocampus protects must match what the network experienced.
\item[\textbullet] \textbf{It is not a smaller step (F).} Moving each synapse group by exactly the method's amount, at the same steps, in the plain direction, collapses the earlier tasks as well. The benefit is not a lower effective learning rate: the memory organises the \emph{direction} of change.
\item[\textbullet] \textbf{Restoring the read does not restore lost performance (D).} The memory holds no copy of the knowledge, and reconnecting the read does not by itself restore access that was already lost. This separates two functions: holding the knowledge that produces the current answer, which lives in the network, and regulating how much the next experience damages it, which the hippocampus does.
\item[\textbullet] \textbf{Each sleep selectively protects the episode it encoded (E).} Against the normal condition after the same task 4, blocking only task 3's sleep costs task 3 32.0 points (70.7 $\rightarrow$ 38.7) and the older tasks 2.5 points (83.8 $\rightarrow$ 81.3).
\end{itemize}

\subsubsection*{A class-free geometric core carries most of the effect: a two-sided constraint and a residual floor}

Extended Data Table 4 identifies which parts of the operator carry the effect while the whole model, including the output head, the tied embedding matrix and the normalisation gains, keeps training under a compressed memory. Its first comparison asks how much of the method needs class information at all.

\begin{itemize}[leftmargin=1.4em,itemsep=1pt,topsep=2pt,parsep=0pt]
\item[\textbullet] \textbf{The geometric core alone carries most of the effect; the class-conditioned parts add a further 4.5 points.} Removing every class-conditioned part together (the class records, class-pattern constraint, conflict detection with the update gate, comparator, joint-share target and erase) leaves the residual-aware index and its two constraints, with the protected subspace constrained exactly and the residual floor outside it (Box 1). Over three independent runs this class-free core reaches 67.1 ± 2.2 (acquisition 74.4, retention 80.9) against 71.6 ± 0.5 (77.9, 82.1) for the full method: of the full method's 36.5-point gain over sequential fine-tuning (35.1), the core delivers 32.0 and the class-conditioned parts the remaining 4.5, at a quarter of the run-to-run spread (SD 0.5 against 2.2), and on ViT-B/16 ImageNet-R the same core ends at 63.39 against 63.95 for the full method in the matched run. In the vision transformer the classifier follows the same rule: earlier heads are constrained by the class-free operator on their input, and the present head is kept orthogonal to the pooled mean of all earlier decision states, a single direction that carries no class identity. Nothing in the core reads a class label, so what protects earlier knowledge here is the geometry of past inputs itself. What the core does not do is keep near-duplicate classes apart; that is the role of the class-conditioned parts, examined in the next section.
\end{itemize}

\begin{itemize}[leftmargin=1.4em,itemsep=1pt,topsep=2pt,parsep=0pt]
\item[\textbullet] \textbf{Within the full system, the update constraint is worth 9.4 points of final AA and 22 points of retention.} With the gradient constrained but Adam's update applied as it comes, retention falls from 82.1 to 60.3 while acquisition rises from 77.9 to 87.3 (three runs): constraining the gradient alone does not constrain the update that is applied. At the first transition, where both runs start from the same state, the applied update without the update constraint carries 3.0–80 times the interference energy, measured on the memory's stored model of earlier inputs rather than on held-out data (by synapse group: attention input 4.2×, attention output 3.0×, MLP input 3.8×, MLP down 3.3×, input embedding matrix 80×), while its own energy is only 1.6–1.9 times larger. This is Proposition 1 at the level of the whole task sequence. The loss holds in each of the three runs (final AA 62.1 against 71.6). The converse fails as well: constraining only the realised update, with the gradient passed to Adam unconstrained, loses 29 points of retention (82.1 $\rightarrow$ 53.1) and 15.2 points of final AA over three runs. The gradient constraint shapes the moments Adam accumulates; the update constraint re-imposes it on the update those moments produce. Neither does the other's work.
\item[\textbullet] \textbf{Protecting the residual subspace carries retention, in every run.} What matters is the complement of the stored span, not how the span itself is protected. Making the protected subspace hard while keeping the residual plasticity factor leaves retention intact (82.9 against 82.1, three runs); keeping the graded span but opening the residual subspace lowers retention to 79.2, while mean acquisition rises from 77.9 to 85.3: what the open complement learns, it learns at the expense of earlier tasks. The directions outside the remembered span are not individually stored, yet protecting them in proportion to the mean energy that earlier inputs left there, $f/(f + \mu\tau)$, is what keeps earlier tasks while new ones are learnt; projection methods that protect a stored span and leave its complement free, as GPM does, miss exactly this part. This establishes a design principle for compressed memories: the complement of what is stored must not be treated as free. The less the operator protects explicitly, the more the residual floor carries. When it protects only the top quarter of each synapse group's stored directions (the memory itself unchanged; the dropped energy joins the residual energy), the residual floor is worth 18.9 points of final AA (68.7 against 49.8 with the residual subspace open) and 26.6 of retention over three runs; with the residual subspace open the earlier tasks collapse from task 2 on. A smaller protected subspace still costs retention even with the floor (79.7 against 82.1), so the floor complements the stored span rather than replacing it.
\item[\textbullet] \textbf{The same conditions hold outside the decoder LLMs.} In BERT-base on TACRED and ResNet-50 on CIFAR-100, removing the update constraint costs 3.1 and 30.6 points of final AA, removing the gradient constraint 9.8 and 37.4, and opening the residual subspace 15.8 and 41.0; in ViT-B/16 on CIFAR-100, removing the update constraint costs 17.0 points over three runs, with retention falling from 91.6 to 79.3 (Extended Data Table 5). Without the update constraint or the residual floor, new tasks are still learnt (acquisition 74–98) while the earlier ones are lost, the same signature as in the language models.
\end{itemize}

\textbf{Erase lowers the stored energy; on these sequences the memory is far from saturation.} At each sleep the gated delta rule also erases old content along the keys it writes, less where conflicting sentences of the task used them (Methods: Sleep). We rebuilt the memory on each variant's own trajectory (Supplementary Table 1). Without erase the memory ends with about 80\% more stored energy (96 against 54), and a non-selective erase that removes the same energy without using the conflict signal loses 1.2 points of final AA. Because the protected subspace stays well below the full dimension of each synapse group (0.284 $\rightarrow$ 0.291 of its input dimension over the TACRED sequence; 0.46–0.49 after the 15 CLINC150 tasks of the 8B models), an accumulating memory reaches the same result on these sequences, and acquisition, update rate and the residual floor are nearly unchanged: erase lowers the energy of directions the memory already protects, while new learning flows mainly through the residual floor. Erase is designed for the regime these sequences do not reach, in which a longer sequence or a fixed memory budget saturates an accumulating index, much as active forgetting and sleep-dependent synaptic downscaling are proposed to keep a limited-capacity store usable\textsuperscript{62–64}. That less than half of each synapse group's input dimension carries 15 tasks is itself a practical property: most input directions stay open while the network keeps learning.

\subsubsection*{Every compared method loses earlier near-duplicate classes; conflict detection keeps them}

Average accuracy averages over tasks and their test sentences, so the loss of one earlier class, one of forty, is diluted in it and barely registers the failure described in the Introduction. We therefore measure that failure directly.

\textbf{Near-duplicate classes and absorption.} Operationally, we count two relations as near-duplicates when their labels, after the person/organisation prefix, differ in a single word (a person's \emph{city of birth} and \emph{city of death}) and are learnt in different tasks. The set is fixed by this rule on the label strings alone, without reference to any model's predictions. On TACRED, 15 of the 40 relations have a near-duplicate in another task, among them a person's \emph{city of birth}, \emph{city of death} and \emph{city of residence}, the \emph{city}, \emph{state or province} and \emph{country} of an organisation's headquarters, and a person's \emph{date of birth} and \emph{date of death} (all 15 are listed in Supplementary Note 2). A sentence naming a person and a city can express any of these relations; what separates them is the role the entity plays, which the label names in a word or two. The memory never reads label strings or token ids: a collision is detected from the network's answer state alone (Methods: Conflict detection). Eight of the 15 are learnt before one of their near-duplicates; we call each of them an earlier class. When a later near-duplicate is learnt, its answer states fall on the earlier class's decision region and, left alone, it takes the region over: the earlier class's test sentences are then answered as the later class (absorption). From the per-sentence answers at the end of the task sequence (Llama-3.2-1B; three runs of each Table 1 configuration), we report for every method the share of the earlier classes' test sentences answered correctly, the share answered as one of their later near-duplicates (absorbed), and, among the errors on all 15 relations, the share that name one of the relation's near-duplicates. The first two are computed per class and then averaged over the eight classes. Every method first learns these classes: right after their own task it answers 72–96\% of their test sentences correctly (DeltaHippo 84\%). The end-of-sequence values therefore measure what is lost after the classes were learnt; how the loss unfolds over the sequence is followed for DeltaHippo and its class-free core in Fig. 4d–f.

\textbf{Every compared method loses them, whatever its regime.} Whatever it relies on, every compared method answers fewer of the earlier classes' sentences correctly than DeltaHippo (0.9–63.0\% against 76.6\%), absorbs 3 to 14 times as many of them into their later near-duplicates (9.0–40.5\% against 2.8\%, 6.2–37.7 percentage points more), and, when it errs on these relations, names a near-duplicate up to 2.8 times as often (21–32\% against 12\%; Extended Data Table 6, Fig. 4a–c). The same holds at every scale: on all six decoders, DeltaHippo answers more of the earlier classes' test sentences correctly than SEQ* (73.5–81.2\% against 51.1–64.5\%; Extended Data Table 7).

\begin{itemize}[leftmargin=1.4em,itemsep=1pt,topsep=2pt,parsep=0pt]
\item[\textbullet] Without protection, or with distillation from the previous model alone, the later class takes the shared region outright: sequential fine-tuning absorbs 31\% of the earlier classes' sentences and L2KD 40\%.
\item[\textbullet] Replay with a teacher re-presents earlier outputs, yet the later class still takes part of the region (LAMOL 11\%, PCLL 30\%).
\item[\textbullet] Protecting subspaces of the weights or of the earlier inputs cannot separate two classes that use the same directions: Sculpting Subspaces absorbs 25\% and CODE-CL 21\%.
\item[\textbullet] A frozen backbone keeps the space that the first task left, and class prototypes placed in that fixed space do not separate a pair that the space does not already separate: SEQ* absorbs 9\%, and among a person's three city relations \emph{city of death} is answered as \emph{city of residence} in 37\% of its sentences and \emph{city of birth} as \emph{city of death} in 26\%, whereas DeltaHippo answers them correctly in 89 and 95\% (Fig. 4c).
\end{itemize}

The failure is the same in every regime. A later class whose answer state falls on an earlier class's decision region takes that region, and nothing in these methods registers when it happens: they protect what earlier tasks produced (their outputs, their weights' subspaces, their inputs, or the whole backbone) but not the boundary between two classes that only the later task puts under pressure.

\textbf{Conflict detection is what keeps them.} The index alone does not keep near-duplicates apart, because they rely on nearly the same directions (previous section). Conflict detection supplies what it lacks: it flags the answer-token positions of the present task whose state falls inside an earlier class's record, which is exactly where absorption begins, and there the update gate, the CA1 comparator and the joint-share target change how the network learns (Fig. 1e; Methods: Conflict detection; Learning at detected conflicts).

\textbf{Without the class-conditioned parts the established class is lost; with them both classes stay in play.} Removing every class-conditioned part, conflict detection among them, gives the class-free core of the previous section (Extended Data Table 4). Under the class-free core, the task-2 class \emph{state or provinces of residence} falls to 0 of 41 in all three runs when its near-duplicate \emph{state or province of birth} is learnt, whereas the full method keeps 29 on average (up to 39). In all three runs it is task 2, the home of these near-duplicate classes, that the core lets fall (44–57 at the end, against 64–83 under the full method). The near-duplicate pair splits winner-take-all: under the core the newcomer takes their shared decision region (7–8 of its 9 test sentences) and the established class is lost; the full method leaves the decision region with the established class or shares it. With both classes weighted equally, the full method keeps the pair better in every run: its mean recall is 59, 70 and 48\% against 44, 39 and 39\% under the core, and the lower of the two recalls is 29, 44 and 11\% against 0\% (Fig. 4d–f). The core learns the newcomer by giving up the established class entirely; the full method keeps both classes in play. Even average accuracy, which dilutes the loss of a few classes over forty, separates the two systems by 4.5 points.

\textbf{The comparator is what lets colliding classes be learnt.} Over the full task sequence, switching only the comparator off lowers the final AA by 12.9 points over three runs, from 71.6 ± 0.5 to 58.7 ± 0.6. The loss is acquisition: at task 6 the new classes are answered as their earlier near-duplicates (\emph{city of death} answered as \emph{city of residence} 21 times, against at most 3 with the comparator). Over the full task sequence, applying the comparator at the same number of positions but at positions conflict detection did not select lowers mean acquisition to 74.3, against 77.9 normally and 60.6 with the comparator off (three runs). Where the comparator acts matters: placing it by conflict detection raises acquisition by 3.5 points over placing it elsewhere. Conflict detection is the signal that places selective learning where the present experience overlaps memory.

\textbf{The joint-share target caps the share a new class takes of a shared decision region.} At answer-token positions where a new class shares its decision region with an earlier one, it drives the gold probability to the share set by the two classes' joint experience instead of to 1 (Methods: Learning at detected conflicts).

Because the whole network keeps adapting while conflict detection holds these boundaries, the earlier distinction stays usable and the later class is still acquired.

\subsubsection*{Earlier knowledge stays usable when representation and output are guided together}

The geometric core and conflict detection protect earlier classes; what they protect, in the end, is the compatibility between the changing representation and the output that answers with it.

\textbf{The collapse is mainly a loss of access, not of linear separability.} Returning to the task-4 intervention (Fig. 3), a probe fitted on each condition's own final-layer states at the end of task 4 still separates the earlier classes nearly as well as in the normal run (A), although generation on them is 0–1.2\% with the read off (B), the content scrambled (C) or the step norm-matched (F): without guidance, generation on tasks 0–2 falls by 80–82 points while the own-condition probe changes by only \textminus{}4.5 to +1.6 (\textminus{}5.0 to +3.9 on task 3). The normal run's probe applied unchanged to the other conditions does not separate them (Supplementary Table 18): the earlier classes move, and a decision rule fitted on the normal run no longer finds them. What the hippocampus preserves, above all, is the alignment between the backbone's changing representation and the output pathway that answers with it.

\textbf{Access needs both sides.} Guiding only the LM head (input embedding rows and final norm), or only the backbone, over the whole task sequence:

Mean ± sample standard deviation over three independent runs; retention as defined in Extended Data Table 4.

\begin{table*}[!t]
{\tabsize\setlength{\tabcolsep}{3pt}\begin{tabular*}{\textwidth}{@{\extracolsep{\fill}}llll@{}}
\toprule
\textbf{Guidance on} & \textbf{Final AA} & \textbf{Mean acquisition} & \textbf{Retention}\\
\midrule
Whole model (DeltaHippo) & 71.6 ± 0.5 & 77.9 ± 2.0 & 82.1 ± 1.7\\
Backbone only & 53.6 ± 11.0 & 90.3 ± 1.0 & 53.4 ± 9.8\\
Head only & 12.1 ± 0.2 & 91.3 ± 1.2 & 1.7 ± 0.7\\
\bottomrule
\end{tabular*}}
\end{table*}

Both partial runs learn each new task well and lose the earlier ones, yet at task 4 a probe fitted on each run's own states still separates the earlier tasks 0–3 (Fig. 5a: 82.4 for backbone only and 85.5 for head only, against 83.3 for the whole model) while generation on them is 37.7 and 0.0. To locate which side breaks the model's own access, we take each run's last-layer hidden states before the final norm at the end of task 4, at the teacher-forced gold answer positions of the test sets of tasks 0–3 (745 answers), and apply every run's actual output head (its final norm and its tied embedding rows, over the full vocabulary) to every run's states, with the models themselves left unchanged. Entries are label-level accuracy: an answer counts only if every answer token is the argmax, which equals the greedy answer under these fixed states. Each diagonal entry reproduces that model's own teacher-forced accuracy to within 0.5 points. Values (\%), pooled over the 745 answers, mean ± sample standard deviation over the three runs (Fig. 5b):

\begin{table*}[!t]
{\tabsize\setlength{\tabcolsep}{3pt}\begin{tabular*}{\textwidth}{@{\extracolsep{\fill}}llll@{}}
\toprule
\textbf{States from \textbackslash{} output head from} & \textbf{Whole model} & \textbf{Backbone only} & \textbf{Head only}\\
\midrule
Whole model & 80.4 ± 2.8 & 43.4 ± 1.8 & 80.7 ± 2.7\\
Backbone only & 61.8 ± 20.3 & 23.8 ± 12.4 & 60.9 ± 20.1\\
Head only & 0.6 ± 1.1 & 0.0 ± 0.0 & 0.6 ± 1.1\\
\bottomrule
\end{tabular*}}
\end{table*}

With states held fixed, the output head matters: a guided one still reads the whole-model states (80.7), while the unguided one of the backbone-only run reads them at 43.4. With the head held fixed, the states matter: the whole-model head reads guided backbone states partly (61.8) but unguided backbone states hardly at all, both after a whole task sequence (head only: 0.6) and after a single unguided task (read off, scrambled and norm-matched conditions at task 4: 0.0, 1.5 and 0.0), although a probe re-fitted on those states still separates the classes. Restoring the whole-model head recovers the backbone-only states from 23.8 to 61.8 but not to 80.4, so neither side's guidance keeps its part intact on its own: the two pathways learn interdependently. The pattern holds in each of the three runs. A re-fitted probe tolerates state changes that the model's fixed answer-token rows do not: high probe accuracy on a state does not mean the model's own output can still use it. In Llama-3.2-1B the LM head is tied to the input embedding matrix, so "head side" here means the tied input/output embedding rows and the final norm gain, and "backbone" every other synapse group; access is kept only when the memory guides both. This is why the hippocampus indexes every synapse group of the model, output head included.

\begin{itemize}[leftmargin=1.4em,itemsep=1pt,topsep=2pt,parsep=0pt]
\item[\textbullet] \textbf{Lost access is recovered by a low-rank correction but not by the calibrations tested.} With each condition's own output head fixed, we fit restricted maps on the earlier tasks' training sets (an analysis outside the method) and measure how much of the gap between the condition's own accuracy and a re-fitted probe each closes (Supplementary Table 19). The low-rank map adds a rank-$K$ linear change to the last-layer state before the fixed head, $h \to h + (hV)U^\top$ with $V$ and $U$ of width $K$. A bias on the label-token rows, with or without a temperature, closes at most 0.50 of the gap (backbone only: 24.7 $\rightarrow$ 56.5), and almost none (0.00–0.03) in the head-only run and the read-off condition. A rank-1 correction closes 0.95 of it in the backbone-only run, and a rank-4 correction closes 0.85–0.98 in every condition; the head-only run needs more than rank 1 (0.32). Lost access is thus more than a prior shift between earlier and newer classes that the bias and temperature tested here could undo: the class information the output no longer reaches is recovered by correcting the states along a few directions. Because these maps are fitted with labels on the earlier tasks' training sets, this establishes that a low-rank correction suffices, a statement about the correction rather than about the rank of the drift itself. Nor is the correction a general realignment. Maps fitted on one group of the earlier classes do not restore the other group (backbone only: 18–22\% against 29\% without a map) and lower it even in the whole model (85 to 33–36\%), whereas maps fitted on the evaluated classes reach 94–96\% in the whole model and the backbone-only run (Supplementary Table 20): the supervised correction relearns the decision rule for the classes it is fitted on. A large part of the backbone-only collapse appears when its answer competes with classes outside the evaluated set: restricting the answer to the evaluated classes' label tokens raises it from 29 to 73\%.
\item[\textbullet] \textbf{An untied model separates the two pathways cleanly.} In Qwen3-8B the output head (final norm and LM head) is separate from the input embedding matrix, so the backbone run also guides the input embedding matrix and the head run guides only the output head. Through task 4 of TACRED (one run), guiding only the output head protects almost nothing: each earlier task falls to 0 once a later task is learnt (task 1 alone stays partly readable, at 28–38, until task 4), and the average accuracy at task 4 is 18.0, although a probe fitted on its own features still reads the earlier tasks at 86.3. Guiding the backbone keeps most of the method's accuracy (average accuracy at task 4: 79.2 against 83.1). Its states are read by a guided head as well as the whole model's (85.1 against 84.4 on tasks 0–3), whereas its own unguided head reads them at 79.1, and reads the whole model's states at only 72.5. In this model the bulk of usable knowledge lives in the backbone's states, and keeping the output head aligned with them adds the rest; neither guided alone matches the whole model. The two pathways' contributions differ between the two models, and in both, joint guidance keeps earlier classes most usable.
\end{itemize}

\textbf{A common linear probe.} At every checkpoint we fit the same logistic-regression probe, over all classes seen so far, on the final-layer state just before the first answer token, and compare it with the same probe on the pre-trained model and on the backbone frozen at the end of task 0 (Fig. 5c; Supplementary Table 2).

\begin{itemize}[leftmargin=1.4em,itemsep=1pt,topsep=2pt,parsep=0pt]
\item[\textbullet] \textbf{Full fine-tuning keeps the representation, and DeltaHippo keeps it usable.} Every parameter was trained on seven further tasks, yet the earlier tasks remain as linearly separable at the end as in a backbone frozen right after task 0 (77.2 vs 75.4; newest task 84.0 vs 83.0) and well above the pre-trained model (72.0 and 75.5). Fig. 5a shows that this separability largely survives even without guidance; what DeltaHippo adds is that the separated classes stay reachable by the model's own output.
\end{itemize}

\section*{Supplementary Note 5: Software, hardware and runs}

\textbf{Software.} Python 3.12, PyTorch 2.9.1, Transformers 4.57.6, Accelerate 1.13.0 and FlashAttention 2.8.3. The full environment is released with the code as a frozen requirements file.

\textbf{Hardware.} Each run uses one GPU. The 8B decoders run on one NVIDIA B200 GPU; every other model runs on one NVIDIA H100 80GB GPU.

\textbf{Runs.} Unless stated otherwise, every reported number is the mean over three runs. Decoders and encoders use seeds 42, 11 and 3; vision models use seeds 1993, 42 and 3. The seed fixes the data order, the update gate's draws and, for encoders and vision models, the initialisation of the output heads. Each run is evaluated on the test split after every task; no checkpoint is selected. Supplementary Tables 13–17 list every run behind the means of the main text and Extended Data.

\textbf{Configurations.} Every run is launched from one configuration file that sets the model, benchmark, method and seed; the configurations are released with the code (tag v1.0).

\section*{Supplementary Note 6: Algorithm}

The procedure below restates Methods in executable order. The hippocampus $H$ is used only during training; at evaluation the plain model generates on its own.

\textbf{Algorithm 1 | DeltaHippo over a task sequence.}

\textbf{Input:} pre-trained network $W$; tasks $0, \dots, T-1$, each with its own training set $\mathcal{D}_e$; learning rate $\eta$; epochs $E = 3$.

\begin{itemize}[leftmargin=1.4em,itemsep=1pt,topsep=2pt,parsep=0pt]
\item[1.] $H \leftarrow$ empty; AdamW state $\leftarrow$ empty. \emph{One optimizer for the whole sequence.}
\item[2.] \textbf{for} $e = 0, \dots, T-1$:
\begin{itemize}[leftmargin=1.2em,itemsep=1pt,topsep=2pt,parsep=0pt]
\item[1.] \textbf{if} $e = 0$: train $W$ on $\mathcal{D}_0$ with plain AdamW for $E$ epochs. \emph{The memory is still empty.}
\item[2.] \textbf{else}: WAKE\textbackslash{}\_START($H$), \emph{building the operators from the memory (once per task)}; then \textbf{for} epoch $= 1, \dots, E$ and each batch $B$ in $\mathcal{D}_e$: TRAIN\textbackslash{}\_STEP($W$, $B$, $H$).
\item[3.] Evaluate the plain model on the test sets of tasks $0, \dots, e$ (greedy generation; $H$ detached).
\item[4.] \textbf{if} $e < T-1$: SLEEP($W$, $\mathcal{D}_e$, $H$). \emph{Write and erase the memory from task $e$'s own data.}
\end{itemize}
\end{itemize}

\textbf{Algorithm 2 | WAKE\_START: operators read from the memory (task start).}

\begin{itemize}[leftmargin=1.4em,itemsep=1pt,topsep=2pt,parsep=0pt]
\item[1.] $\bar c \leftarrow \operatorname{mean}_j H_j$; $M_j \leftarrow (H_j - \bar c) / \lVert H_j - \bar c \rVert$.
\item[2.] $r_j \leftarrow \max\{\, M_j^\top M_{j'} : \text{same offset, other class} \,\}$. \emph{Neighbourhood radius of each record.}
\item[3.] \textbf{for} each linear synapse group $a$:
\begin{itemize}[leftmargin=1.2em,itemsep=1pt,topsep=2pt,parsep=0pt]
\item[1.] $Q_a \leftarrow \operatorname{orth}(P^a)$. \emph{Earlier class patterns.}
\item[2.] $R_a \leftarrow \operatorname{orth}\big((I - U_a U_a^\top)\, Q_a\big)$; $W_a \leftarrow [U_a, R_a]$. \emph{Protected basis (one fp64 QR of $[U_a, P^a]$ on wide models).}
\item[3.] $\mu \leftarrow n_a / n_a^{\mathrm{last}}$; $\tau \leftarrow \mathrm{tail}_a / (d_a - r_a)$.
\item[4.] $a_i \leftarrow f_a / (f_a + \mu \lambda_i)$ on $U_a$; $a_i \leftarrow 0$ on $R_a$; $b \leftarrow f_a / (f_a + \mu \tau)$.
\item[5.] $T_a \leftarrow b\, I + W_a\, \mathrm{diag}(a - b)\, W_a^\top$. \emph{Stored as $(W_a, a, b)$; never formed as $d \times d$.}
\end{itemize}
\item[4.] \textbf{for} each embedding row $v$: $s_v \leftarrow f_{\mathrm{tok}} / (f_{\mathrm{tok}} + \mu_{\mathrm{tok}}\, \nu_v)$.
\end{itemize}

\textbf{Algorithm 3 | TRAIN\_STEP: one constrained update.}

\begin{itemize}[leftmargin=1.4em,itemsep=1pt,topsep=2pt,parsep=0pt]
\item[1.] Forward (bf16): logits, LM-head inputs $h_t$, synapse-group inputs $x_t$.
\item[2.] \textbf{for} each answer-token position $t$:
\begin{itemize}[leftmargin=1.2em,itemsep=1pt,topsep=2pt,parsep=0pt]
\item[1.] $\hat h_t \leftarrow (h_t - \bar c) / \lVert h_t - \bar c \rVert$; $s_j \leftarrow \hat h_t^\top M_j$.
\item[2.] $\operatorname{conflict}(t) \leftarrow \exists\, j$ with $o_j = o_t$ and $s_j > r_j$; $j^* \leftarrow \arg\max\, (s_j - r_j)$.
\item[3.] $\psi_t \leftarrow N_{j^*} / (N_{j^*} + n_c / E_{\mathrm{done}})$ if $\operatorname{conflict}(t)$, else $0$.
\item[4.] \textbf{if} $o_t > 0$ and $\operatorname{conflict}(t)$ and the present (class, $o_t$) has been seen in conflict with its gold losing: $a_t \leftarrow \operatorname{clip}(E_{\mathrm{rec}}\, \psi_t, 0, 1)$. \emph{Joint-share target.}
\end{itemize}
\item[3.] $A \leftarrow \{\, t : \operatorname{conflict}(t) \text{ and } z_{\mathrm{gold}} \le z_{\mathrm{rival}} \,\}$. \emph{Comparator positions.}
\item[4.] $p \leftarrow \operatorname{clip}\big(\operatorname{mean}_t (1 - \psi_t), 0, 1\big)$. \emph{Update gate.}
\item[5.] Backward:
\begin{itemize}[leftmargin=1.2em,itemsep=1pt,topsep=2pt,parsep=0pt]
\item[1.] logit gradient $(q - y_t)$, with $y_t$ the joint-share target where $a_t > 0$, else cross-entropy;
\item[2.] for each linear map of synapse group $a$: $\partial L / \partial W \leftarrow \sum_t \delta_t^\top \tilde x_t$, with $\tilde x_t = x_t T_a$ ($t \notin A$) or $X_p[t]$ ($t \in A$);
\item[3.] embedding rows $\times\, D_{\mathrm{tok}}$; gains $\times\, D$ (rebuilt at steps 1, 2, 4, 8, …).
\end{itemize}
\item[6.] Optimizer:
\begin{itemize}[leftmargin=1.2em,itemsep=1pt,topsep=2pt,parsep=0pt]
\item[1.] AdamW moments $m$, $v$ updated with the constrained gradient; $u \leftarrow \hat m / (\sqrt{\hat v} + \epsilon)$;
\item[2.] $u' \leftarrow u\, T_a + (u\, \Gamma_a)\, \tilde Q_a^\top$. \emph{Update constraint, then the comparator's restoration;}
\item[3.] $u_v \leftarrow s_v\, u_v$ (embedding rows); $u \leftarrow u \odot D$ (gains);
\item[4.] with probability $p$: $W \leftarrow W - \eta_m\, u'$. \emph{Otherwise only the moments advance.}
\end{itemize}
\end{itemize}

\textbf{Algorithm 4 | SLEEP: writing and erasing the memory after task $e$.}

\begin{itemize}[leftmargin=1.4em,itemsep=1pt,topsep=2pt,parsep=0pt]
\item[1.] Read 1 (no gradients, inference mode, $\mathcal{D}_e$ only):
\begin{itemize}[leftmargin=1.2em,itemsep=1pt,topsep=2pt,parsep=0pt]
\item[1.] per linear synapse group: episode moment $\hat C_e$; class-level records $N_j$, $\Sigma h_j$, $\Sigma x^a_j$;
\item[2.] gain energies; token use $\nu_v$;
\item[3.] keys $K$, energies $\lambda_K$ and floor $f_e$ from the spectral cutoff of $\hat C_e$.
\end{itemize}
\item[2.] Read 2 (from task 1 on): $\rho_i \leftarrow$ share of key $i$ used by conflicting answer positions.
\item[3.] Gated delta update per synapse group:
\begin{itemize}[leftmargin=1.2em,itemsep=1pt,topsep=2pt,parsep=0pt]
\item[1.] $\beta_e \leftarrow n_e / (n + n_e)$; $B \leftarrow \operatorname{orth}([U, K])$; $S \leftarrow (B^\top U)\, \mathrm{diag}(\lambda_U)\, (B^\top U)^\top$;
\item[2.] erase: $S \leftarrow G S G^\top$, $G = I - \sum_i \big(1 - \sqrt{1 - \beta_e (1 - \rho_i)}\big)\, \kappa_i \kappa_i^\top$;
\item[3.] write: $S \leftarrow S + \beta_e \sum_i \lambda_{K,i}\, \kappa_i \kappa_i^\top$;
\item[4.] prune: keep the eigen-directions of $S$ above $f_e$ as $U$, $\lambda_U$; update $\mathrm{tail}$, $n$, $n^{\mathrm{last}}$, $f$.
\end{itemize}
\item[4.] Discard $\hat C_e$. \emph{No $d \times d$ matrix and no sentence-level state is kept.}
\end{itemize}

\section*{Supplementary Tables}
\begin{table*}[!tp]
\caption{\textbf{Memory-update variants (Llama-3.2-1B, TACRED; one run each).} Each variant rebuilds the memory on its own trajectory.}
{\tabsize\setlength{\tabcolsep}{3pt}\begin{tabularx}{\textwidth}{@{}>{\raggedright\arraybackslash}Xlll>{\raggedright\arraybackslash}X>{\raggedright\arraybackslash}X>{\raggedright\arraybackslash}X@{}}
\toprule
\textbf{Memory update} & \textbf{Final AA} & \textbf{Mean acquisition} & \textbf{Retention} & \textbf{Stored energy, after task 0 $\rightarrow$ 6} & \textbf{Protected subspace (fraction of each synapse group's dimension), after task 0 $\rightarrow$ 6} & \textbf{Residual-to-floor ratio, after task 0 $\rightarrow$ 6}\\
\midrule
Conflict-aware write–erase (DeltaHippo) & 71.27 & 78.8 & 80.8 & 41 $\rightarrow$ 54 & 0.284 $\rightarrow$ 0.291 & 0.35 $\rightarrow$ 0.49\\
No erase (accumulate) & 71.32 & 78.4 & 81.0 & 41 $\rightarrow$ 96 & 0.284 $\rightarrow$ 0.296 & 0.35 $\rightarrow$ 0.49\\
Non-selective erase, energy-matched & 70.12 & 78.3 & 80.5 & 41 $\rightarrow$ 54 & 0.284 $\rightarrow$ 0.292 & 0.35 $\rightarrow$ 0.49\\
\bottomrule
\end{tabularx}}
\par\vspace{2pt}{\fontsize{7bp}{8.5bp}\selectfont Stored energy is the sum of the protected energies $\lambda$ over all synapse groups (relative units); the residual-to-floor ratio is the median over synapse groups of $\tau/f$, which sets the residual plasticity factor $b = f/(f + \mu\tau)$.\par}
\end{table*}

\begin{table*}[!tp]
\caption{\textbf{Common linear probe at the end of the task sequence (Llama-3.2-1B, TACRED).} One run; a logistic-regression probe over all classes seen so far, on the final-layer state before the first answer token; generated answers, the trained model's own generation. Accuracy (\%).}
{\tabsize\setlength{\tabcolsep}{3pt}\begin{tabular*}{\textwidth}{@{\extracolsep{\fill}}lllll@{}}
\toprule
\textbf{At the end of the task sequence} & \textbf{Trained model (all parameters, 8 tasks)} & \textbf{Frozen after task 0} & \textbf{Pre-trained} & \textbf{Generated answers}\\
\midrule
Earlier tasks & 77.2 & 75.4 & 72.0 & 70.1\\
Newest task & 84.0 & 83.0 & 75.5 & 79.8\\
\bottomrule
\end{tabular*}}
\end{table*}

\begin{table*}[!tp]
\caption{\textbf{Layers of the method.}}
{\tabsize\setlength{\tabcolsep}{3pt}\begin{tabularx}{\textwidth}{@{}>{\raggedright\arraybackslash}X>{\raggedright\arraybackslash}X>{\raggedright\arraybackslash}X@{}}
\toprule
\textbf{Layer} & \textbf{Parts} & \textbf{Where}\\
\midrule
Principle & a residual-aware metric of past activations constrains the gradient and the realised update & Operators\\
Core method (class-free; alone 67.1 against 71.6 over three runs) & low-rank-plus-residual activation memory; the protected subspace and a residual floor outside it (Box 1); gradient constraint and post-optimizer update constraint & What the memory stores; Operators; The constrained update\\
Class-incremental extension & class-conditioned index (class state records and input patterns, which feed the class-pattern constraint); state-only conflict detection and comparator; joint-share target (caps a new class's share of a decision region shared with an earlier one) & Operators; Conflict detection; Learning at detected conflicts\\
Memory maintenance & sleep encoding (required, Fig. 3); the gated-delta erase policy (erase lowers the stored energy, for sequences that approach saturation) & Sleep\\
Scaling rules & relative step; width rule; protected basis orthonormalised in fp64; embedding-row and gain plasticity factors & Model training; Operators; Cost\\
\bottomrule
\end{tabularx}}
\end{table*}

\begin{table*}[!tp]
\caption{\textbf{Hard constraints and the regimes they exclude.}}
{\tabsize\setlength{\tabcolsep}{3pt}\begin{tabularx}{\textwidth}{@{}l>{\raggedright\arraybackslash}Xl>{\raggedright\arraybackslash}X@{}}
\toprule
\textbf{\#} & \textbf{Hard constraint} & \textbf{Regime it excludes} & \textbf{Baselines that use it}\\
\midrule
1 & \textbf{No cross-task replay} of stored, generated or reconstructed samples, and no stored per-sentence activations. The only re-reading is one gradient-free encoding phase over the just-finished task, implemented with one or more forward traversals; no earlier task is revisited and none of its samples is kept afterwards & keeping old data & DER++, CLS-ER, LAMOL, PCLL; SEQ* on the encoders\\
2 & \textbf{No teacher, model copy or distillation}: no previous-model snapshot or copy, even read-only, and no distillation from an earlier model & keeping old models & L2KD, LAMOL, PCLL, CLS-ER\\
3 & \textbf{No freezing} of the backbone or of any module: every parameter tensor remains in the optimization; plasticity is constrained directionally and continuously rather than switched off, and selected directions or coordinates may receive an exact zero plasticity factor & not changing the model & SEQ*, EASE, SD-LoRA\\
4 & \textbf{No LoRA, prompts or adapters}, and no task-specific parameters or heads beyond the standard expanding classifier & not changing the model & EASE, SD-LoRA\\
5 & \textbf{No task identity and no memory at test time}: the plain model is evaluated (greedy generation for LLMs, the plain classifier for encoders and vision), with the hippocampus detached & not answering alone & none of the compared baselines\\
\bottomrule
\end{tabularx}}
\end{table*}

\begin{table*}[!tp]
\caption{\textbf{Synapse groups of a transformer layer (Qwen3-0.6B).}}
{\tabsize\setlength{\tabcolsep}{3pt}\begin{tabular*}{\textwidth}{@{\extracolsep{\fill}}lll@{}}
\toprule
\textbf{Synapse group} & \textbf{Linear maps} & \textbf{Input dimension (Qwen3-0.6B)}\\
\midrule
qkv & query, key, value & 1024 (hidden)\\
gu & MLP gate, MLP up & 1024\\
o & attention output & 2048 (heads × head dim)\\
down & MLP down & 3072 (intermediate)\\
\bottomrule
\end{tabular*}}
\end{table*}

\begin{table*}[!tp]
\caption{\textbf{Synapse-group counts per decoder.}}
{\tabsize\setlength{\tabcolsep}{3pt}\begin{tabular*}{\textwidth}{@{\extracolsep{\fill}}llllll@{}}
\toprule
\textbf{Model} & \textbf{Layers} & \textbf{Linear synapse groups} & \textbf{Linear maps} & \textbf{Gain synapse groups} & \textbf{Embedding}\\
\midrule
Qwen3-0.6B & 28 & 113 & 197 & 113 & tied\\
Llama-3.2-1B & 16 & 65 & 113 & 33 & tied\\
Llama-3.2-3B & 28 & 113 & 197 & 57 & tied\\
Qwen3-4B & 36 & 145 & 253 & 145 & tied\\
Qwen3-8B & 36 & 145 & 253 & 145 & untied\\
Llama-3.1-8B & 32 & 129 & 225 & 65 & untied\\
\bottomrule
\end{tabular*}}
\end{table*}

\begin{table*}[!tp]
\caption{\textbf{Responses at detected conflicts.}}
{\tabsize\setlength{\tabcolsep}{3pt}\begin{tabularx}{\textwidth}{@{}l>{\raggedright\arraybackslash}X>{\raggedright\arraybackslash}X>{\raggedright\arraybackslash}X@{}}
\toprule
\textbf{Response} & \textbf{Acts at} & \textbf{What it does} & \textbf{Evidence}\\
\midrule
Update gate & every step, through the conflicts of its answer-token positions & applies the step with probability $p$ & removed with the other class-conditioned parts: 67.1 against 71.6 over three runs (class-free core)\\
Comparator & conflicting positions whose gold does not win yet & gradient and update along the input's mismatch with every earlier pattern & final AA 71.6 $\rightarrow$ 58.7 over three runs; \emph{city of death} answered as \emph{city of residence} 21 times instead of at most 3\\
Joint-share target & answer-token positions after the first, for near-duplicate records seen in conflict while gold lost & drives the gold probability to $1 - a_t$ instead of 1 & removed with the other class-conditioned parts: 67.1 against 71.6 over three runs (class-free core)\\
\bottomrule
\end{tabularx}}
\end{table*}

\begin{table*}[!tp]
\caption{\textbf{Components per architecture.} –, the component is not used in that architecture (ResNet-50 has no embedding table and no attention).}
{\tabsize\setlength{\tabcolsep}{3pt}\begin{tabularx}{\textwidth}{@{}l>{\raggedright\arraybackslash}X>{\raggedright\arraybackslash}X>{\raggedright\arraybackslash}X>{\raggedright\arraybackslash}X@{}}
\toprule
\textbf{Component} & \textbf{LLMs} & \textbf{BERT / RoBERTa} & \textbf{ViT-B/16} & \textbf{ResNet-50}\\
\midrule
Backbone operator & minimum-interference & novelty factor $c\,(I - W W^\top)$ & minimum-interference & novelty factor $c\,(I - W W^\top)$\\
Comparator restoration & full & minimum-interference plasticity factor & minimum-interference plasticity factor & the backbone rule's plasticity factor\\
Comparator active at & conflicts where gold does not win & every input whose gold does not win & conflicts where gold does not win & conflicts where gold does not win\\
Joint-share target & answer-token positions $o_t > 0$ & the decision feature & – & –\\
Embedding rows & $s_v$ (tied: with LM-head operator) & $s_v$ in the units of the residual stream they feed & [CLS]/position rows in residual-stream units & –\\
Weight decay & decoupled & folded into the update before the constraint & folded & folded\\
Attention-sink weight & – & decision feature $\times\, (1 + \sum_i a_{i0}^2)$ & decision feature $\times\, (1 + \sum_i a_{i0}^2)$ & –\\
\bottomrule
\end{tabularx}}
\end{table*}

\begin{table*}[!tp]
\caption{\textbf{Cost per run.}}
{\tabsize\setlength{\tabcolsep}{3pt}\begin{tabular*}{\textwidth}{@{\extracolsep{\fill}}lllllll@{}}
\toprule
\textbf{Run} & \textbf{Tasks} & \textbf{Training} & \textbf{Sleep} & \textbf{Evaluation} & \textbf{Wall time} & \textbf{Peak GPU memory (training / sleep)}\\
\midrule
Llama-3.2-1B, TACRED & 8 & 171 s (45.6\%) & 153 s (40.8\%) & 51 s (13.6\%) & 0.10 h & 33.7 / 25.0 GB\\
Llama-3.2-1B, FewRel & 8 & 634 s (57.3\%) & 250 s (22.5\%) & 224 s (20.2\%) & 0.31 h & 30.6 / 24.9 GB\\
Llama-3.1-8B, TACRED & 8 & 486 s (42.4\%) & 601 s (52.4\%) & 60 s (5.2\%) & 0.32 h & 138.6 / 117.9 GB\\
Llama-3.1-8B, CLINC150 & 15 & 750 s (28.8\%) & 1723 s (66.1\%) & 134 s (5.1\%) & 0.72 h & 144.7 / 137.1 GB\\
\bottomrule
\end{tabular*}}
\end{table*}

\begin{table*}[!tp]
\caption{\textbf{Profiled run (Llama-3.2-1B, TACRED, 8 tasks; seed 42).}}
{\tabsize\setlength{\tabcolsep}{3pt}\begin{tabular*}{\textwidth}{@{\extracolsep{\fill}}ll@{}}
\toprule
\textbf{Item} & \textbf{Value}\\
\midrule
Wall time & 375 s = 0.10 GPU-hours: training 45\%, sleep 41\%, evaluation 14\%\\
Sleep & 13 gradient-free passes over task data in total, 0.82 M tokens (training: 1.44 M tokens over 3 epochs)\\
Linear algebra in sleep & eigendecomposition 89 s (dominated by the 8192 × 8192 MLP down-projection moments), QR 8 s\\
Linear algebra in training & about 1.3 s over 702 steps\\
Peak GPU memory & 33.7 GB training, 25.0 GB sleep\\
Peak host memory & 7.7 GB (subspace memory 1.40 GB fp32 on the host, about 0.7 GB bf16 on the device)\\
Inference & plain architecture without the memory: per-token cost identical to the plain model\\
\bottomrule
\end{tabular*}}
\end{table*}

\begin{table*}[!tp]
\caption{\textbf{Benchmarks.}}
{\tabsize\setlength{\tabcolsep}{3pt}\begin{tabular*}{\textwidth}{@{\extracolsep{\fill}}llll@{}}
\toprule
\textbf{Benchmark} & \textbf{Classes} & \textbf{Tasks} & \textbf{Classes per task}\\
\midrule
TACRED & 40 & 8 & 5\\
FewRel & 80 & 8 & 10\\
Banking77 & 77 & 7 & 11\\
CLINC150 & 150 & 15 & 10\\
CIFAR-100 & 100 & 10 & 10\\
ImageNet-R & 200 & 10 & 20\\
\bottomrule
\end{tabular*}}
\end{table*}

\begin{table*}[!tp]
\caption{\textbf{Correspondence with hippocampal accounts.}}
{\tabsize\setlength{\tabcolsep}{3pt}\begin{tabularx}{\textwidth}{@{}l>{\raggedright\arraybackslash}X>{\raggedright\arraybackslash}X>{\raggedright\arraybackslash}X>{\raggedright\arraybackslash}X@{}}
\toprule
\textbf{Function} & \textbf{Account} & \textbf{Mechanism in DeltaHippo} & \textbf{Evidence in this study} & \textbf{Where}\\
\midrule
Division of labour & \textbf{Complementary learning systems}: long-term knowledge in the cortex; a fast hippocampal trace governs how it is laid down\textsuperscript{65–68} & The network holds all task knowledge and is trained with plain cross-entropy; the hippocampus is read at every wake step and shapes only the update; the plain model answers & Without the read, every earlier task collapses within a quarter of the next task (Fig. 3, condition B) & Model training\\
Index & \textbf{Hippocampal indexing}: the hippocampus stores an index to cortical patterns, not the patterns\textsuperscript{69,70} & Per synapse group, a low-rank subspace with energies and a residual floor; per class, the mean answer state and input pattern. No sample is stored & Rotating the stored directions with the spectrum kept collapses earlier tasks (Fig. 3, condition C) & Synapse groups; What the memory stores\\
Index & \textbf{Sleep-dependent consolidation}: experience is reactivated offline after learning\textsuperscript{71,72} & After each task, a gradient-free re-read of that task writes the index & Blocking one sleep makes its own episode selectively vulnerable (Fig. 3, condition E) & Sleep\\
Regulation & \textbf{Memory strength and metaplasticity}: consolidated memories are harder to overwrite, and a synapse's plasticity depends on its history\textsuperscript{73–75} & Each direction keeps the plasticity factor $f/(f + \mu\lambda)$; the space outside the stored span keeps $f/(f + \mu\tau)$; embedding rows and gains are protected according to their earlier use & Opening the space outside the span loses retention, more so the smaller the explicitly protected span (Extended Data Table 4) & Operators\\
Conflict & \textbf{Match–mismatch novelty detection}\textsuperscript{76,77} & An answer state is in conflict when it falls inside an earlier class's record; only the state is read & Placing the comparator where conflict detection fires raises acquisition by 3.5 points over the same number of other positions (three runs) & Conflict detection\\
Conflict & \textbf{Cholinergic balance of encoding and retrieval}\textsuperscript{78} & A match with a strongly recorded class lowers the update probability & Applied whole with probability $p$; removed with the other class-conditioned parts, final AA falls from 71.6 to 67.1 over three runs (class-free core; Extended Data Table 4) & Learning at detected conflicts\\
Conflict & \textbf{CA1 as a comparator} between recall and input\textsuperscript{77,78} & At conflicting positions whose gold does not yet win, the update learns along the mismatch of the input with all earlier class patterns & Without it, a task whose classes collide with earlier ones is not learnt (task 6: \emph{city of death}; final AA \textminus{}12.9 over three runs) & Learning at detected conflicts; The constrained update\\
Conflict & Protecting what the index points to (computational device; no specific counterpart) & The base update is exactly orthogonal to earlier class patterns outside the stored span (class-pattern constraint); the joint-share target keeps a new class from taking over a decision region it shares with an earlier one & Removed with the other class-conditioned parts, the established class falls to 0/41 in every run (Fig. 4d–f) & Operators; Learning at detected conflicts\\
Capacity & \textbf{Forgetting that keeps a limited store usable}\textsuperscript{62–64} via the \textbf{gated delta rule}\textsuperscript{36,38} & Sleep erases stored energy along new keys, less where conflicting positions relied on them & Erase lowers stored energy (54 against 96); it is designed for sequences that saturate the store & Sleep\\
\bottomrule
\end{tabularx}}
\end{table*}

\begin{table*}[!tp]
\caption{\textbf{Decoder language models, every run.} Final average accuracy (\%) of each run, seeds 42 / 11 / 3, after the last task; the means and standard deviations of these runs are those reported in the main tables.}
{\tabsize\setlength{\tabcolsep}{3pt}\begin{tabular*}{\textwidth}{@{\extracolsep{\fill}}llllll@{}}
\toprule
\textbf{Model} & \textbf{Method} & \textbf{TACRED} & \textbf{FewRel} & \textbf{Banking77} & \textbf{CLINC150}\\
\midrule
Qwen3-0.6B & DeltaHippo & 70.92 / 74.43 / 72.01 & 69.39 / 71.76 / 71.88 & 83.38 / 82.60 / 82.76 & 87.91 / 83.87 / 89.62\\
 & SEQ* & 52.12 / 54.09 / 53.83 & 69.17 / 69.62 / 68.85 & 72.50 / 75.49 / 75.42 & 87.96 / 87.78 / 87.62\\
 & SEQ & 40.85 / 34.41 / 36.35 & 24.22 / 25.53 / 23.62 & 19.94 / 19.64 / 20.13 & 40.40 / 38.42 / 38.18\\
 & L2KD & 11.21 / 9.44 / 11.34 & 11.23 / 13.08 / 11.79 & 17.66 / 17.53 / 17.37 & 25.09 / 24.31 / 25.09\\
 & LAMOL & 32.96 / 33.53 / 33.77 & 29.24 / 26.66 / 26.85 & 31.04 / 29.74 / 27.18 & 54.29 / 54.89 / 56.09\\
 & PCLL & 33.78 / 31.41 / 33.29 & 23.71 / 25.03 / 25.83 & 34.16 / 39.32 / 37.99 & 56.29 / 58.11 / 53.64\\
 & Sculpting & 41.06 / 41.03 / 41.89 & 42.55 / 40.51 / 38.09 & 38.12 / 37.89 / 35.97 & 67.96 / 62.04 / 61.73\\
 & CODE-CL & 44.52 / 39.73 / 35.74 & 37.35 / 31.26 / 35.72 & 46.92 / 51.10 / 43.80 & 71.31 / 72.60 / 67.87\\
Llama-3.2-1B & DeltaHippo & 71.27 / 72.14 / 71.31 & 81.33 / 76.63 / 79.29 & 80.97 / 82.92 / 82.95 & 88.60 / 86.82 / 85.71\\
 & SEQ* & 55.93 / 53.33 / 55.58 & 72.34 / 72.37 / 72.96 & 80.88 / 79.74 / 80.23 & 90.47 / 90.40 / 90.38\\
 & SEQ & 36.73 / 31.82 / 36.71 & 36.77 / 39.87 / 39.96 & 24.77 / 26.01 / 23.93 & 57.69 / 59.27 / 57.78\\
 & L2KD & 21.15 / 20.50 / 21.48 & 33.68 / 29.33 / 34.51 & 21.49 / 20.52 / 20.71 & 58.29 / 58.98 / 57.93\\
 & LAMOL & 55.73 / 50.60 / 51.55 & 41.37 / 39.84 / 45.68 & 64.35 / 63.41 / 65.10 & 74.49 / 71.93 / 72.16\\
 & PCLL & 40.34 / 33.59 / 32.98 & 43.93 / 44.80 / 41.62 & 46.36 / 42.66 / 41.88 & 63.24 / 66.11 / 63.62\\
 & Sculpting & 34.78 / 24.90 / 19.93 & 39.06 / 36.08 / 44.02 & 33.93 / 39.77 / 33.18 & 52.80 / 61.80 / 58.42\\
 & CODE-CL & 69.20 / 69.28 / 47.83 & 60.61 / 57.33 / 56.19 & 60.78 / 57.01 / 50.94 & 85.58 / 81.33 / 81.71\\
Qwen3-4B & DeltaHippo & 76.15 / 78.12 / 75.27 & 79.69 / 78.47 / 71.39 & 86.69 / 87.89 / 87.79 & 92.42 / 92.47 / 92.11\\
 & SEQ* & 57.70 / 56.71 / 57.65 & 77.49 / 77.36 / 77.88 & 78.67 / 79.51 / 77.27 & 90.49 / 90.62 / 90.62\\
 & SEQ & 42.83 / 44.74 / 43.28 & 40.38 / 40.10 / 38.57 & 33.67 / 31.07 / 31.66 & 62.09 / 59.89 / 63.93\\
Llama-3.2-3B & DeltaHippo & 75.99 / 75.40 / 77.81 & 80.33 / 83.30 / 80.79 & 85.97 / 85.32 / 86.23 & 92.51 / 91.69 / 90.84\\
 & SEQ* & 55.49 / 53.60 / 52.87 & 75.26 / 76.55 / 75.73 & 81.43 / 80.55 / 81.36 & 91.84 / 91.76 / 92.18\\
 & SEQ & 43.93 / 40.88 / 39.47 & 42.99 / 41.05 / 42.05 & 30.13 / 28.64 / 28.83 & 59.18 / 61.44 / 59.20\\
Qwen3-8B & DeltaHippo & 74.32 / 78.18 / 77.86 & 85.04 / 86.29 / 86.52 & 88.05 / 88.64 / 88.38 & 92.82 / 94.27 / 94.00\\
 & SEQ* & 58.68 / 59.84 / 59.45 & 77.68 / 77.82 / 78.31 & 78.57 / 75.49 / 78.05 & 92.02 / 92.58 / 92.27\\
 & SEQ & 43.63 / 45.49 / 40.83 & 41.74 / 38.64 / 38.50 & 32.89 / 31.82 / 35.16 & 69.13 / 70.87 / 65.64\\
Llama-3.1-8B & DeltaHippo & 76.80 / 78.12 / 77.13 & 83.84 / 80.95 / 83.97 & 87.31 / 88.38 / 87.73 & 93.29 / 92.36 / 91.93\\
 & SEQ* & 67.13 / 64.34 / 69.37 & 80.01 / 80.01 / 81.43 & 84.48 / 84.81 / 84.48 & 93.78 / 92.93 / 92.53\\
 & SEQ & 41.25 / 33.64 / 37.95 & 46.46 / 47.79 / 48.28 & 29.55 / 30.58 / 29.19 & 67.93 / 62.09 / 65.44\\
\bottomrule
\end{tabular*}}
\end{table*}

\begin{table*}[!tp]
\caption{\textbf{Text encoders, every run.} Final average accuracy (\%) of each run, seeds 42 / 11 / 3, after the last task; the means and standard deviations of these runs are those reported in the main tables.}
{\tabsize\setlength{\tabcolsep}{3pt}\begin{tabular*}{\textwidth}{@{\extracolsep{\fill}}llllll@{}}
\toprule
\textbf{Model} & \textbf{Method} & \textbf{TACRED} & \textbf{FewRel} & \textbf{Banking77} & \textbf{CLINC150}\\
\midrule
BERT-base & DeltaHippo & 62.76 / 63.18 / 61.93 & 69.17 / 63.64 / 68.14 & 85.07 / 85.19 / 84.81 & 90.11 / 90.36 / 90.80\\
 & SEQ & 18.00 / 17.81 / 18.32 & 15.85 / 17.84 / 18.53 & 20.39 / 20.32 / 20.65 & 14.13 / 15.58 / 15.20\\
 & SEQ* & 28.98 / 31.36 / 29.69 & 47.48 / 45.81 / 51.25 & 56.62 / 61.01 / 57.73 & 63.18 / 63.73 / 60.58\\
 & DER++ & 41.44 / 43.04 / 41.55 & 50.59 / 48.78 / 52.35 & 53.70 / 55.03 / 54.12 & 75.93 / 76.00 / 73.73\\
 & CLS-ER & 45.58 / 48.39 / 40.94 & 52.71 / 53.88 / 50.41 & 58.64 / 59.61 / 60.06 & 73.22 / 76.42 / 76.11\\
RoBERTa-base & DeltaHippo & 55.32 / 44.52 / 59.89 & 75.26 / 67.59 / 71.79 & 83.99 / 85.68 / 83.34 & 91.49 / 92.20 / 91.98\\
 & SEQ & 15.25 / 13.51 / 15.08 & 14.00 / 17.21 / 15.15 & 19.09 / 21.01 / 19.06 & 14.96 / 13.91 / 14.44\\
 & SEQ* & 26.55 / 26.34 / 30.35 & 47.33 / 48.85 / 50.97 & 65.39 / 66.43 / 66.07 & 68.80 / 71.09 / 67.51\\
 & DER++ & 43.31 / 44.09 / 43.55 & 45.80 / 48.37 / 49.46 & 60.91 / 61.33 / 62.01 & 80.42 / 81.98 / 79.11\\
 & CLS-ER & 43.35 / 48.13 / 45.68 & 53.47 / 49.90 / 52.01 & 64.97 / 67.27 / 67.79 & 82.20 / 81.42 / 81.80\\
\bottomrule
\end{tabular*}}
\end{table*}

\begin{table*}[!tp]
\caption{\textbf{Vision models, every run.} Final average accuracy (\%) of each run, seeds 1993 / 42 / 3, after the last of ten tasks.}
{\tabsize\setlength{\tabcolsep}{3pt}\begin{tabular*}{\textwidth}{@{\extracolsep{\fill}}llll@{}}
\toprule
\textbf{Model} & \textbf{Method} & \textbf{CIFAR-100} & \textbf{ImageNet-R}\\
\midrule
ViT-B/16 & DeltaHippo & 88.49 / 88.14 / 88.00 & 63.94 / 64.12 / 64.46\\
 & SEQ & 22.16 / 21.11 / 20.47 & 23.88 / 22.69 / 23.14\\
 & SEQ* & 81.81 / 81.15 / 81.59 & 64.37 / 64.47 / 63.85\\
 & DER++ & 48.01 / 45.01 / 44.48 & 41.58 / 43.46 / 43.74\\
 & CLS-ER & 79.55 / 81.56 / 80.15 & 52.58 / 51.57 / 51.23\\
 & EASE & 86.99 / 87.04 / 87.02 & 67.69 / 68.01 / 67.74\\
 & SD-LoRA & 87.44 / 87.67 / 87.30 & 66.63 / 66.47 / 66.60\\
ResNet-50 & DeltaHippo & 68.09 / 68.96 / 67.54 & 61.92 / 61.31 / 60.69\\
 & SEQ & 8.86 / 9.16 / 9.21 & 8.35 / 8.81 / 8.48\\
 & SEQ* & 46.55 / 45.31 / 48.66 & 41.22 / 40.84 / 39.77\\
 & DER++ & 17.78 / 18.21 / 18.90 & 23.84 / 23.77 / 23.86\\
 & CLS-ER & 53.70 / 54.52 / 55.59 & 40.39 / 40.28 / 39.62\\
\bottomrule
\end{tabular*}}
\end{table*}

\begin{table*}[!tp]
\caption{\textbf{The core ablations, every run} (Llama-3.2-1B, TACRED; seeds 42 / 11 / 3). Final AA, mean acquisition and retention as in Extended Data Table 4.}
{\tabsize\setlength{\tabcolsep}{3pt}\begin{tabular*}{\textwidth}{@{\extracolsep{\fill}}llll@{}}
\toprule
\textbf{Variant} & \textbf{Final AA} & \textbf{Mean acquisition} & \textbf{Retention}\\
\midrule
DeltaHippo & 71.27 / 72.14 / 71.31 & 78.81 / 75.50 / 79.24 & 80.80 / 84.02 / 81.46\\
Class-free geometric core only & 69.57 / 65.82 / 65.83 & 73.96 / 70.48 / 78.64 & 82.72 / 82.16 / 77.86\\
No update constraint (gradient constrained only) & 60.05 / 64.37 / 62.01 & 85.32 / 88.20 / 88.47 & 56.75 / 63.30 / 60.84\\
No gradient constraint (update constraint only) & 68.23 / 42.40 / 58.62 & 87.28 / 92.15 / 92.53 & 60.05 / 50.18 / 49.11\\
Hard constraint inside the protected subspace, residual floor kept & 70.74 / 69.82 / 72.66 & 76.40 / 73.05 / 78.34 & 82.46 / 83.18 / 82.96\\
Graded inside, residual subspace fully open & 69.77 / 74.02 / 69.76 & 85.37 / 86.65 / 83.96 & 77.53 / 80.51 / 79.42\\
Top 25\% of stored directions protected, residual floor kept & 66.51 / 70.62 / 68.87 & 76.15 / 77.09 / 79.22 & 78.13 / 81.95 / 78.88\\
Top 25\% of stored directions protected, residual subspace open & 52.44 / 53.16 / 43.70 & 91.91 / 91.24 / 90.98 & 51.37 / 57.88 / 49.79\\
Comparator off & 59.31 / 58.63 / 58.21 & 59.92 / 61.75 / 60.20 & 81.25 / 81.12 / 79.87\\
\bottomrule
\end{tabular*}}
\end{table*}

\begin{table*}[!tp]
\caption{\textbf{Class order, every run} (Llama-3.2-1B, TACRED; seeds 42 / 11 / 3). Final average accuracy (\%) under the protocol order and two further fixed random orders of the 40 relations, the same for both methods.}
{\tabsize\setlength{\tabcolsep}{3pt}\begin{tabular*}{\textwidth}{@{\extracolsep{\fill}}llll@{}}
\toprule
\textbf{Order} & \textbf{DeltaHippo} & \textbf{SEQ*} & \textbf{Margin (mean)}\\
\midrule
Protocol order & 71.27 / 72.14 / 71.31 (71.58 ± 0.49) & 55.93 / 53.33 / 55.58 (54.95 ± 1.41) & +16.6\\
Further order 1 & 72.70 / 74.97 / 69.05 (72.24 ± 2.99) & 58.31 / 59.68 / 59.75 (59.25 ± 0.81) & +13.0\\
Further order 2 & 71.32 / 72.42 / 74.11 (72.62 ± 1.41) & 54.41 / 54.75 / 55.14 (54.77 ± 0.36) & +17.9\\
\bottomrule
\end{tabular*}}
\end{table*}

\begin{table*}[!tp]
\caption{\textbf{Generation and linear probes after task 4 under the memory interventions (Llama-3.2-1B, TACRED).} One run; logistic-regression probes on final-layer pre-answer states, fitted on the seen tasks' training sets and evaluated on their test sets. The pre-trained model (before task 0) and the backbone frozen after task 0 are references for the representation alone: neither is trained on tasks 1–4, so generation and the normal run's probe do not apply to them (–).}
{\tabsize\setlength{\tabcolsep}{3pt}\begin{tabular*}{\textwidth}{@{\extracolsep{\fill}}llll@{}}
\toprule
\textbf{Condition} & \textbf{Generation, tasks 0–2 / 3} & \textbf{Own probe, tasks 0–2 / 3} & \textbf{Normal run's probe applied, tasks 0–2 / 3}\\
\midrule
A. Normal & 81.6 / 69.1 & 85.9 / 75.7 & 85.9 / 75.7\\
B. Read off & 0.0 / 0.0 & 81.4 / 70.7 & 10.6 / 2.8\\
C. Scrambled content & 1.2 / 0.0 & 87.5 / 79.6 & 27.4 / 34.3\\
F. Norm-matched plain & 0.0 / 0.0 & 85.1 / 75.7 & 28.7 / 8.3\\
Pre-trained model & – & 78.3 / 71.8 & –\\
Backbone frozen after task 0 & – & 83.4 / 74.6 & –\\
\bottomrule
\end{tabular*}}
\end{table*}

\begin{table*}[!tp]
\caption{\textbf{Restricted recovery of lost access (Llama-3.2-1B, TACRED, tasks 0–3 at the end of task 4).} Each condition's final norm and output rows are fixed; the maps are fitted on the earlier tasks' training sets and evaluated on their test sets (label-level accuracy, \%). In brackets, the share of the gap between the condition's own head and a re-fitted probe that the map closes. Bias, one bias per label-token row; bias + temperature, one logit scale in addition; rank-$K$ map, $h \to h + (hV)U^\top$ on the last-layer state. Whole model, backbone only and head only: mean ± sample standard deviation over the three runs of each guidance setting (Fig. 5a,b); read off, scrambled and norm-matched: the task-4 intervention conditions, one run each. Own head and maps: mean over tasks 0–3 (Fig. 5b pools the 745 answers). Re-fitted probe: a class-level logistic-regression probe on the first answer position's state, as in Supplementary Table 18.}
{\tabsize\setlength{\tabcolsep}{3pt}\begin{tabular*}{\textwidth}{@{\extracolsep{\fill}}llllllll@{}}
\toprule
\textbf{Condition} & \textbf{Own head} & \textbf{Bias} & \textbf{Bias + temperature} & \textbf{Rank-1 map} & \textbf{Rank-4 map} & \textbf{Rank-16 map} & \textbf{Re-fitted probe}\\
\midrule
Whole model & 80.6 ± 2.8 & 82.6 ± 2.0 (0.25) & 82.9 ± 1.9 (0.29) & 85.4 ± 0.7 (0.60) & 88.4 ± 1.1 (0.97) & 88.6 ± 0.1 (1.00) & 88.6 ± 0.5\\
Backbone only & 24.7 ± 12.5 & 47.8 ± 17.8 (0.37) & 56.5 ± 17.2 (0.50) & 84.4 ± 1.7 (0.95) & 86.5 ± 1.4 (0.98) & 88.1 ± 0.2 (1.00) & 87.8 ± 0.2\\
Head only & 0.6 ± 1.1 & 1.2 ± 1.1 (0.01) & 3.3 ± 2.8 (0.03) & 28.9 ± 6.0 (0.32) & 75.5 ± 1.6 (0.85) & 83.4 ± 0.5 (0.94) & 88.9 ± 0.2\\
Read off & 0.0 & 0.0 (0.00) & 0.0 (0.00) & 26.8 (0.32) & 73.4 (0.87) & 76.6 (0.90) & 84.8\\
Scrambled content & 0.9 & 3.5 (0.03) & 11.3 (0.12) & 54.6 (0.61) & 86.7 (0.97) & 86.3 (0.97) & 89.4\\
Norm-matched plain & 0.0 & 5.4 (0.06) & 9.7 (0.11) & 56.2 (0.63) & 85.9 (0.97) & 87.6 (0.99) & 88.5\\
\bottomrule
\end{tabular*}}
\end{table*}

\begin{table*}[!tp]
\caption{\textbf{A low-rank correction does not transfer to classes it was not fitted on (Llama-3.2-1B, TACRED, tasks 0–3 at the end of task 4).} The classes of tasks 0–3 are split into two fixed groups (12 and 8 classes). Maps as in Supplementary Table 19 are fitted on the other group (held-out) or on the evaluated group (same classes) and evaluated on the evaluated group's test sentences with the condition's own head, teacher-forced and label-level (every answer token must be correct). Own head and map columns: each answer token is the argmax over the full vocabulary; restricted decoding: the argmax over the evaluated group's label tokens only; re-fitted probe: a class-level logistic-regression probe on the first answer position. Accuracy (\%), mean over three runs (read off: one run).}
{\tabsize\setlength{\tabcolsep}{3pt}\begin{tabular*}{\textwidth}{@{\extracolsep{\fill}}llllll@{}}
\toprule
\textbf{Condition} & \textbf{Own head} & \textbf{Own head, restricted decoding} & \textbf{Re-fitted probe} & \textbf{Held-out rank-1 / 4 / 16} & \textbf{Same-class rank-1 / 4 / 16}\\
\midrule
Whole model & 84.9 & 92.0 & 89.4 & 33.4 / 35.6 / 34.5 & 94.2 / 95.5 / 96.3\\
Backbone only & 29.4 & 72.6 & 89.7 & 20.5 / 17.9 / 21.5 & 94.3 / 95.3 / 95.9\\
Head only & 2.0 & 14.9 & 90.3 & 0.3 / 0.3 / 0.0 & 45.6 / 92.3 / 95.2\\
Read off & 0.0 & 4.3 & 86.1 & 0.0 / 0.0 / 0.0 & 44.2 / 90.7 / 92.0\\
\bottomrule
\end{tabular*}}
\end{table*}

\section*{Supplementary References}
{\fontsize{7.5bp}{9bp}\selectfont
\begin{itemize}[leftmargin=2.2em,labelsep=0.5em,itemsep=0.5pt,topsep=2pt,parsep=0pt]
\item[1.] Zhou, D.-W., Cai, Z.-W., Ye, H.-J., Zhan, D.-C. \& Liu, Z. Revisiting class-incremental learning with pre-trained models: generalizability and adaptivity are all you need (SimpleCIL / APER). \emph{Int. J. Comput. Vis.} (2024).
\item[2.] McDonnell, M. D., Gong, D., Parvaneh, A., Abbasnejad, E. \& van den Hengel, A. RanPAC: Random projections and pre-trained models for continual learning. In \emph{Advances in Neural Information Processing Systems} (2023).
\item[3.] Wang, Z. et al. Learning to prompt for continual learning. In \emph{Proc. IEEE/CVF Conference on Computer Vision and Pattern Recognition} (2022).
\item[4.] Wang, Z. et al. DualPrompt: Complementary prompting for rehearsal-free continual learning. In \emph{Proc. European Conference on Computer Vision} (2022).
\item[5.] Smith, J. S. et al. CODA-Prompt: COntinual decomposed attention-based prompting for rehearsal-free continual learning. In \emph{Proc. IEEE/CVF Conference on Computer Vision and Pattern Recognition} (2023).
\item[6.] Wang, L. et al. Hierarchical decomposition of prompt-based continual learning: rethinking obscured sub-optimality. In \emph{Advances in Neural Information Processing Systems} (2023).
\item[7.] Gao, Z., Cen, J. \& Chang, X. Consistent prompting for rehearsal-free continual learning. In \emph{Proc. IEEE/CVF Conference on Computer Vision and Pattern Recognition} (2024).
\item[8.] Zhou, D.-W., Sun, H.-L., Ye, H.-J. \& Zhan, D.-C. Expandable subspace ensemble for pre-trained model-based class-incremental learning. In \emph{Proc. IEEE/CVF Conference on Computer Vision and Pattern Recognition} (2024).
\item[9.] Tan, Y. et al. Semantically-shifted incremental adapter-tuning is a continual ViTransformer. In \emph{Proc. IEEE/CVF Conference on Computer Vision and Pattern Recognition} (2024).
\item[10.] Sun, H.-L. et al. MOS: Model surgery for pre-trained model-based class-incremental learning. In \emph{Proc. AAAI Conference on Artificial Intelligence} (2025).
\item[11.] Wang, Y., Zhou, D.-W. \& Ye, H.-J. Integrating task-specific and universal adapters for pre-trained model-based class-incremental learning. In \emph{Proc. IEEE/CVF Conference on Computer Vision and Pattern Recognition} (2025).
\item[12.] Liang, Y.-S. \& Li, W.-J. InfLoRA: Interference-free low-rank adaptation for continual learning. In \emph{Proc. IEEE/CVF Conference on Computer Vision and Pattern Recognition} (2024).
\item[13.] Wu, Y. et al. SD-LoRA: Scalable decoupled low-rank adaptation for class incremental learning. In \emph{Proc. International Conference on Learning Representations} (2025).
\item[14.] Wang, X. et al. Orthogonal subspace learning for language model continual learning. In \emph{Findings of the Association for Computational Linguistics: EMNLP} (2023).
\item[15.] Kirkpatrick, J. et al. Overcoming catastrophic forgetting in neural networks. \emph{Proc. Natl Acad. Sci. USA} \textbf{114}, 3521–3526 (2017).
\item[16.] Zenke, F., Poole, B. \& Ganguli, S. Continual learning through synaptic intelligence. In \emph{Proc. International Conference on Machine Learning} (2017).
\item[17.] Zeng, G., Chen, Y., Cui, B. \& Yu, S. Continual learning of context-dependent processing in neural networks. \emph{Nat. Mach. Intell.} \textbf{1}, 364–372 (2019).
\item[18.] Farajtabar, M., Azizan, N., Mott, A. \& Li, A. Orthogonal gradient descent for continual learning. In \emph{Proc. International Conference on Artificial Intelligence and Statistics} (2020).
\item[19.] Saha, G., Garg, I. \& Roy, K. Gradient projection memory for continual learning. In \emph{Proc. International Conference on Learning Representations} (2021).
\item[20.] Wang, S., Li, X., Sun, J. \& Xu, Z. Training networks in null space of feature covariance for continual learning. In \emph{Proc. IEEE/CVF Conference on Computer Vision and Pattern Recognition} (2021).
\item[21.] Saha, G. \& Roy, K. Continual learning with scaled gradient projection. In \emph{Proc. AAAI Conference on Artificial Intelligence} \textbf{37}, 9677–9685 (2023).
\item[22.] Wang, S., Li, X., Sun, J. \& Xu, Z. Training networks in null space of feature covariance with self-supervision for incremental learning (Adam-NSCL, Adam-SFCL). \emph{IEEE Trans. Pattern Anal. Mach. Intell.} \textbf{47}, 2563–2580 (2025).
\item[23.] Zhang, Y., Chen, X., Ren, C. \& Zhang, Q. A tropical geometry view of forgetting: a per-unit projector for knowledge-preserving fine-tuning. Preprint at \url{https://arxiv.org/abs/2610.04670} (2026).
\item[24.] Apolinario, M. P. E., Choudhary, S. \& Roy, K. CODE-CL: Conceptor-based gradient projection for deep continual learning. In \emph{Proc. IEEE/CVF International Conference on Computer Vision} (2025).
\item[25.] Nayak, N. S. et al. Sculpting subspaces: constrained full fine-tuning in LLMs for continual learning. Preprint at \url{https://arxiv.org/abs/2504.07097} (2025).
\item[26.] Cheng, Q., Wan, Y., Wu, L., Hou, C. \& Zhang, L. Continuous subspace optimization for continual learning. In \emph{Advances in Neural Information Processing Systems} (2025).
\item[27.] Zheng, J., Cai, X., Qiu, S. \& Ma, Q. Spurious forgetting in continual learning of language models. In \emph{Proc. International Conference on Learning Representations} (2025).
\item[28.] Palit, V., Draye, F., Zucchet, N., Jin, Z. \& Schölkopf, B. When forgetting is not catastrophic: on the mechanics of spurious forgetting. Preprint at \url{https://arxiv.org/abs/2610.08718} (2026).
\item[29.] Cosentino, R. PLATE: Plasticity-tunable efficient adapters for geometry-aware continual learning. Preprint at \url{https://arxiv.org/abs/2602.03846} (2026).
\item[30.] Luo, H. et al. Representation finetuning for continual learning. Preprint at \url{https://arxiv.org/abs/2603.11201} (2026).
\item[31.] Yoo, S., Kim, K., Joo, H., Bang, J. \& Kim, H. SPACE-LoRA: Allocating activation-subspace protection for continual learning. Preprint at \url{https://arxiv.org/abs/2609.34453} (2026).
\item[32.] Wang, Y. et al. Geometry Conflict: Explaining and controlling forgetting in LLM continual post-training. Preprint at \url{https://arxiv.org/abs/2605.09608} (2026).
\item[33.] Zheng, J., Qiu, S. \& Ma, Q. Learn or recall? Revisiting incremental learning with pre-trained language models. In \emph{Proc. Annual Meeting of the Association for Computational Linguistics} (2024).
\item[34.] Chen, Y. et al. HippoTune: A hippocampal associative loop–inspired fine-tuning method for continual learning. In \emph{Proc. International Conference on Learning Representations} (2026).
\item[35.] Kapoor, K., Mackey, W., Aloimonos, Y. \& Lin, X. HiCL: Hippocampal-inspired continual learning. In \emph{Proc. AAAI Conference on Artificial Intelligence} (2026).
\item[36.] Widrow, B. \& Hoff, M. E. Adaptive switching circuits. In \emph{IRE WESCON Convention Record} Part 4, 96–104 (1960).
\item[37.] Schlag, I., Irie, K. \& Schmidhuber, J. Linear transformers are secretly fast weight programmers. In \emph{Proc. International Conference on Machine Learning} (2021).
\item[38.] Yang, S., Kautz, J. \& Hatamizadeh, A. Gated delta networks: improving Mamba2 with delta rule. In \emph{Proc. International Conference on Learning Representations} (2025).
\item[39.] Sun, F.-K., Ho, C.-H. \& Lee, H.-Y. LAMOL: Language modeling for lifelong language learning. In \emph{Proc. International Conference on Learning Representations} (2020).
\item[40.] Zhao, Y. et al. Prompt conditioned VAE: enhancing generative replay for lifelong learning in task-oriented dialogue. In \emph{Proc. Conference on Empirical Methods in Natural Language Processing} (2022).
\item[41.] Chuang, Y.-S., Su, S.-Y. \& Chen, Y.-N. Lifelong language knowledge distillation. In \emph{Proc. Conference on Empirical Methods in Natural Language Processing} (2020).
\item[42.] Buzzega, P., Boschini, M., Porrello, A., Abati, D. \& Calderara, S. Dark experience for general continual learning: a strong, simple baseline. In \emph{Advances in Neural Information Processing Systems} (2020).
\item[43.] Arani, E., Sarfraz, F. \& Zonooz, B. Learning fast, learning slow: a general continual learning method based on complementary learning system. In \emph{Proc. International Conference on Learning Representations} (2022).
\item[44.] Zhang, G., Wang, L., Kang, G., Chen, L. \& Wei, Y. SLCA: Slow learner with classifier alignment for continual learning on a pre-trained model. In \emph{Proc. IEEE/CVF International Conference on Computer Vision} (2023).
\item[45.] Yang, G. et al. Tensor Programs V: Tuning large neural networks via zero-shot hyperparameter transfer. Preprint at \url{https://arxiv.org/abs/2203.03466} (2022).
\item[46.] van de Ven, G. M. \& Tolias, A. S. Three scenarios for continual learning. Preprint at \url{https://arxiv.org/abs/1904.07734} (2019).
\item[47.] Kim, G., Xiao, C., Konishi, T., Ke, Z. \& Liu, B. A theoretical study on solving continual learning. In \emph{Advances in Neural Information Processing Systems} (2022).
\item[48.] Zhang, Y., Zhong, V., Chen, D., Angeli, G. \& Manning, C. D. Position-aware attention and supervised data improve slot filling. In \emph{Proc. Conference on Empirical Methods in Natural Language Processing} (2017).
\item[49.] Han, X. et al. FewRel: a large-scale supervised few-shot relation classification dataset with state-of-the-art evaluation. In \emph{Proc. Conference on Empirical Methods in Natural Language Processing} (2018).
\item[50.] Casanueva, I., Temčinas, T., Gerz, D., Henderson, M. \& Vulić, I. Efficient intent detection with dual sentence encoders. In \emph{Proc. 2nd Workshop on Natural Language Processing for Conversational AI} (2020).
\item[51.] Larson, S. et al. An evaluation dataset for intent classification and out-of-scope prediction. In \emph{Proc. Conference on Empirical Methods in Natural Language Processing and the International Joint Conference on Natural Language Processing} (2019).
\item[52.] Krizhevsky, A. Learning multiple layers of features from tiny images. Technical report, Univ. Toronto (2009).
\item[53.] Hendrycks, D. et al. The many faces of robustness: a critical analysis of out-of-distribution generalization. In \emph{Proc. IEEE/CVF International Conference on Computer Vision} (2021).
\item[54.] Zhou, D.-W., Wang, F.-Y., Ye, H.-J. \& Zhan, D.-C. PyCIL: a Python toolbox for class-incremental learning. \emph{Sci. China Inf. Sci.} \textbf{66}, 197101 (2023).
\item[55.] Yang, A. et al. Qwen3 technical report. Preprint at \url{https://arxiv.org/abs/2505.09388} (2025).
\item[56.] Grattafiori, A. et al. The Llama 3 herd of models. Preprint at \url{https://arxiv.org/abs/2407.21783} (2024).
\item[57.] Devlin, J., Chang, M.-W., Lee, K. \& Toutanova, K. BERT: pre-training of deep bidirectional transformers for language understanding. In \emph{Proc. Conference of the North American Chapter of the Association for Computational Linguistics} (2019).
\item[58.] Liu, Y. et al. RoBERTa: a robustly optimized BERT pretraining approach. Preprint at \url{https://arxiv.org/abs/1907.11692} (2019).
\item[59.] Dosovitskiy, A. et al. An image is worth 16x16 words: transformers for image recognition at scale. In \emph{Proc. International Conference on Learning Representations} (2021).
\item[60.] He, K., Zhang, X., Ren, S. \& Sun, J. Deep residual learning for image recognition. In \emph{Proc. IEEE/CVF Conference on Computer Vision and Pattern Recognition} (2016).
\item[61.] Gavish, M. \& Donoho, D. L. The optimal hard threshold for singular values is 4/$\surd$3. \emph{IEEE Trans. Inf. Theory} \textbf{60}, 5040–5053 (2014).
\item[62.] Tononi, G. \& Cirelli, C. Sleep and the price of plasticity: from synaptic and cellular homeostasis to memory consolidation and integration. \emph{Neuron} \textbf{81}, 12–34 (2014).
\item[63.] Akers, K. G. et al. Hippocampal neurogenesis regulates forgetting during adulthood and infancy. \emph{Science} \textbf{344}, 598–602 (2014).
\item[64.] Richards, B. A. \& Frankland, P. W. The persistence and transience of memory. \emph{Neuron} \textbf{94}, 1071–1084 (2017).
\item[65.] McClelland, J. L., McNaughton, B. L. \& O'Reilly, R. C. Why there are complementary learning systems in the hippocampus and neocortex. \emph{Psychol. Rev.} \textbf{102}, 419–457 (1995).
\item[66.] Squire, L. R. \& Alvarez, P. Retrograde amnesia and memory consolidation: a neurobiological perspective. \emph{Curr. Opin. Neurobiol.} \textbf{5}, 169–177 (1995).
\item[67.] Frankland, P. W. \& Bontempi, B. The organization of recent and remote memories. \emph{Nat. Rev. Neurosci.} \textbf{6}, 119–130 (2005).
\item[68.] Kumaran, D., Hassabis, D. \& McClelland, J. L. What learning systems do intelligent agents need? \emph{Trends Cogn. Sci.} \textbf{20}, 512–534 (2016).
\item[69.] Teyler, T. J. \& DiScenna, P. The hippocampal memory indexing theory. \emph{Behav. Neurosci.} \textbf{100}, 147–154 (1986).
\item[70.] Teyler, T. J. \& Rudy, J. W. The hippocampal indexing theory and episodic memory: updating the index. \emph{Hippocampus} \textbf{17}, 1158–1169 (2007).
\item[71.] Diekelmann, S. \& Born, J. The memory function of sleep. \emph{Nat. Rev. Neurosci.} \textbf{11}, 114–126 (2010).
\item[72.] Rasch, B. \& Born, J. About sleep's role in memory. \emph{Physiol. Rev.} \textbf{93}, 681–766 (2013).
\item[73.] Fusi, S., Drew, P. J. \& Abbott, L. F. Cascade models of synaptically stored memories. \emph{Neuron} \textbf{45}, 599–611 (2005).
\item[74.] Benna, M. K. \& Fusi, S. Computational principles of synaptic memory consolidation. \emph{Nat. Neurosci.} \textbf{19}, 1697–1706 (2016).
\item[75.] Abraham, W. C. \& Bear, M. F. Metaplasticity: the plasticity of synaptic plasticity. \emph{Trends Neurosci.} \textbf{19}, 126–130 (1996).
\item[76.] Kumaran, D. \& Maguire, E. A. Match–mismatch processes underlie human hippocampal responses to associative novelty. \emph{J. Neurosci.} \textbf{27}, 8517–8524 (2007).
\item[77.] Lisman, J. E. \& Grace, A. A. The hippocampal-VTA loop: controlling the entry of information into long-term memory. \emph{Neuron} \textbf{46}, 703–713 (2005).
\item[78.] Hasselmo, M. E. The role of acetylcholine in learning and memory. \emph{Curr. Opin. Neurobiol.} \textbf{16}, 710–715 (2006).
\end{itemize}
}

\end{document}